\documentclass[journal]{IEEEtran}

\usepackage{times}
\usepackage{epsfig}
\usepackage{graphicx}
\usepackage{amsmath}
\usepackage{amssymb}
\usepackage{enumitem}
\usepackage{tabu}
\usepackage{booktabs}
\usepackage{multirow}
\usepackage{array}
\usepackage{adjustbox}
\usepackage{xcolor}
\usepackage{cite}
\usepackage{url}

\newcommand{\red}[1]{\color{red}{#1}}
\newcommand{\blue}[1]{\underline{\color{blue}{#1}}}

\definecolor{XH_color}{rgb}{1, 0.27, 0}

\begin{document}
%
% paper title
% Titles are generally capitalized except for words such as a, an, and, as,
% at, but, by, for, in, nor, of, on, or, the, to and up, which are usually
% not capitalized unless they are the first or last word of the title.
% Linebreaks \\ can be used within to get better formatting as desired.
% Do not put math or special symbols in the title.
\title{GridDehazeNet+: An Enhanced Multi-Scale Network with Intra-Task Knowledge Transfer for Single Image Dehazing}
%
%
% author names and IEEE memberships
% note positions of commas and nonbreaking spaces ( ~ ) LaTeX will not break
% a structure at a ~ so this keeps an author's name from being broken across
% two lines.
% use \thanks{} to gain access to the first footnote area
% a separate \thanks must be used for each paragraph as LaTeX2e's \thanks
% was not built to handle multiple paragraphs
%

\author{Xiaohong~Liu,~
Zhihao Shi,~
Zijun Wu, and~Jun~Chen,~\IEEEmembership{Senior Member,~IEEE}% <-this % stops a space
\thanks{X. Liu, Z. Shi, Z. Wu, and J. Chen (Corresponding Author) are with the Department of Electrical and Computer Engineering, McMaster University, Hamilton, ON., L8S 4K1, Canada (e-mail: \{liux173,  shiz31, wuz146, chenjun\}@mcmaster.ca). This work was
supported in part by the Natural Sciences and Engineering Research Council
of Canada through a Discovery Grant.}}% <-this % stops a space
\maketitle

% As a general rule, do not put math, special symbols or citations
% in the abstract or keywords.
\begin{abstract}
We propose an enhanced multi-scale network, dubbed GridDehazeNet+, for single image dehazing. The proposed dehazing method does not rely on the Atmosphere Scattering Model (ASM), and an explanation as to why it is not necessarily performing the dimension reduction offered by this model is provided. GridDehazeNet+ consists of three modules: pre-processing, backbone, and post-processing. The trainable pre-processing module can generate learned inputs with better diversity and more pertinent features as compared to those derived inputs produced by hand-selected pre-processing methods. 
The backbone module implements multi-scale estimation with two major enhancements: 1) a novel grid structure that effectively alleviates the bottleneck issue via dense connections across different scales; 2) a spatial-channel attention block that can facilitate adaptive fusion by consolidating dehazing-relevant features. The post-processing module helps to reduce the artifacts in the final output. 
Due to domain shift, the model trained on synthetic data may not generalize well on real data. To address this issue, we shape the distribution of synthetic data to match that of real data, and use the resulting translated data to finetune our network. We also propose a novel intra-task knowledge transfer mechanism that can memorize and take advantage of synthetic domain knowledge to assist the learning process on the translated data. Experimental results demonstrate that the proposed method outperforms the state-of-the-art on several synthetic dehazing datasets, and achieves the superior performance on real-world hazy images after finetuning. 
Codes will be released online upon the publication of the present work.

%can be easily embedded in our network to adapt feature fusions, where the features advantageous to dehazing are consolidated. 

\end{abstract}


%%%%%%%%% BODY TEXT

\section{Introduction}

% a general introduction

\IEEEPARstart{T}{he} image dehazing problem has received significant attention in the research community~\cite{zhang2019joint,zhang2019single,yin2019color,zhao2020pyramid} over the past two decades. It aims to restore the clear version of a hazy image, thus helps mitigate the impact of image distortion induced by the environmental conditions on various visual analysis tasks.

The Atmosphere Scattering Model (ASM)~\cite{scatteringfn01, scatteringfn02, scatteringfn03} provides a simple approximation of the haze effect. Specifically, it assumes that
\begin{align}
	I_c(x)=J_c(x)t(x)+A(1-t(x)),\quad c=1,2,3,\label{eq:asm}
\end{align}
where $I_c(x)$ ($J_c(x)$) is the intensity of the $c$th color channel of pixel $x$ in the hazy (clear) image, $t(x)$ is the transmission map, and $A$ is the global atmospheric light intensity. In addition, we have $t(x)=e^{-\beta d(x)}$, where $\beta$ and $d(x)$ are the atmosphere scattering parameter and the scene depth, respectively. %$d(x)$ as follows
%\begin{align*}
%t(x)=e^{-\beta d(x)}.
%\end{align*}
This model indicates that image dehazing is in general an underdetermined problem without the knowledge of $A$ and $t(x)$.

\begin{figure}[t]
	\centering
	
	\begin{minipage}[h]{0.49\linewidth}
		\centering
		\includegraphics[width=\linewidth]{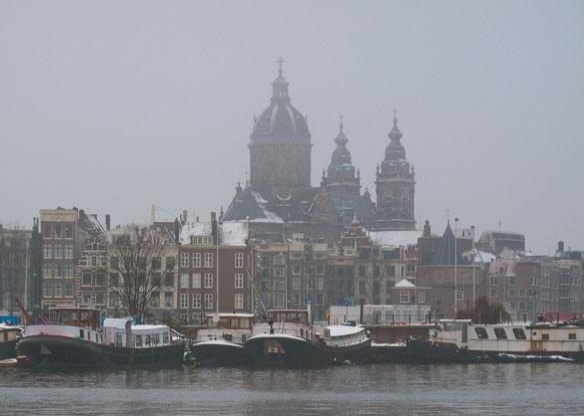}
		\scriptsize{(a) Real hazy image}
	\end{minipage}
	\begin{minipage}[h]{0.49\linewidth}
		\centering
		\includegraphics[width=\linewidth]{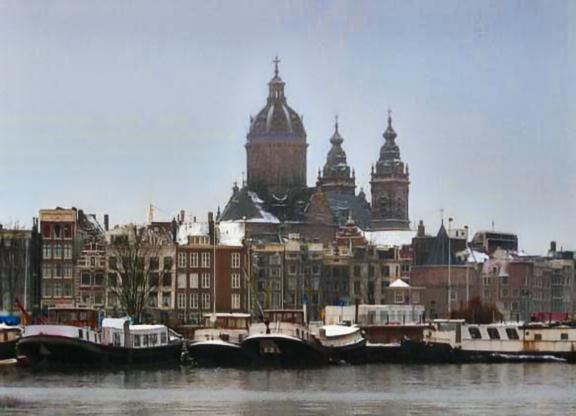}
		\scriptsize{(b) DADN~\cite{shao2020domain}}
	\end{minipage}
	\begin{minipage}[h]{0.49\linewidth}
		\centering
		\includegraphics[width=\linewidth]{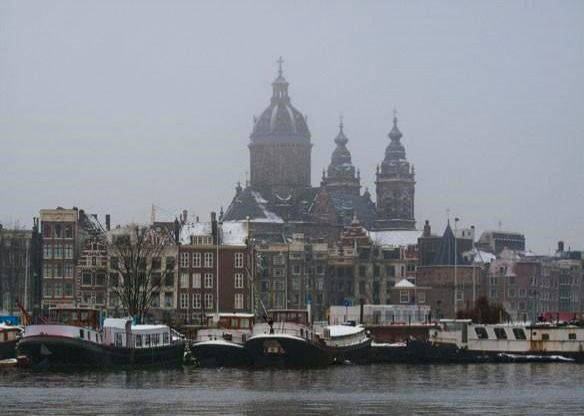}
		\scriptsize{(c) ACER-Net~\cite{wu2021contrastive}}
	\end{minipage}
	\begin{minipage}[h]{0.49\linewidth}
		\centering
		\includegraphics[width=\linewidth]{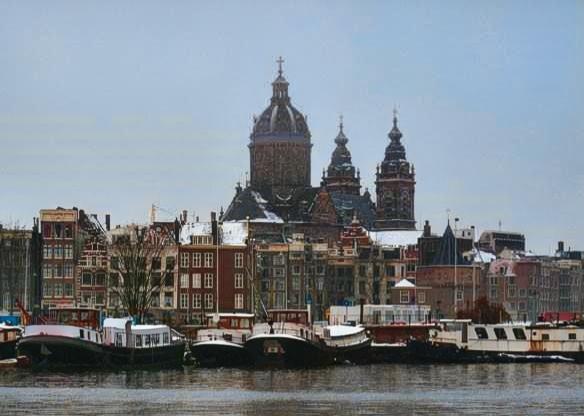}
		\scriptsize{(d) GDN+}
	\end{minipage}
	%	\begin{minipage}[h]{0.32\linewidth}
	%		\centering
	%		\includegraphics[width=\linewidth]{images/first_page/gt_0001.png}
	%		\scriptsize{(c) Ground truth}
	%	\end{minipage}
%	\vspace{0.3em}
	\caption{Dehazing results for a real hazy image from URHI dataset~\cite{li2019benchmarking}: (a) a real hazy image, (b) the result based on DADN~\cite{shao2020domain}, (c) the result based on ACER-Net~\cite{wu2021contrastive}, and (d) our result. The GDN+ achieves the best visual performance against the others in terms of haze removal and enhanced color contrast.}
	\label{fig:first_page}
%	\vspace{-1.5em}
\end{figure}

As a canonical example of image restoration, the dehazing problem can be tackled using a variety of techniques that are generic in nature. Moreover, many misconceptions and difficulties encountered in image dehazing manifest in other restoration problems as well.
Therefore, it is instructive to examine the relevant issues in a broader context, four of which are highlighted below.

\subsubsection{Role of Physical Model} Many data-driven approaches to image restoration require synthetic data for training. To produce such data, it is necessary to have a physical model of the relevant image degradation process ({\textit{e}.\textit{g}.,  the ASM for the haze effect).  A natural question arises whether the design of the image restoration algorithm itself should rely on this physical model. Apparently a model-dependent algorithm may suffer inherent performance loss on real images due to model mismatch. However, it is often taken for granted that such an algorithm must have advantages on synthetic images produced using the same physical model.
	
%Apparently a model-dependent algorithm may suffer inherent performance loss on real-world images due to model mismatch. However,
	
\subsubsection{Selection of Pre-Processing Method} Pre-processing is widely used in image preparation to facilitate follow-up operations~\cite{tong2017image, iebmgated01}. It can also be used to generate several variants of the given image, providing a certain form of diversity that can be harnessed via proper fusion. However, the pre-processing methods are often selected based on heuristics, thus are not necessarily best suited to the problem under consideration.

\subsubsection{Bottleneck of Multi-Scale Estimation} 	
Image restoration requires an explicit/implicit knowledge of the statistical relationship between the distorted image and the original clear version. The statistical model needed to capture this relationship often has a huge number of parameters, comparable or even more than the available training data. As such, directly estimating these parameters based on the training data is often  unreliable. Multi-scale estimation~\cite{shen2018deep, chen2018learning} tackles this problem by i) approximating the high-dimensional statistical model with a low-dimensional one, ii) estimating the parameters of the low-dimensional model based on the training data, iii) parameterizing the neighborhood of the estimated low-dimensional model,  performing a refined estimation, and repeating this procedure if needed. It is clear that the estimation accuracy on one scale will affect that on the next scale. Since multi-scale estimation is commonly done in a successive manner, its performance is often limited by a certain bottleneck.

\subsubsection{Effect of Domain Shift}
The effectiveness of supervised learning for image restoration has been widely observed. However, building a large-scale real dataset of distorted images paired with their ground-truth is very expensive and sometimes not even possible~\cite{zhu2017unpaired}. 
Therefore, in practice one commonly resorts to synthetic data for network training. However, due to  domain shift, there is no guarantee that a network trained on synthetic data can generalize well to real data.

%Since the synthesized haze appearance might not conform to the real one, a network trained on synthetic data is probably not generalized well on real data in testing.

% With the supervised learning, promising restoration results have been achieved. However, collecting a large-scale dataset that pairs the distorted image and its corresponding ground-truth is expensive or even not practical~\cite{zhu2017unpaired}. Therefore, synthetic data is commonly utilized to train the network. %generated from a mathematical formula that simulates the degradation process. 
%Since the synthesized haze appearance might not conform to the real one, a network trained on synthetic data is probably not generalized well on real data in testing. %Owing to domain shift, the better dehazing performance we achieve on synthetic data, the worse we have on real data
%Therefore, an effective way should be proposed to deal with this issue.

The present work can be viewed as a product of our attempt to address the aforementioned generic issues in image restoration.  Its main contributions can be summarized as follows:
\begin{enumerate}
\item The proposed GridDehazeNet+ (abbreviated as GDN+)  does not rely on the ASM for haze removal, yet is capable of outperforming the existing model-dependent dehazing methods even on synthetic images. We also experimentally demonstrate that the dimension reduction offered by the ASM is not necessarily beneficial to  network learning, owing to the introduction of undesirable local minima.

%A possible explanation, together with some supporting experimental results, is provided for this puzzling phenomenon.

\item In contrast to hand-selected pre-processing methods, the pre-processing module in the GDN+ is fully trainable, thus can offer more flexible and pertinent image enhancement.

\item The implementation of attention-based multi-scale estimation on a densely connected grid network allows efficient information exchange across different scales and alleviates the bottleneck issue. 

\item To cope with domain shift, certain translated data are generated, by shaping the distribution of synthetic data to match that of real-world hazy images, to finetune our network. Moreover, a novel Intra-Task Knowledge Transfer (ITKT) mechanism is proposed to help the finetuning process on translated data.

%\item To alleviate the domain shift and improve the dehazing performance on real hazy images, a novel intra-task knowledge transferring is proposed that leverages the distilled knowledge from synthetic data to assist the learning process on translated data.
  
\end{enumerate}

Benefiting from the overall design, the proposed GDN+ outperforms the State-Of-The-Art (SOTA) methods on several synthetic dehazing datasets and achieves superior performance on real-world hazy images after finetuning.
An example is shown in Fig.~\ref{fig:first_page}, where our method delivers the most visually appealing dehazing result for a real hazy image from URHI dataset~\cite{li2019benchmarking}.

%the real hazy image after dehazing has better visibility than the current SOTA~\cite{shao2020domain}.
\section{Related Work}
Early works on image dehazing either require \textit{multiple} images of the same scene taken under different conditions~\cite{multiimagepolar01, multiimagepolar02, scatteringfn02, multiimageweather02, multiimageweather03} or side information acquired from other sources~\cite{multiimagedepth02,multiimagedepth01}.  Recent years have seen increasing interest in  \textit{single} image dehazing without side information, which is considerably more challenging. 
To place our work in a proper context,
we give a review of existing prior-based and learning-based methods for single image dehazing as well as the recent developments of knowledge distillation and transfer. 

\subsection{Prior-Based Single Image Dehazing}

A conventional strategy for single image dehazing is to estimate the transmission map $t(x)$ and the global atmospheric light intensity $A$ (or their variants) based on certain assumptions or priors. Then, Eq.~(\ref{eq:asm}) is inverted to obtain the dehazed image. Representative works along this line of research include~\cite{iebmcontrast01,iebmalbedo01,iebmdcp01,iebmrf01,iebmlinear01}. Specifically, \cite{iebmcontrast01} proposed a local contrast maximization method for dehazing, motivated by the observation that clear images tend to have higher contrast as compared to their hazy counterparts. \cite{iebmalbedo01} realized haze removal via the analysis of albedo under the assumption that the transmission map and surface shading are locally uncorrelated. \cite{iebmdcp01} proposed the Dark Channel Prior (DCP), which asserted that pixels in non-haze patches have low intensity in at least one color channel. \cite{iebmrf01} suggested a machine learning approach that exploits four haze-related features using a random forest regressor. \cite{iebmlinear01} proposed a color attenuation prior that is beneficial to modeling the scene depth of hazy images. Although these methods have enjoyed varying degrees of success, their performances are inherently limited by the accuracy of the adopted assumptions/priors with respect to the target scenes.

\subsection{Learning-Based Single Image Dehazing}
With the advance in deep learning techniques and the availability of large synthetic datasets~\cite{iebmrf01}, recent years have witnessed the increasing popularity of data-driven methods for image dehazing. These methods largely follow the conventional strategy mentioned above but with reduced reliance on hand-crafted priors. For example, \cite{iebmmscnn01} employed a Multi-Scale CNN (MSCNN) that first predicted a holistic transmission map, and refined it locally. \cite{iebmdehazenet01} proposed a three-layer Convolutional Neural Network (CNN), named DehazeNet, to directly estimate the transmission map from the given hazy image. \cite{zhang2018densely} embedded the ASM into a neural network for joint learning of the transmission map, atmospheric light intensity, and dehazed result. \cite{dong2020physics} explored the physical model in the feature space (instead of the pixel space) to perform image dehazing.
 
  %applied the physical model on the feature space to explore useful features for image dehazing.
  
  % Instead of considering the physical model in the image space, 

The AOD-Net~\cite{iebmaod01} represents a departure from the conventional strategy. Specifically, a reformulation of Eq.~(\ref{eq:asm}) was introduced in~\cite{iebmaod01} to bypass the estimation of the transmission map and  atmospheric light intensity. 
A close inspection reveals that this reformulation in fact renders the ASM completely superfluous (though this point is not recognized in~\cite{iebmaod01}). In~\cite{iebmgated01}, the proposed Gated Fusion Network (GFN) went one step further by explicitly abandoning the ASM in its design, and leveraged several hand-selected pre-processing methods (\textit{i.e.}, white balance, contrast enhancing, and gamma correction) to improve the dehazing results. Recent works mostly followed this model-agnostic design principle and tackled image dehazing with various techniques. By regarding image dehazing as image-to-image translation, \cite{qu2019enhanced} constructed an Enhanced Pix2pix Dehazing Network (EPDN) based on the Generative Adversarial Network (GAN), which does not rely on any physical model.  \cite{qin2020ffa} capitalized on the attention mechanism and put forward a feature fusion attention network with the flexibility to regulate different types of information. By leveraging the boosting strategy, \cite{dong2020multi} proposed a boosted decoder that can progressively restore the haze-free image. \cite{wu2021contrastive} treated hazy and clear images as negative and positive samples to train the proposed AECR-Net jointly, and the adopted contrastive regularization can be applied to other dehazing methods to further improve their performance.

While there is increasing evidence that model-agnostic image dehazing methods are able to outperform their model-dependent counterparts even if only synthetic data (produced using the physical model) are concerned, the reason behind this puzzling phenomenon is still unclear. In this paper, we attempt to lift the veil by providing a possible explanation together with some supporting experiments. 

%Since collecting a large set of hazy-clean image pairs in  real-world scenarios is a formidable task, 

In addition, owing to domain shift, learning-based methods trained on synthetic data tend to generalize poorly over to real data. To mitigate the detrimental effect caused by domain shift, \cite{li2019semi} proposed a hybrid approach, where a CNN is trained on synthetic data in a supervised manner, and on real data in an unsupervised manner. To support  unsupervised learning, physical priors (\textit{i.e.}, dark channel loss and total variation loss) were employed. \cite{chen2021psd} followed this line of ideas and proposed a principled synthetic-to-real dehazing framework to finetune a model trained on synthetic data, aiming at improving the generalization performance on real data. However, involving real data in training does not fully address the domain-shift issue. In~\cite{shao2020domain}, a Domain Adaptation Dehazing Network (DADN) was proposed by adopting the CycleGAN~\cite{zhu2017unpaired} to deal with the discrepancies between the synthetic domain and real domain.

%\XH{Instead of leveraging CycleGAN to shape the distribution of synthetic data to real data, \cite{chen2021psd} proposes a Principled Synthetic-to-real Dehazing (PSD) framework that adopts three physical-prior-based losses (\textit{i.e.,} dark channel prior loss, bright channel prior loss, and contrast limited adaptive histogram equalization loss) to finetune the model trained on synthetic images, aiming at improving the generalization performance on real-world ones.}
	
In view of the fact that  unsupervised finetuning guided by physical priors may cause significant artifacts, the GDN+  proposed in the present paper exploits supervised finetuning on translated data to improve the dehazing performance on real data. 

%As the translated data created by CycleGAN is quite natural without perceptible artifact, rather than leveraging the physical priors

%We also show how to use so-called translated data to reduce the domain gap

 %introduced a domain adaptation paradigm that employed the CycleGAN~\cite{zhu2017unpaired} to reduce domain shift between synthetic and real data.

%employed a semi-supervised learning algorithm, where a CNN is trained on synthetic data in a supervised manner, and on real data in an unsupervised manner. However, involving real data in training does not fully address the domain-shift issue. Their subsequent work~\cite{shao2020domain} introduced a domain adaptation paradigm that employed the CycleGAN~\cite{zhu2017unpaired} to reduce domain shift between synthetic and real data.

%also propose an enhanced grid network and use the translated images from real domain for a supervised learning to further improve the dehazing performance on real scenario. 

%While building model-agnostic methods for single image dehazing has become a growing trend since more appealing results are achieved as compared to the ones from model-dependent methods even if only the synthetic data are concerned, the reason behind this puzzling phenomenon is still unclear. In this paper, we attempt to lift the veil by providing a possible explanation with some supporting experiments. We also propose an enhanced grid network and use the translated images from real domain for a supervised learning to further improve the dehazing performance on real scenario. 

\subsection{Knowledge Distillation and Transfer}

One popular application of knowledge distillation~\cite{hinton2015distilling} is for network compression, where the learned logits from a large network (\textit{i.e.,} teacher network) is transferred to a small network (\textit{i.e.,} student network). Compared to the teacher network, the student network is much easier to deploy, possibly at the cost of a potential performance drop. \cite{romero2014fitnets} suggested that the intermediate representations from the teacher network can be leveraged to further improve the training process of the student network. In recent years, knowledge distillation has been proved useful not only for network compression, but also for various computer vision tasks, including object detection~\cite{wang2019distilling}, semantic segmentation~\cite{wang2020intra}, image synthesis~\cite{yin2020dreaming},  style transfer~\cite{chen2020optical},~\textit{etc.} Knowledge distillation found its first application to single image dehazing in \cite{wu2020knowledge}, where the teacher and student networks share the same architecture but are responsible for image reconstruction and image dehazing tasks, respectively. In contrast, for the Knowledge Distilling Dehazing Network (KDDN) proposed in \cite{hong2020distilling}, the architectures of teacher and student networks are tailored to the designated tasks; besides, multiple features, rather than only one intermediate feature, are distilled to improve the effectiveness of knowledge transfer.

%instead of distilling one intermediate feature  to~\cite{wu2020knowledge} that only distilled one intermediate feature, multiple features were distilled in~\cite{hong2020distilling} to yield better dehazing performance. 

Different from~\cite{wu2020knowledge, hong2020distilling}, where knowledge transfer is carried out among heterogeneous tasks, we perform  ITKT with teacher and student networks  working on the \textit{same} task (\textit{i.e}, dehazing) but taking \textit{different} data as inputs. Intuitively, the synthetic domain knowledge yields useful insights into translated data, where the haze effect does not admit a simple mathematical characterization. Therefore, the characteristics of intermediate features distilled from the teacher network can greatly benefit the learning process of the student network, enabling it to deliver satisfactory dehazing results on real-world hazy images.

%which further improves the dehazing performance of our network in real-world scenarios.

 %hazy inputs from synthetic and real domains. Since the synthetic data is usually generated by the ASM, it is easier to be tackled than the translated data in real domain, where the haze effect is hard to be modeled by a simple mathematical formula. Therefore, distilling the characteristic of intermediate features from the teacher network on synthetic data can greatly help the learning process of the student network on translated data, which further improves the dehazing performance of our network in real scenario.

% To align the intermediate features, the same architecture is adopted for teacher and student networks in this paper.
\section{Method}

\subsection{Overview}

Here we highlight the following aspects of the proposed GDN+.

\subsubsection{No Reliance on Atmosphere Scattering Model} Although  the model-agnostic approach to single image dehazing has become increasingly popular, no convincing reason has been provided why there is any advantage in ignoring the ASM, as far as the dehazing performance on synthetic images is concerned.
The argument put forward in~\cite{iebmgated01} is that estimating $t(x)$ from a hazy image is an ill-posed problem. Nevertheless, this is puzzling since estimating $t(x)$ (which is color-channel-independent) is presumably easier than $J_c(x)$, $c=1,2,3$. 
In Fig.~\ref{fig:losssurface}, we offer a possible explanation why it could be problematic if one blindly uses the constraint that $t(x)$ is color-channel-independent to narrow down the search space and why it might be potentially advantageous to relax this constraint in the search of the optimal $t(x)$. However, with this relaxation, the ASM offers no dimension reduction in the estimation procedure. 
More fundamentally, it is known that the loss surface of a CNN is generally well-behaved in the sense that the local minima are often almost as good as the global minimum~\cite{choromanska2015loss,draxler2018essentially,nguyen2018loss}. On the other hand, by incorporating the ASM into a CNN, one basically introduces a nonlinear component that is heterogeneous in nature from the rest of the network, which may create an undesirable loss surface. To support this explanation, we provide some experimental results in Section \ref{sec:asm}.

\begin{figure}[t]
	\centering
	\begin{minipage}[h]{0.49\linewidth}
		\centering
		\includegraphics[width=\linewidth]{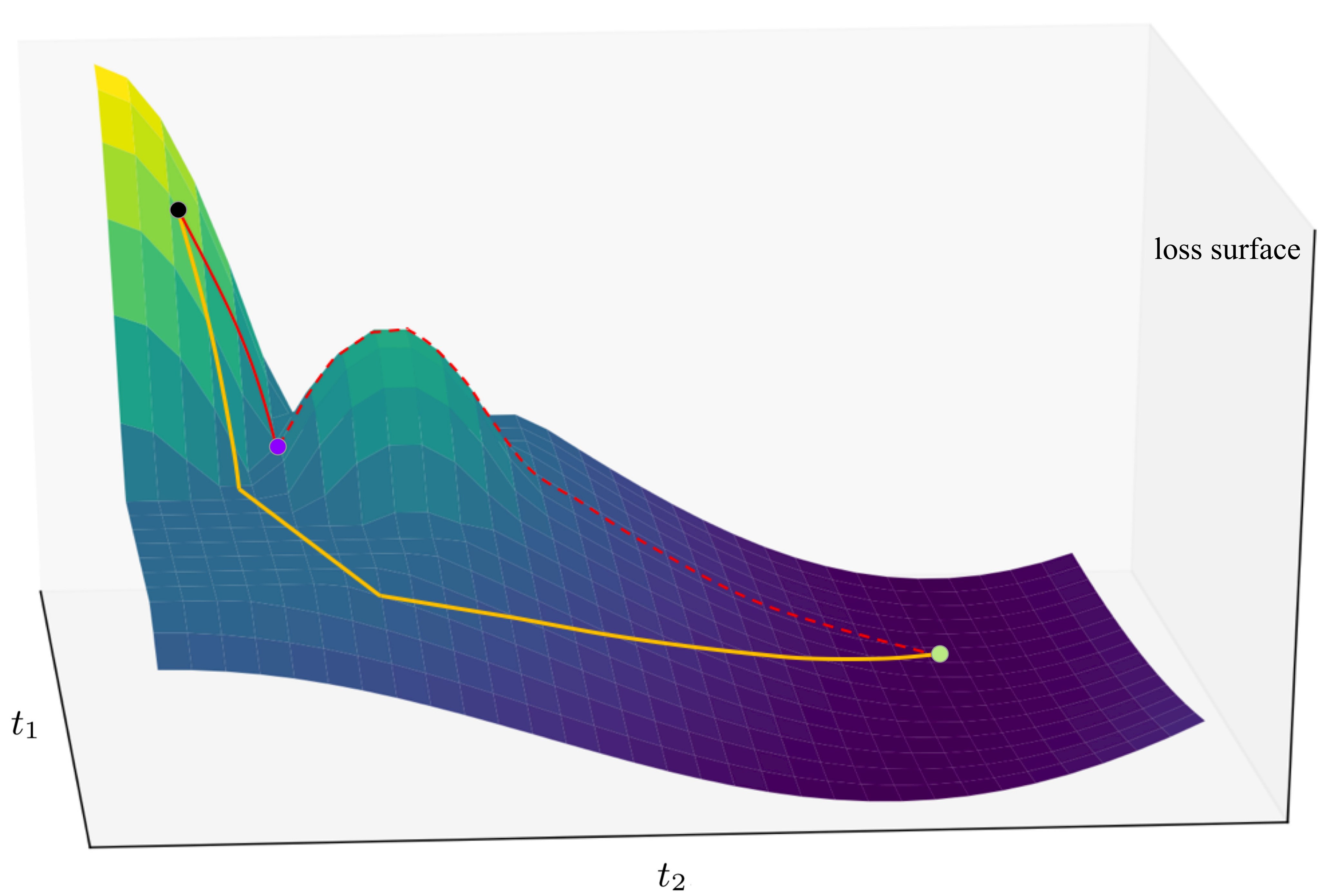}
		\scriptsize{(a) Loss surface}
	\end{minipage}
	\begin{minipage}[h]{0.49\linewidth}
		\centering
		\includegraphics[width=\linewidth]{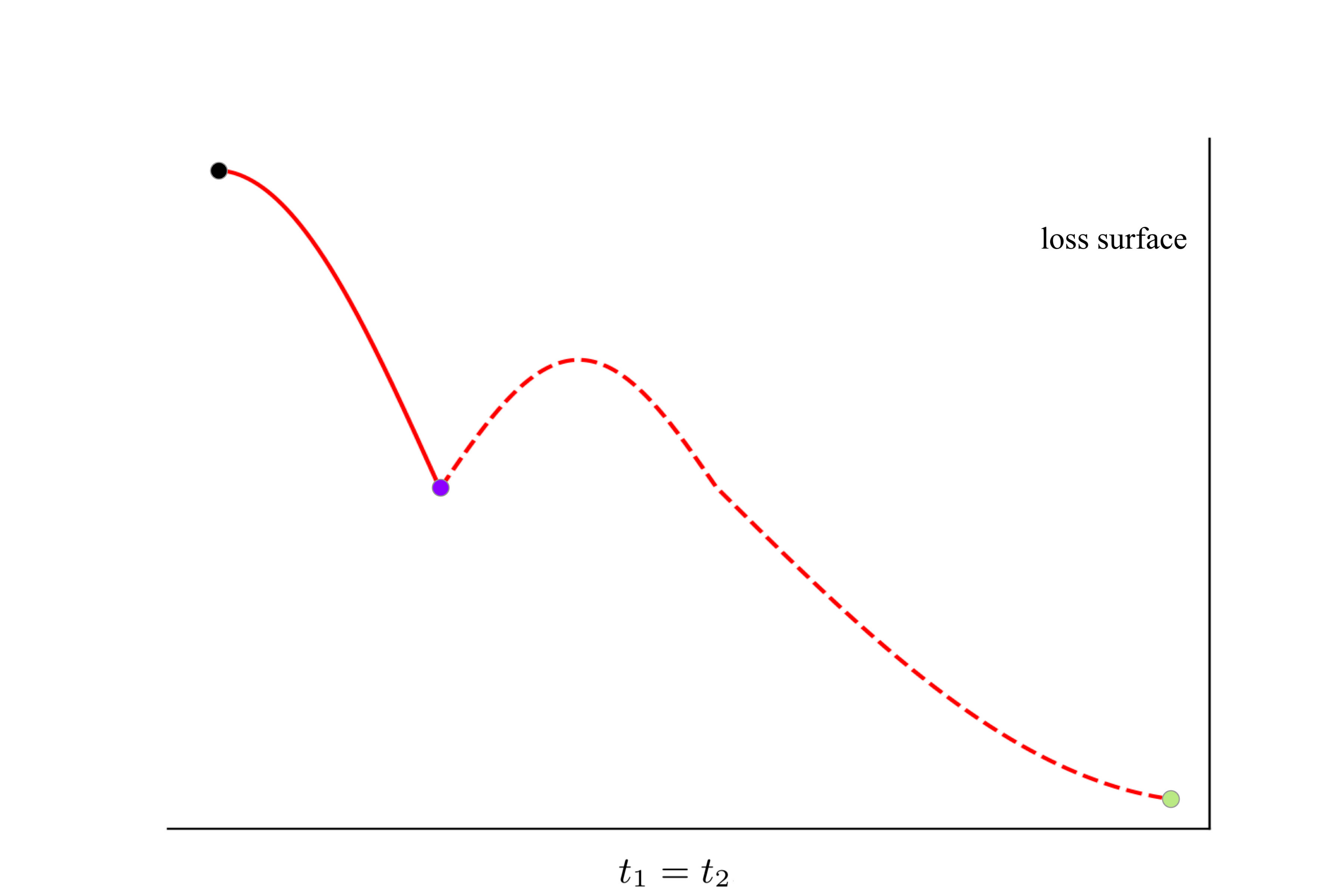}
		\scriptsize{(b) Constrained loss surface}
	\end{minipage}
	\caption{On the potential detrimental effect of using the ASM for image dehazing. For illustration purposes, we focus on two color channels of a single pixel and denote the respective transmission maps by $t_1$ and $t_2$. Fig.~\ref{fig:losssurface} (a) plots the loss surface as a function of $t_1$ and $t_2$. It can be seen that the global minimum is attained at a point (see the green dot) satisfying $t_1=t_2$, which agrees with the ASM. With the black dot as the starting point, one can readily find this global minimum using gradient descent (see the yellow path). However, a restricted search based on the ASM along the $t_1=t_2$ direction (see the red path) will get stuck at a point indicated by the purple dot (see Fig.~\ref{fig:losssurface} (b)). Note that this point is a local minimum in the constrained space but not in the original space, and it becomes an obstruction simply due to the adoption of the ASM.}
	\label{fig:losssurface}
\end{figure}

\begin{figure*}[t]
	\centering
	\includegraphics[width=\linewidth]{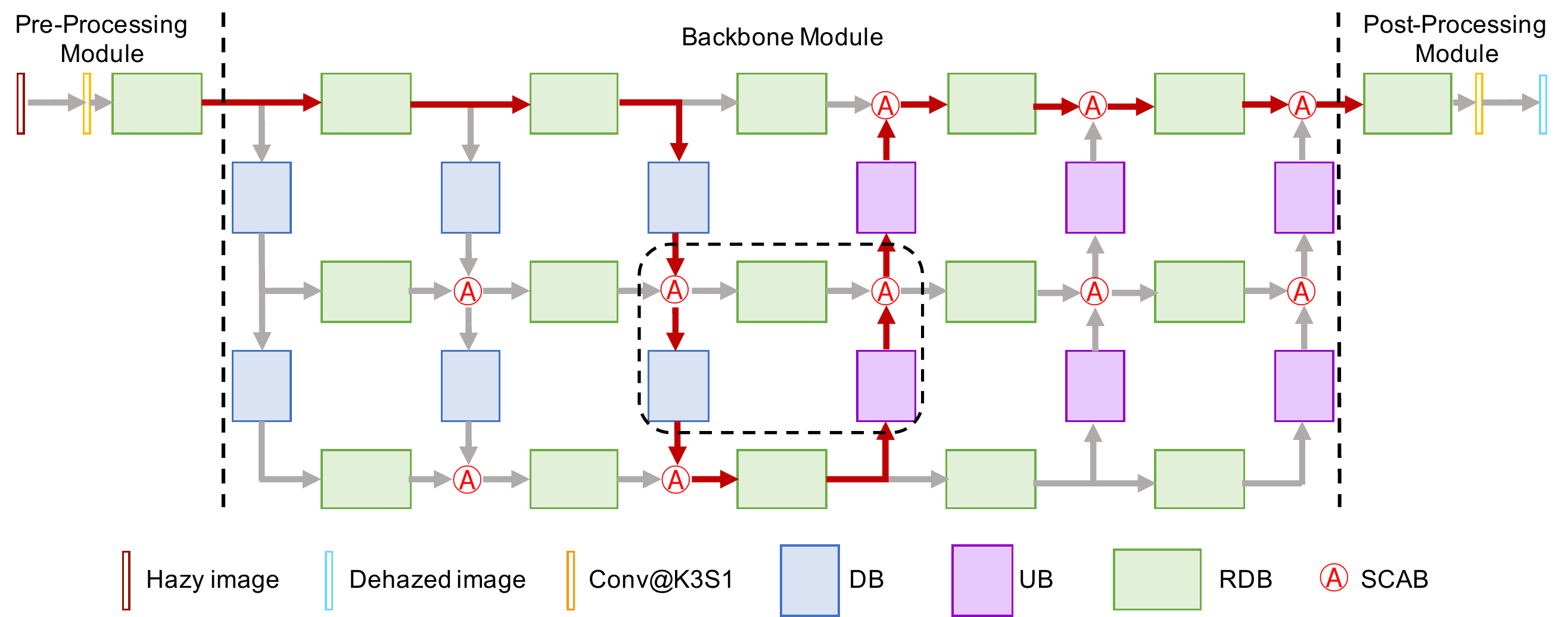}
	\caption{The architecture of the proposed GridDehazeNet+ (GDN+). Here Conv@K$n$S$m$ indicates a $n \times n$ convolution with stride $m$.}
	\label{fig:GDN_main}
\end{figure*}

\begin{figure}[t]
	\centering
	\includegraphics[width=\linewidth]{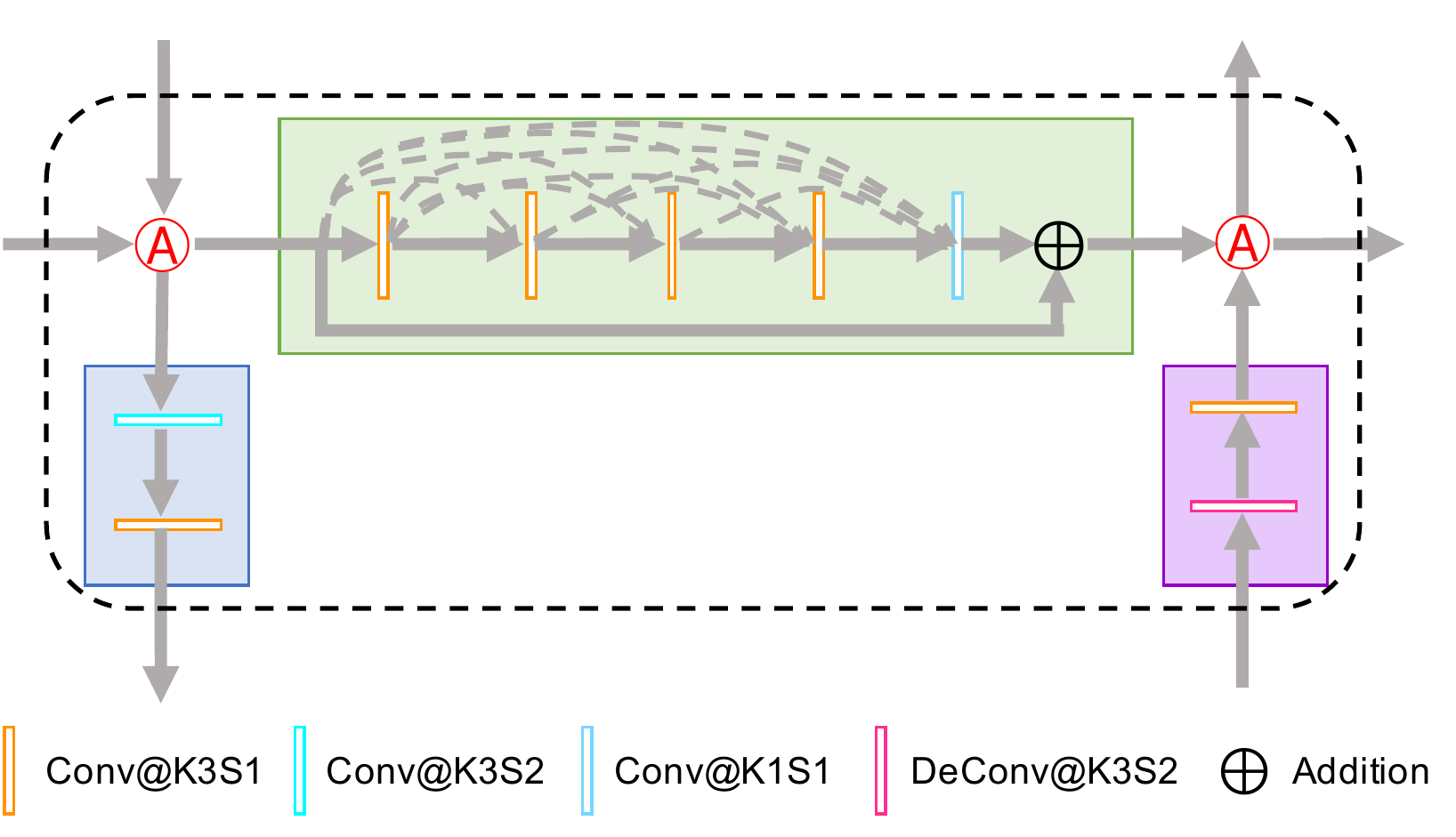}
	\caption{Illustration of the  dashed box in Fig.~\ref{fig:GDN_main}. Here Conv(DeConv)@K$n$S$m$ indicates a $n \times n$ convolution (deconvolution) with stride $m$.}
	\label{fig:GDN_block}
\end{figure}

\subsubsection{Trainable Pre-Processing Module} The pre-processing module effectively converts the single image dehazing problem to a multi-image dehazing  problem by generating several variants of the given hazy image, each highlighting a different aspect of this image and making the relevant feature information more evidently exposed. In contrast to those hand-selected pre-processing methods adopted in the existing works ({\textit{e}.\textit{g}., \cite{iebmgated01}), the proposed pre-processing module is made fully trainable, which is in line with the general preference of  data-driven methods over prior-based methods as shown by recent developments in image dehazing.	
Note that hand-selected processing methods typically aim to enhance certain concrete features that are visually recognizable. However, the exclusion of abstract features is not justifiable. Indeed, there might exist abstract transform domains that better suit the follow-up operations than the image domain.
A trainable pre-processing module has the freedom to identify transform domains over which more diversity gain can be harnessed.
	
\subsubsection{Enhanced Multi-Scale Estimation} Here the meaning of word \textit{enhanced} is two-fold. First, inspired by \cite{iebmgridnet01}, we \textit{enhance} the conventional multi-scale network using a novel grid structure. This grid structure has clear advantages over the encoder-decoder structure and the conventional multi-scale structure extensively used in image restoration~\cite{iebmkpn01,iebmrdn01,Tao_2018_CVPR,iebmgated01}. In particular, the information flow in the encoder-decoder structure or the conventional multi-scale structure often suffers from the bottleneck effect due to the hierarchical architecture whereas the grid structure circumvents this issue via dense connections across different scales using up-sampling/down-sampling blocks. Second, we further \textit{enhance} the network with  Spatial-Channel Attention Blocks (SCABs) that are placed at the junctions where features are exchanged and aggregated. These SCABs enable the network to better exploit the diversity created by the pre-processing module and the information most relevant to the dehazing task. 

%can be easily embedded in the position where the feature is exchanging and aggregating. This attention block enables the network to better harness the feature that is more significant to dehazing process.

\subsubsection{Intra-Task Knowledge Transfer} ITKT refers to leveraging the knowledge acquired from a certain task on \textit{one dataset} to facilitate the learning process of the same task on \textit{another dataset}. In the current context, it is observed that the synthetic domain knowledge is beneficial for handling translated data. Rather than directly finetuning the network on translated data, a teacher-student structure is used to memorize and take advantage of synthetic domain knowledge. To the best of our knowledge, this is the first work that leverages ITKT to improve the dehazing performance on real-world hazy images.

%with the identical architecture for the sake of facilitating the alignment of their intermediate features.

\begin{figure*}[t]
	\centering
	\includegraphics[width=\linewidth]{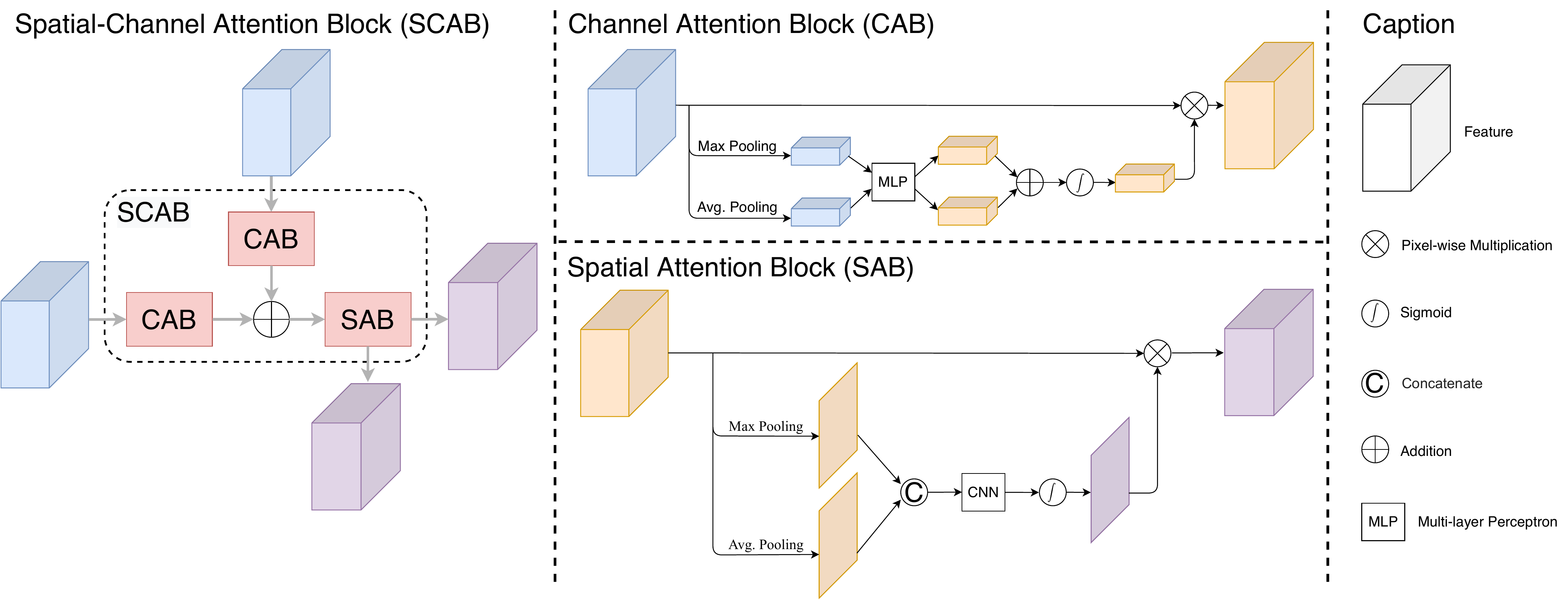}
	\caption{Illustration of the Spatial-Channel Attention Block (SCAB).}
	\label{fig:SCAB}
\end{figure*}

In comparison to the preliminary work GDN~\cite{liu2019griddehazenet}, the GDN+ is improved in two aspects. First, the GDN only adopts channel-wise attention with the learned weights independent to the target features~\cite{liu2019griddehazenet}.
In contrast, the GDN+ employs the self-attention mechanism~\cite{wang2018non, woo2018cbam}, encapsulated in SCABs, to generate feature-adaptive weights. Second, the GDN tends to suffer significant performance drop on real-world hazy images, possibly due to the domain shift between synthetic data in training and real data in testing.
To address this issue, we shape the distribution of synthetic data to match that of real data, and use the resulting translated data to finetune the network. Moreover, to memorize and take advantage of synthetic domain knowledge, we propose an ITKT mechanism to assist the learning process on translated data.

%Our previous work (\textit{i.e.,} GDN)~\cite{liu2019griddehazenet} exists two underlying drawbacks. One is that although the channel-wise attention adopted in~\cite{liu2019griddehazenet} is effective, it is quite simple, where the attention weight is learned independently, and not associated with the applied feature \textit{per se}. Therefore, the learned weight only reach the equilibrium to achieve the relatively best dehazing performance on the entire training dataset, not adaptive to each of them. In recent years, the self-attention~\cite{wang2018non,woo2018cbam} has achieved promising results, and can be employed for image dehazing. 
%Since the real haze might be anisotropic, 
%Therefore, we elaborately designed a spatial-channel attention block (SCAB) that can be effortlessly embedded in our network.
%different from~\cite{liu2019griddehazenet} that only performed the channel-wise attention. 

% is that~\cite{liu2019griddehazenet} is not generalized well while dealing with real hazy images. A possible reason is the domain shift in which synthetic hazy images are used in training, but real hazy images are considered in testing. To address this issue, inspired by~\cite{shao2020domain}, we convert synthetic data to translated data that has the same haze distribution as real data. This translated data is then utilized in training. In addition, we noticed that the learned knowledge from synthetic data can be regarded as a prior that helps the learning process on translated data. Therefore, a teacher-student structure is leveraged to transfer this knowledge.
	
\subsection{Network Architecture} \label{3.2}

The GDN+ consists of three modules, \textit{i.e}, the pre-processing module, the backbone module and the post-processing module. %Let $I(x)$ and $\hat{J}(x)$ denote the input hazy image and output dehazed image in RGB color space respectively. %The functional representation of the proposed GridDehazeNet with end-to-end relation can be formulated as 
%\begin{equation}
%\hat{J}(x) = F_{GDN}(I(x); \theta_{GDN}),
%\end{equation}
%where $F_{GDN}$ represents the dehazing process in GridDehazeNet and $ \theta_{GDN}$ denotes the ensemble of the network parameters. 
Fig.~\ref{fig:GDN_main} shows the overall architecture of the proposed network. 

The pre-processing module consists of a $3 \times 3$ convolution with stride $1$ (denoted as Conv@K$3$S$1$) and a Residual Dense Block (RDB)~\cite{iebmrdn01}. 
% Except for this convolution, all other convolutions in our network are activated by ReLU function. 
It generates 16 feature maps, which will be referred to as the learned inputs, from the given hazy image.

The backbone module is an improved version of GridNet~\cite{iebmgridnet01} originally proposed for semantic segmentation. It performs  enhanced multi-scale estimation based on the learned inputs. We choose a grid structure with three rows and six columns. Each row corresponds to a different scale and consists of five RDBs with the number of feature maps unaltered. Each column can be regarded as a bridge that connects different scales via Upsampling Blocks (UBs) or Downsampling Blocks (DBs). In each UB (DB), the size of feature maps is increased (decreased) by a factor of 2 while the number of feature maps is decreased (increased) by the same factor. Here upsampling/downsampling is implemented  using  convolution instead of traditional methods such as bilinear or bicubic interpolation.
Fig.~\ref{fig:GDN_block} provides a detailed illustration of the RDB, UB, and DB in the dash box in Fig.~\ref{fig:GDN_main}. Each RDB consists of five convolutions: the first four are used to increase the number of feature maps while the last one fuses these feature maps. The output is then combined with the input of this RDB via channel-wise addition. Following~\cite{iebmrdn01}, the growth rate in RDB is set to 16. The UB and DB are structurally the same except that they respectively use Convolution (Conv) and DeConvolution (DeConv)) to adjust the size of feature maps. In GDN+, except for the first convolution in the pre-processing module and the $1\times1$ convolution in each RDB, all other convolutions are activated by ReLU.  To strike a balance between the output size and the computational complexity, we set the number of feature maps at three different scales to 16, 32, and 64, respectively.

Since dehazed images constructed directly from the output of the backbone module tend to contain artifacts, we introduce a post-processing module to further improve the quality. The structure of the post-processing module is symmetrical to that of the pre-processing module. 

It is worth noting that the GDN+ subsumes some existing networks as special cases. For example, the red path in Fig.~\ref{fig:GDN_main}  shows an encoder-decoder network that can be obtained by pruning the GDN+. As another example,  removing the exchange branches ({\textit{i}.\textit{e}.,  the middle four columns in the backbone module) from GDN+ leads to a conventional multi-scale network.
	
%	Indeed, if we prune (or deactivate) a portion of the GDN+, it can be turned into an encoder-decoder network or a conventional multi-scale network. 

\subsection{Feature Fusion with Spatial-Channel Attention Blocks} \label{3.3}

Since different channels and regions of learned features may not be of the same importance for the dehazing process, we embed certain judiciously constructed SCABs into the network to enable adaptive feature fusion. The SCAB employs spatial and channel-wise attentions~\cite{woo2018cbam}, realized respectively by the Spatial Attention Block (SAB) and the Channel Attention Block (CAB). The SAB applies the average and max poolings along the channel axis to aggregate the local information on different feature maps, and the two pooled results are concatenated and fed into a convolution to generate the spatial attention map. The CAB applies the average and max poolings along the spatial axis instead; the pooled features are adjusted by a shared multi-layer perceptron which explores  the inter-channel relationship to consolidate the important information; the adjusted versions are then added together and passed through a Sigmoid function to produce the channel attention coefficients. Finally, the spatial attention map and channel attention coefficients act back on the corresponding input features to enable self-adaptation.

As illustrated in Fig.~\ref{fig:SCAB}, each SCAB consists of two CABs and one SAB. The features from horizontal and vertical streams are first accommodated by two distinct CABs to strengthen the relevant characteristics via channel-wise attention. The outputs of the two CABs are added together and then fed into a SAB for spatial adaptation. Let $F_{i,j}^h$ and $F_{i,j}^v$ denote respectively the features from the horizontal stream and vertical stream at the fusion position $(i,j)$ in the backbone module, where $i=0,1,2$ and $j=0,1,\dotsc, 5$. Let  $f_{i,j}^h(F \mid \Theta_{i,j}^h)$ and $f_{i,j}^v(F \mid \Theta_{i,j}^v)$ denote respectively the CAB operations for the horizontal stream and vertical stream at the fusion position $(i,j)$, where $F$ represents an arbitrary input feature, and $\Theta_{i,j}^h$, $\Theta_{i,j}^v$ are the trainable weights. Similarly, let $g_{i,j}(F \mid \Phi_{i,j})$ denote the SAB operation at the fusion position $(i,j)$, where $\Phi_{i,j}$ is the trainable weight. The proposed SCAB can be expressed as 
\begin{equation}
\tilde{F}_{i,j} = g_{i,j}(f_{i,j}^c(F_{i,j}^c \mid \Theta_{i,j}^c) + f_{i,j}^r(F_{i,j}^r \mid \Theta_{i,j}^r) \mid \Phi_{i,j}),
\end{equation}
where $\tilde{F}_{i,j}$ is the output feature of the SCAB. Note that  SCABs endow the GDN+ with the ability to fuse features from different scales adaptively. Quite remarkably, our experimental results indicate that it suffices to use SCABs with a small number of trainable weights to substantially boost the overall performance.

%Our experimental results also validate that the performance of the proposed GDN+ can be remarkably improved by only introducing a small number of trainable weights.

\begin{figure}[t]
	\centering
	\includegraphics[width=\linewidth]{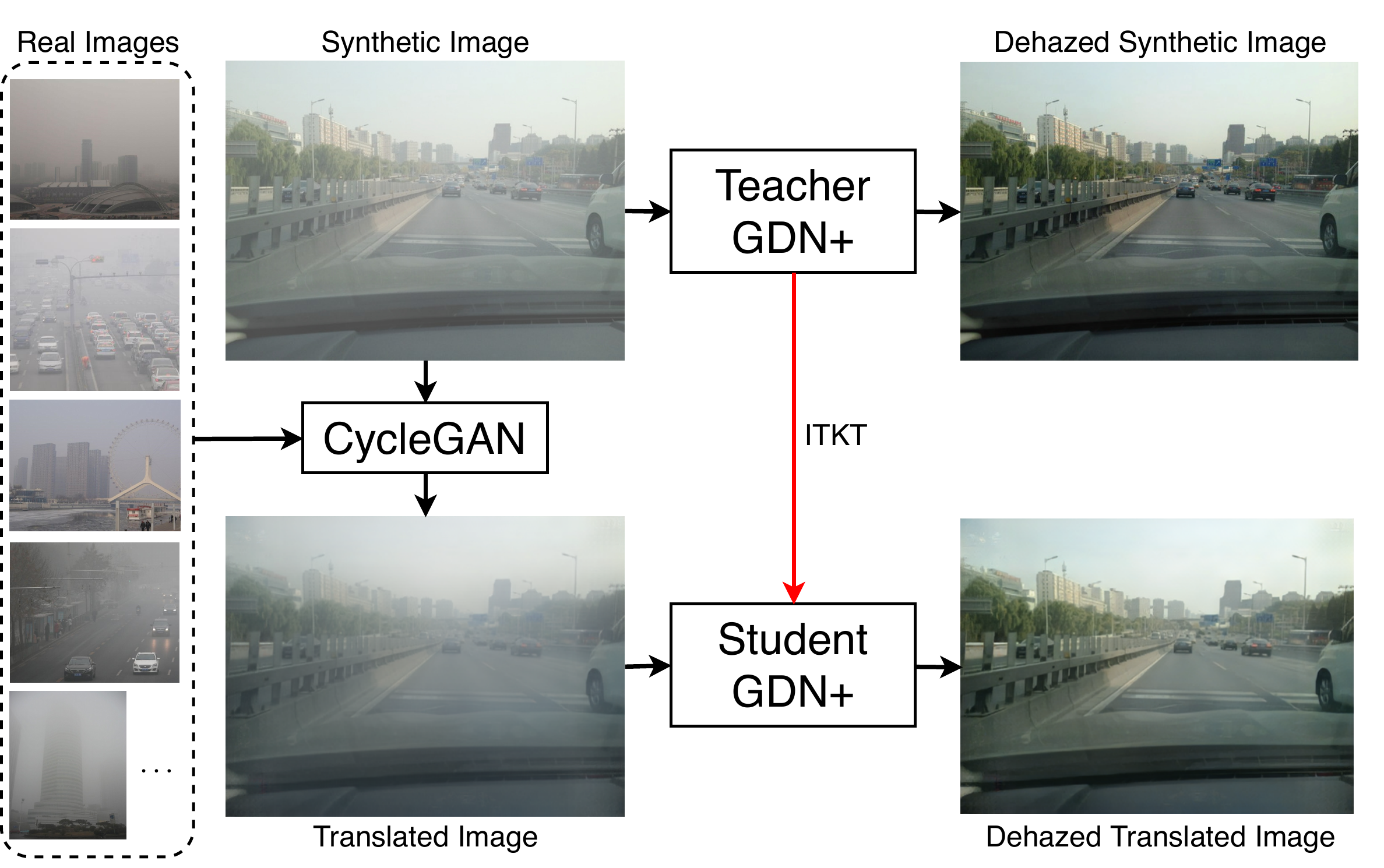}
	\caption{The flowchart of the proposed ITKT mechanism.}
	\label{fig:KT}
\end{figure}

\subsection{Intra-Task Knowledge Transfer}

We use the CycleGAN~\cite{zhu2017unpaired} to convert ASM-based synthetic data to more realistic-looking translated data, which can be regarded as samples from the distribution of real-world hazy images. As the real haze effect captured by translated data does not admit a simple mathematical characterization, the learning process on translated data is more difficult than that on synthetic data. Therefore, to memorize and take advantage of synthetic domain knowledge, we propose an ITKT mechanism to reduce the finetuning difficulty on translated data. The overall flowchart of the proposed ITKT mechanism is demonstrated in~Fig.~\ref{fig:KT}. the teacher GDN+ is pre-trained on synthetic data, and its learned weights are utilized to initialize the student GDN+. During the finetuning process, the teacher GDN+ is responsible for memorizing and providing the synthetic domain knowledge to the student GDN+, thus its weights are fixed. The student GDN+, equipped with this knowledge, is finetuned on translated data in a supervised manner to improve the dehazing performance on real-world hazy images. Note that the teacher and student networks have the freedom to adopt their own architectures as long as the synthetic domain knowledge is properly transferred.

As shown in Fig.~\ref{fig:KT} and Fig.~\ref{fig:dataset}, the haze effect of synthetic images is noticeably different from that of translated ones, which is a clear indicator of domain shift. Benefiting from ITKT, the performance drop on real-world hazy images is significantly alleviated.
In Sec.~\ref{sec:ITKT}, we also evaluate the effectiveness of ITKT by directly finetuning the GDN+ on translated data. Our experimental results show that the dehazing performance deteriorates as a consequence of this change.

%-------------------------------------------------------------------------
\subsection{Loss Function} \label{3.4}

In total, three different loss functions are employed to train the proposed network: 1) the fidelity loss $L_F$, 2) the perceptual loss $L_P$, and 3) the intra-task knowledge transfer loss $L_{KT}$. Their definitions and the underlying rationale are detailed below.

\subsubsection{Fidelity Loss} The commonly used fidelity losses include $L_1$ and MSE. The MSE loss is very sensitive to outliers, thus might suffer from gradient explosion~\cite{Girshick_2015_ICCV}. Although the $L_1$ loss does not have this issue, it is not differentiable at zero. The smooth $L_1$ loss can be regarded as an integration of these two losses, thus inherits their merits and avoids their drawbacks. Therefore, we use it as our fidelity loss to quantitatively measure  the difference between the dehazed image and the ground-truth.

Let $\hat{J}_c(x)$ denote the intensity of the $c$th color channel of the pixel $x$ in the dehazed image, and $N$ denote the total number of pixels in one channel. Our fidelity loss can be expressed as
\begin{equation}
	L_{F} = \frac{1}{3N}\sum_{c=1}^3\sum_{x=1}^N h(\hat{J}_c(x)-J_c(x)),
\end{equation}
where 
\begin{eqnarray}h(e)=
	\begin{cases}
		0.5e^2, &\mbox{if }|e|<1,\cr |e|-0.5, &\mbox{otherwise}. \cr\end{cases}
\end{eqnarray}

\subsubsection{Perceptual Loss} 
As a complement to the pixel-level fidelity loss, the perceptual loss~\cite{johnson2016perceptual} leverages multi-scale features extracted from a  pre-trained deep neural network to quantify the overall \textit{perceptual} difference between the dehazed image and the ground-truth. In this work, we use the  VGG16~\cite{simonyan2014very} pre-trained on ImageNet~\cite{russakovsky2015imagenet} as our loss network and extract the features from the last layer of each of the first three stages ({\textit{i}.\textit{e}.,  Conv1-2, Conv2-2 and Conv3-3). The perceptual loss can be defined as 
\begin{equation}
	L_{P} = \frac{1}{3}\sum_{l=1}^3\frac{1}{C_lH_lW_l}||\phi_l(\hat{J})-\phi_l(J)||_2^2,
\end{equation}
where $\phi_l(\hat{J})$ ($\phi_l(J)$), $l=1,2,3$, denote the aforementioned three VGG16 feature maps associated with the dehazed image  $\hat{J}$ (the ground truth $J$), and $C_l$, $H_l$, and $W_l$ specify the dimension of $\phi_l(\hat{J})$ ($\phi_l(J)$).

\subsubsection{Intra-Task Knowledge Transfer Loss} 
To effectively transfer the synthetic domain knowledge, we design an ITKT loss that guides the features from the student network to mimic the ones from the teacher network by reducing their $L_1$ distance. Three intermediate features from the first scale of the backbone module after the SCAB-based fusion are selected.
 According to our experiments, this selection induces the best dehazing performance among the candidates that have been considered. Following the notation in Sec.~\ref{3.3}, we denote these features by $\tilde{F}_{0,3}$, $\tilde{F}_{0,4}$,  $\tilde{F}_{0,5}$, and  use the superscripts $t$ and $s$ to indicate whether they come from the teacher or student network. Our ITKT loss can be expressed as

\begin{equation}
L_{KT} = \frac{1}{3}\sum_{j=3}^{5}||\tilde{F}_{0,j}^s - \tilde{F}_{0,j}^t||_1.
\end{equation}
	
\subsubsection{The Overall Loss} The overall loss $L_S$ of our GDN+ is a linear combination of fidelity loss $L_F$, perceptual loss $L_P$, and  ITKT loss $L_{KL}$, which can be formulated as
\begin{equation}
L_S = L_F + \lambda_P L_P + \lambda_{KL} L_{KL},
\end{equation}
where $\lambda_P$ and $\lambda_{KL}$ are used to balance the loss components. According to our experiments, they are set to $0.04$ and $0.01$ respectively.

%\subsubsection{The Overall Losses for Teacher and Student Networks} The overall loss for the teacher network (denoted as $L_T$) is a linear combination of fidelity and perceptual losses. It can be formulated as
%\begin{equation}
%L_T = L_F + \lambda_P L_P,
%\end{equation}
%where $\lambda_P$ is used to balance the two loss components, and is set to $0.04$. As for the student network, its loss function (denoted as $L_S$) integrates 
%fidelity and perceptual losses as well as the ITKT loss.
%Specifically, we have
%\begin{equation}
%L_S = L_F + \lambda_P L_P + \lambda_{KL} L_{KL},
%\end{equation}
%where $\lambda_{KL}$ plays a similar role as $\lambda_P$,  and is set to $0.01$.

\section{Data Preparation}

\begin{figure}[t]
	\centering
	\begin{minipage}[h]{0.24\linewidth}
		\centering
		\includegraphics[width=\linewidth]{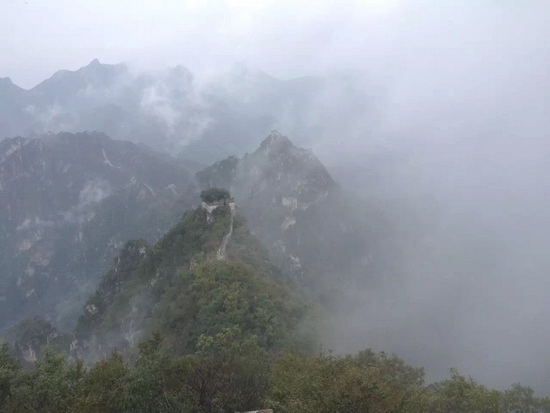}
	\end{minipage}
	\begin{minipage}[h]{0.24\linewidth}
		\centering
		\includegraphics[width=\linewidth]{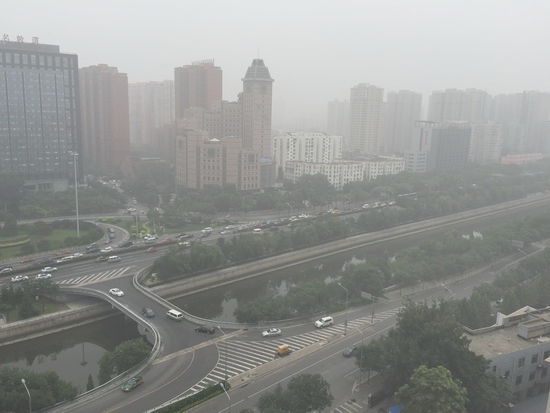}
	\end{minipage}
	\begin{minipage}[h]{0.24\linewidth}
		\centering
		\includegraphics[width=\linewidth]{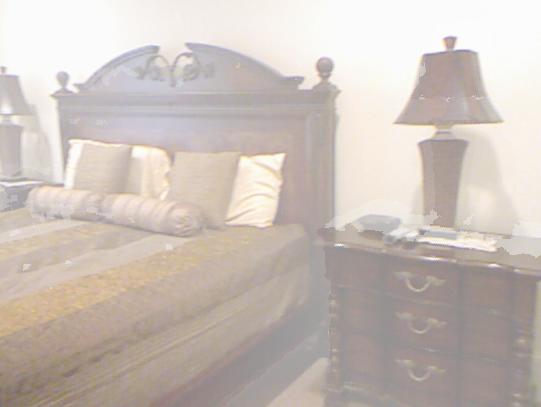}
	\end{minipage}
	\begin{minipage}[h]{0.24\linewidth}
		\centering
		\includegraphics[width=\linewidth]{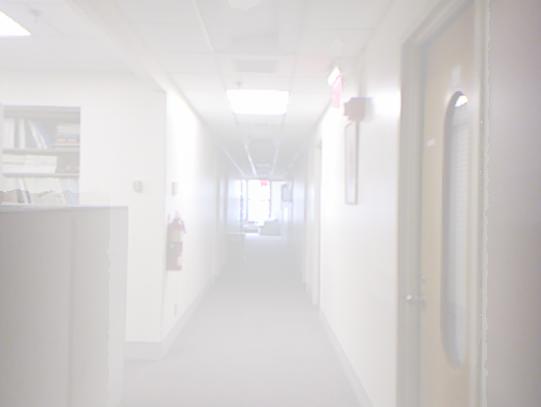}
	\end{minipage}
	
	\vspace{1mm}
	
	\begin{minipage}[h]{0.24\linewidth}
		\centering
		\includegraphics[width=\linewidth]{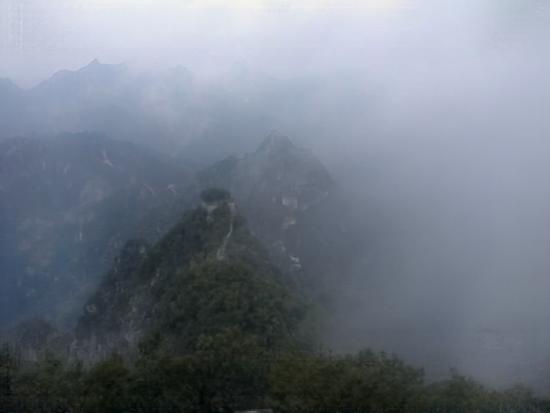}
	\end{minipage}
	\begin{minipage}[h]{0.24\linewidth}
		\centering
		\includegraphics[width=\linewidth]{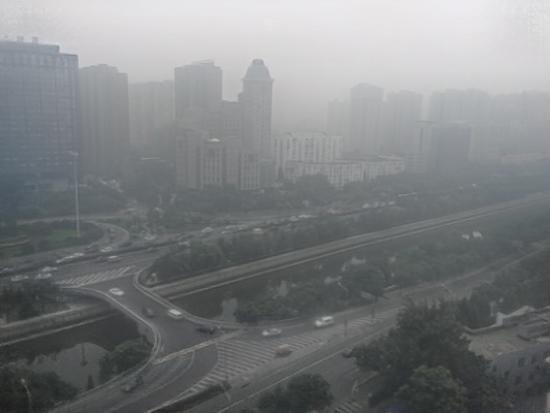}
	\end{minipage}
	\begin{minipage}[h]{0.24\linewidth}
		\centering
		\includegraphics[width=\linewidth]{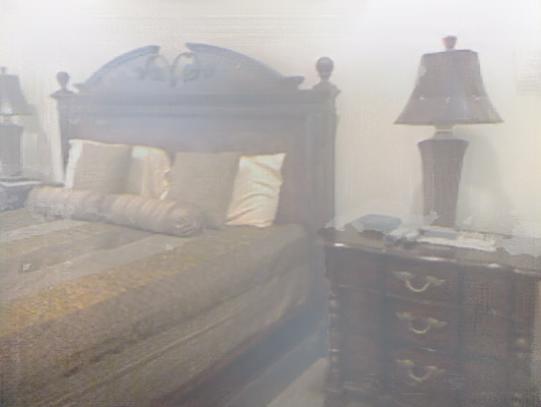}
	\end{minipage}
	\begin{minipage}[h]{0.24\linewidth}
		\centering
		\includegraphics[width=\linewidth]{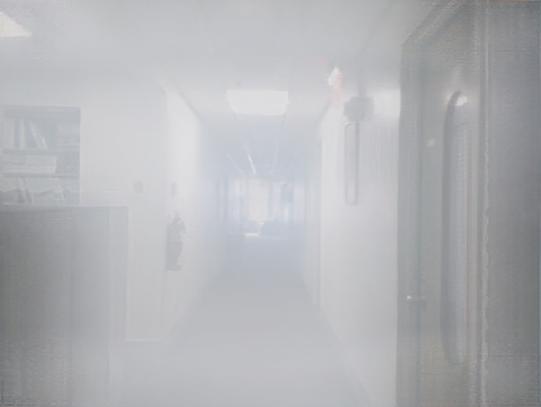}
	\end{minipage}
	\caption{Visualization of the haze effect before (shown in the $1$st row) and after (shown in the $2$rd row) the translation.}
	\label{fig:dataset}
	%	\vspace{-1.5em}
\end{figure}

\begin{table*}
	\centering
	\small
	\caption{\label{tab:syn1}Quantitative Evaluations on Four Dehazing Benchmarks. Average PSNR/SSIM Values Are Reported. \textbf{\red{Red}} and {\blue{Blue}} indicate the best and the second best performance, respectively.}
	\begin{tabular}{cc|cc|cc|cc|cc|cc}
		\hline
		\multicolumn{2}{c|}{\multirow{2}*{Method}}
		& \multicolumn{2}{c|}{SOTS}
		& \multicolumn{2}{c|}{Middlebury}
		& \multicolumn{2}{c|}{HazeRD}
		& \multicolumn{2}{c|}{O-HAZE}
		& \multicolumn{2}{c}{\multirow{2}*{$\#$ Param.(M)}} \\
		\cline{3-8}
		\multicolumn{2}{c|}{~} & PSNR & SSIM & PSNR & SSIM & PSNR & SSIM& PSNR & SSIM & \multicolumn{2}{c}{~} \\
		\hline
		%----------above are top 2 rows -------%
		\multicolumn{2}{c|}{DCP~\cite{iebmdcp01}}& $15.49$ &$0.64$ & $15.91$ & $0.81$ & $14.01$ & $0.39$ & $15.80$ & $0.33$ &\multicolumn{2}{c}{-} \\
		
		%---------- MSCNN-------%
		\multicolumn{2}{c|}{MSCNN~\cite{iebmmscnn01}}& $17.57$ &$0.81$& $13.35$ &$0.78$ & $15.57$ &$0.42$ & $19.60$&$0.63$&\multicolumn{2}{c}{$0.008$} \\
		
		%---------- DehazeNet-------%
		\multicolumn{2}{c|}{DehazeNet~\cite{iebmdehazenet01}}& $21.14$ & $0.85$ & $13.28$ & $0.79$ & $15.54$ & $0.41$ & $19.37$ & $0.64$ &\multicolumn{2}{c}{$0.008$}\\
		
		%---------- AOD-Net-------%
		\multicolumn{2}{c|}{AOD-Net~\cite{iebmaod01}}& $19.06$ & $0.85$ &  $13.86$ & $0.79$ & $15.63$ & $0.45$ & $18.85$ & $0.70$ &\multicolumn{2}{c}{\textbf{$0.002$}} \\
		
		%---------- GFN-------%
		\multicolumn{2}{c|}{GFN~\cite{iebmgated01}}& $22.30$ & $0.88$ & $14.38$ & $0.81$ & $13.98$ & $0.37$ & $22.83$ & $0.77$ &\multicolumn{2}{c}{$0.212$} \\
		
		%---------- EPDN-------%
		\multicolumn{2}{c|}{EPDN~\cite{qu2019enhanced}}& $23.82$ & $0.89$ & $15.11$ & $0.83$ & $17.37$ & $0.56$ & $19.82$ & $0.75$ &\multicolumn{2}{c}{$17.380$} \\

		%---------- KDDN-------%
		\multicolumn{2}{c|}{KDDN~\cite{hong2020distilling}}
		& {$30.09$}&{\underline{\textcolor{blue}{$0.97$}}} &  {$17.27$}&\underline{\textcolor{blue}{$0.87$}} &
		$16.44$ &\underline{\textcolor{blue}{$0.83$}}& 
		$25.45$&$0.78$&\multicolumn{2}{c}{$2.405$} \\
		
		%---------- DADN-------%
		\multicolumn{2}{c|}{DADN~\cite{shao2020domain}}& $27.76$ &$0.93$ & $15.93$ &$0.70$ &\underline{\textcolor{blue}{$18.07$}} & $0.63$ & $21.69$ & $0.74$ &\multicolumn{2}{c}{$54.591$} \\
		
		%---------- ACER-------%
		\multicolumn{2}{c|}{ACER-Net~\cite{wu2021contrastive}}& \textbf{\underline{\textcolor{blue}{$31.61$}}}&\textcolor{red}{$\bf{0.98}$}
		& \underline{\textcolor{blue}{$17.48$}} & $0.86$
		& $15.88$ & $0.81$ 
		& \underline{\textcolor{blue}{$25.87$}} & \underline{\textcolor{blue}{$0.84$}} &
		\multicolumn{2}{c}{$2.614$} \\

		%---------- GDN-------%
		\multicolumn{2}{c|}{GDN~\cite{liu2019griddehazenet}}& {$27.75$}&$0.96$ & {$14.46$}&{$0.83$} & $14.51$&$0.79$ & $23.51$&$0.83$&\multicolumn{2}{c}{$0.958$} \\
		
		\hline
		%---------- Ours-------%
		\multicolumn{2}{c|}{GDN+}&{\textcolor{red}{$\bf{32.15}$}}&{\textcolor{red}{$\bf{0.98}$}} & \textcolor{red}{$\bf{18.04}$}&\textcolor{red}{$\bf{0.88}$} & \textcolor{red}{$\bf{19.54}$}&\textcolor{red}{$\bf{0.87}$} & \textcolor{red}{$\bf{26.10}$}&\textcolor{red}{$\bf{0.85}$}&\multicolumn{2}{c}{$0.961$} \\
		\hline
	\end{tabular}
\end{table*}

\subsection{Training Dataset}
%\XH{A subset of the large-scale synthetic dataset, named RESIDE~\cite{li2019benchmarking}, is used to train the GDN+. The RESIDE dataset contains a Indoor Training Set (ITS) and a Outdoor Training Set (OTS) that are synthesized from clear indoor and outdoor images based on the ASM via proper choices of the scattering coefficient $\beta$ and the atmospheric light intensity $A$.} The ITS has a total of $13,990$ hazy images, generated from $1,399$ clear images with $\beta \in [0.6, 1.8]$ and $A \in [0.7, 1.0]$ (the depth maps $d(x)$ are obtained from the NYU Depth V2 dataset~\cite{silberman2012indoor} and Middlebury Stereo dataset~\cite{scharstein2003high}). We observe that there is an overlap between the OTS and the Synthetic Objective Testing Set (SOTS) in the sense that some clear images are used to generate hazy versions in both sets (though the adopted ASM parameters are different). With the overlapped part removed, the OTS has a total of $296,695$ hazy images, generated from $8,477$ clear images with $\beta \in [0.04, 0.2]$ and $A \in [0.8, 1.0]$ (the depth maps are estimated using the algorithm in~\cite{liu2016learning}). 

The RESIDE~\cite{li2019benchmarking} is a large-scale dataset that contains an Indoor Training Set (ITS), an Outdoor Training Set (OTS), a Synthetic Object Testing Set (SOTS), a set of Unannotated real Hazy Images (URHI), and a real Task-driven Testing Set (RTTS). The ITS and OTS are generated from clear images based on the ASM via proper choices of the scattering coefficient $\beta$ and the atmospheric light intensity $A$. Following DADN~\cite{shao2020domain}, we use the exactly same dataset that consists of $6,000$ images with $3,000$ from ITS and $3,000$ from OTS to train our GDN+. Since different dehazing methods may originally adopt different training datasets (\textit{e.g.}, AOD-Net~\cite{iebmaod01} was trained  using $27,256$ synthetic hazy images while ACER-Net~\cite{wu2021contrastive} was trained on ITS that only has $13,990$ images), for fair comparisons, we laboriously retrain all the methods under consideration on the aforementioned dataset by following their respective training strategies.

To finetune the GDN+, we select $1,000$ real-world hazy images from RTTS~\cite{li2019benchmarking}, and utilize the CycleGAN to convert $6,000$ synthetic images to translated ones with the distribution matched to that of real-world hazy images. Note that these $6,000$ translated images should not be considered as  additionally introduced data since they are generated from the training data \textit{per se}.} Fig.~\ref{fig:dataset} visualizes the haze effect before and after this translation.

\subsection{Testing Dataset}~\label{sec:test_dataset}
For testing, in total $6$ dehazing datasets are used. Four of them are synthetic datasets and the rest two are real datasets. These testing datasets differ in size and haze distribution. We elaborate them as follows:
\begin{itemize}
\item \textbf{SOTS}~\cite{li2019benchmarking}: Synthesized based on ASM, it comprises  $500$  indoor hazy images and $500$ outdoor hazy images roughly of size $620 \times 460$.
\item \textbf{Middlebury}~\cite{ancuti2016d}: Synthesized based on ASM, it consists of $23$ indoor hazy images roughly of size $2,880 \times 1,988$.
\item \textbf{HazeRD}~\cite{zhang2017hazerd}: Synthesized based on ASM, it contains $75$ outdoor hazy images roughly of size $3,873 \times 2,516$.
\item \textbf{O-HAZE}~\cite{ancuti2018haze}: It has $45$ outdoor hazy images roughly of size $5,456 \times 3,632$. Instead of relying on ASM for synthesizing the haze effect, the images in this dataset are produced by a professional haze machine and consequently  more visually realistic.  Since the haze distribution of this dataset is different from that of ASM-based datasets, following the testing protocol of previous dehazing works, we adopt the training/testing splits in~\cite{ancuti2018ntire} to train and test the GDN+ and other methods chosen for comparison.
\item $\mathbf{37}$\textbf{Real}~\cite{fattal2014dehazing}: It collects $37$ real-world hazy images roughly of size $768 \times 512$. This is  a commonly used benchmark for testing the performance of dehazing methods on real data.
\item \textbf{URHI}~\cite{li2019benchmarking}: It contains $4,809$ real-world hazy images of various sizes (ranging from $ 400 \times 350$ to $2,000 \times 1,000$).
\end{itemize}

Unless otherwise specified, the \textit{pre-trained} GDN+ is tested on synthetic datasets to demonstrate the superiority of our network design while the \textit{finetuned} GDN+ is tested on real datasets to verify the mitigation of domain gap attributed to ITKT. For simplicity, we do not explicitly differentiate them since it is easy to tell the difference based on the testing datasets.

\section{Experimental Results}

We conduct extensive experiments to demonstrate that the proposed GDN+ outperforms the SOTA methods on synthetic datasets and delivers visually more satisfactory results on real datasets after finetuning. The experiments also provide useful insights into the constituent modules of the GDN+ and solid justifications for the effectiveness of the proposed ITKT mechanism.
%We end this section with the runtime analysis and the discussion of a potential failure case.

%We first introduce the implementation details, and then evaluate the proposed GDN+ on several benchmarks, where the SOTA dehazing results are achieved on both synthetic and real-world datasets. Moreover, a variety of experiments are conducted to validate the advantages of our overall network design. At last, the runtime analysis and a potential failure case are provided.

\subsection{Experimental Setup}\label{sec:implementation}
The GDN+ is first trained on synthetic data for $100$ epochs and then finetuned on translated data for another $100$ epochs. %We train the teacher and student networks on synthetic and translated data separately. Since there are more than $310$k synthetic images with ITS and OTS combined, it is inefficient to train the teacher network on such a massive dataset. Therefore, we uniformly sample $6,000$ images from ITS and OTS respectively to gather a total number of $12,000$ images for one training epoch. To train the student network, we use the aforementioned $6,000$ translated hazy images. Besides, the corresponding synthetic images are fed into the teacher network to produce the features needed for ITKT. 
We randomly crop a patch of size $240 \times 240$ from each image. For training, the Adam optimizer~\cite{kingma2014adam} is adopted, where $\beta_1$ and $\beta_2$ take the default values of $0.9$ and $0.999$, respectively. The batch size is set to $16$ with the initial learning rate $1e$-$3$ that will be reduced by half every 20 epochs. The training is carried out on a PC with two NVIDIA GeForce GTX $2080$Ti, but only one GPU is used for testing.

We compare the proposed GDN+ with $10$ methods including DCP~\cite{iebmdcp01}, MSCNN~\cite{iebmmscnn01}, DehazeNet~\cite{iebmdehazenet01}, AOD-Net~\cite{iebmaod01}, GFN~\cite{iebmgated01}, EPDN~\cite{qu2019enhanced}, KDDN~\cite{hong2020distilling}, DADN~\cite{shao2020domain}, ACER-Net~\cite{wu2021contrastive}, and GDN~\cite{liu2019griddehazenet}, where the DCP is the only non-learning-based method. Although ACER-Net is the current SOTA, DADN achieves better visual quality on real-world hazy images. Therefore, we consider both of them as the representatives of existing dehazing methods. For quantitative comparisons, we leverage the Peak Signal to Noise Ratio (PSNR) and Structure Similarity Index Measure (SSIM) to evaluate the dehazing results of different methods on synthetic datasets. Since the ground-truth of real-world hazy images are not available in real datasets, the Fog Aware Density Evaluator (FADE)~\cite{choi2015referenceless}, a no-reference image quality assessment tool specifically designed for image dehazing task, is used as an alternative to support quantitative evaluations.
%Note that the official codes of some compared methods are not publicly available. For fair comparisons, we laboriously implement them by ourselves and make sure  that they can reproduce the results reported in their original papers. 

\begin{figure*}[!t]
	\centering
	% SOTS indoor
	\begin{minipage}[h]{0.118\linewidth}
		\centering
		\includegraphics[width=\linewidth]{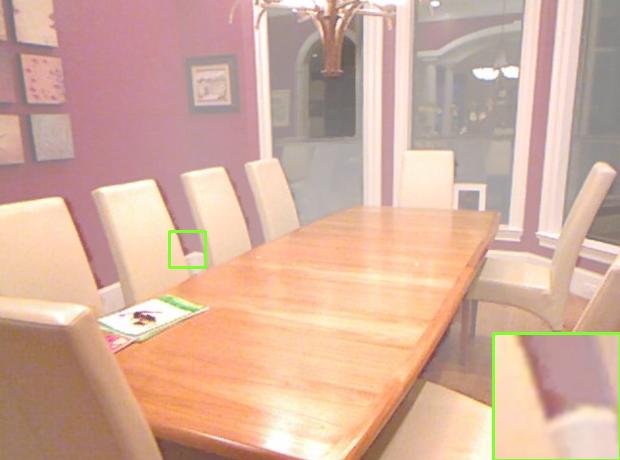}
	\end{minipage}
%	\begin{minipage}[h]{0.118\linewidth}
%		\centering
%		\includegraphics[width=\linewidth]{images/syn/SOTS/1/AOD_1400_7.jpg}
%	\end{minipage}
	\begin{minipage}[h]{0.118\linewidth}
		\centering
		\includegraphics[width=\linewidth]{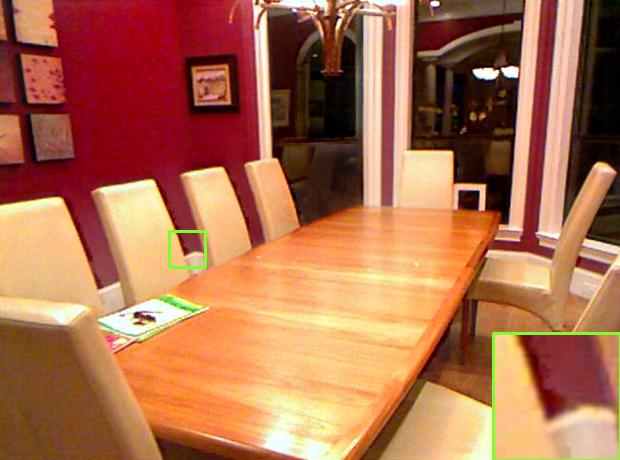}
	\end{minipage}
	\begin{minipage}[h]{0.118\linewidth}
		\centering
		\includegraphics[width=\linewidth]{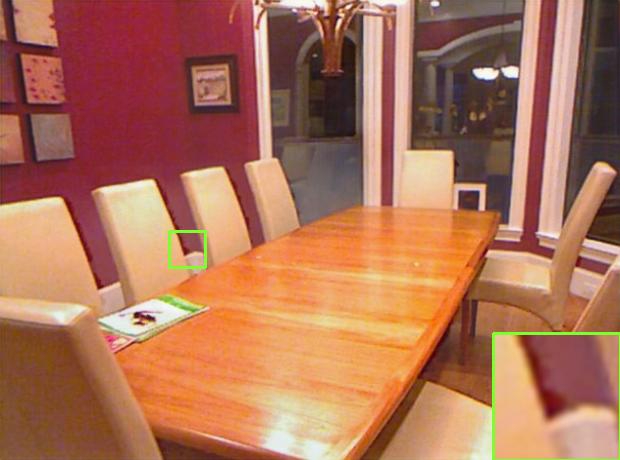}
	\end{minipage}
	\begin{minipage}[h]{0.118\linewidth}
		\centering
		\includegraphics[width=\linewidth]{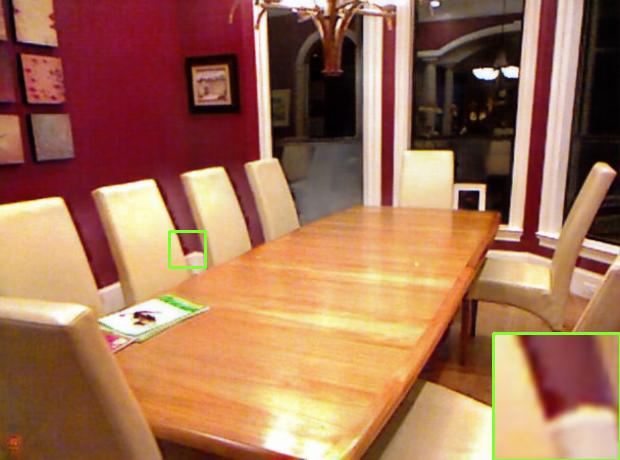}
	\end{minipage}
	\begin{minipage}[h]{0.118\linewidth}
		\centering
		\includegraphics[width=\linewidth]{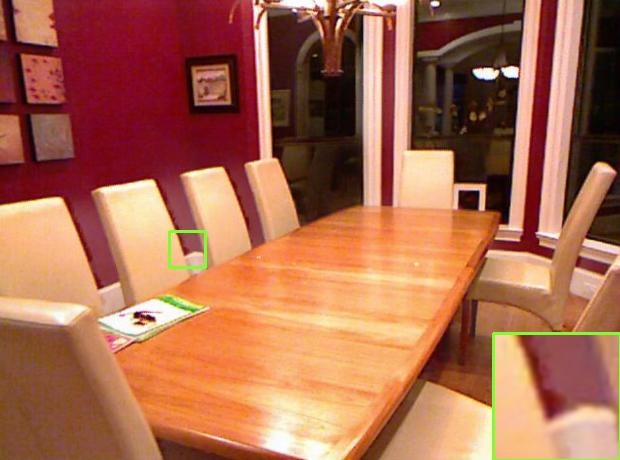}
	\end{minipage}
	\begin{minipage}[h]{0.118\linewidth}
		\centering
		\includegraphics[width=\linewidth]{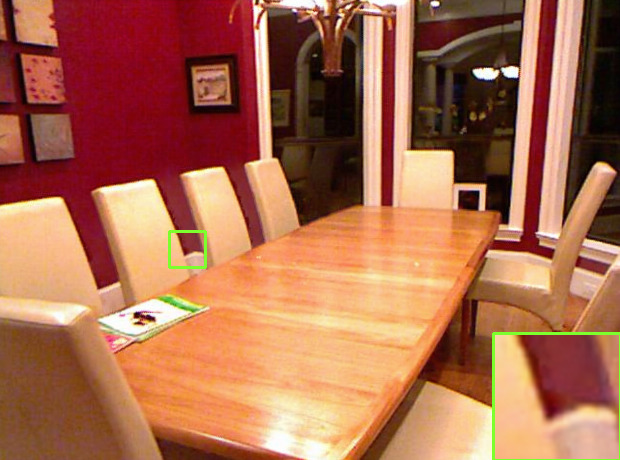}
	\end{minipage}
	\begin{minipage}[h]{0.118\linewidth}
		\centering
		\includegraphics[width=\linewidth]{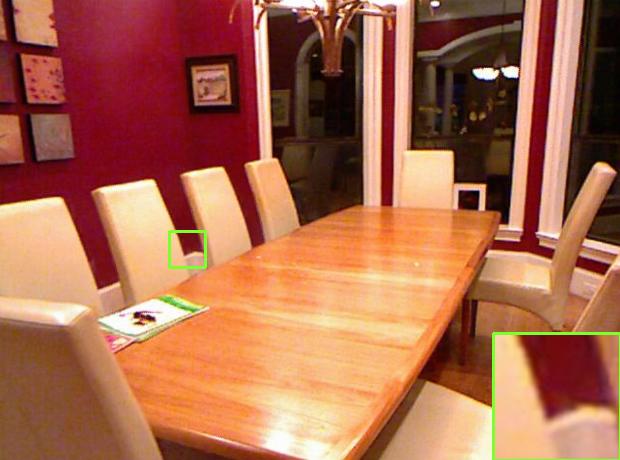}
	\end{minipage}
	\begin{minipage}[h]{0.118\linewidth}
		\centering
		\includegraphics[width=\linewidth]{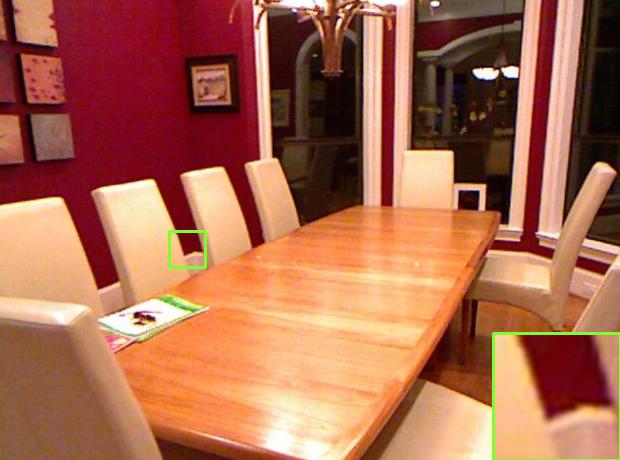}
	\end{minipage}
	\vspace{1mm}
	% SOTS outdoor
	\begin{minipage}[h]{0.118\linewidth}
		\centering
		\includegraphics[width=\linewidth]{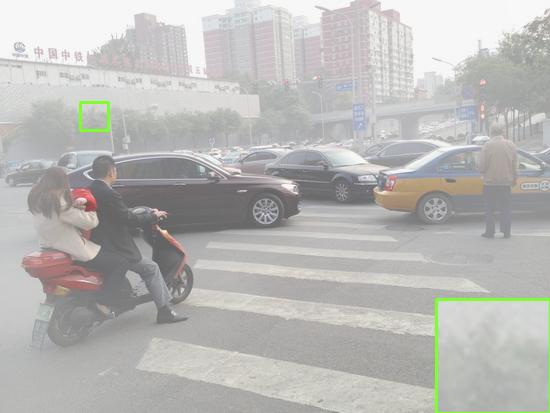}
	\end{minipage}
%	\begin{minipage}[h]{0.118\linewidth}
%		\centering
%		\includegraphics[width=\linewidth]{images/syn/SOTS/2/{AOD_0333_0.95_0.2}.jpg}
%	\end{minipage}
	\begin{minipage}[h]{0.118\linewidth}
		\centering
		\includegraphics[width=\linewidth]{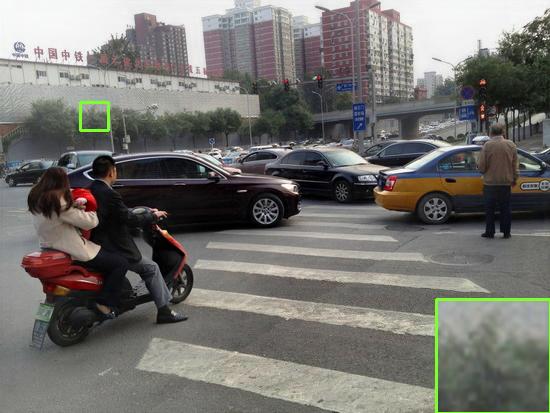}
	\end{minipage}
	\begin{minipage}[h]{0.118\linewidth}
		\centering
		\includegraphics[width=\linewidth]{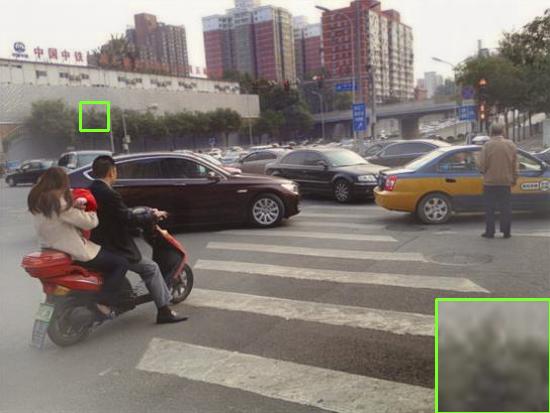}
	\end{minipage}
	\begin{minipage}[h]{0.118\linewidth}
		\centering
		\includegraphics[width=\linewidth]{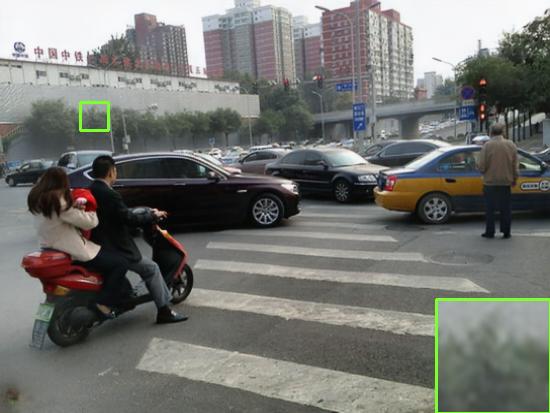}
	\end{minipage}
	\begin{minipage}[h]{0.118\linewidth}
		\centering
		\includegraphics[width=\linewidth]{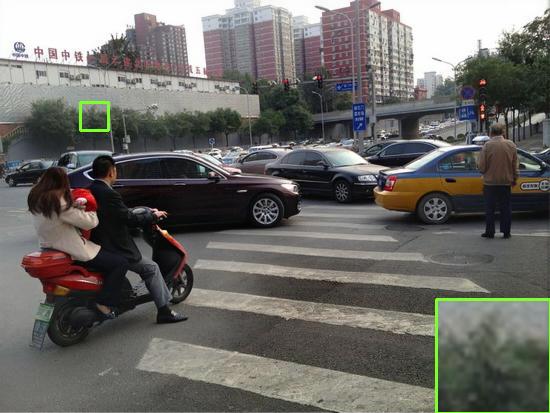}
	\end{minipage}
	\begin{minipage}[h]{0.118\linewidth}
		\centering
		\includegraphics[width=\linewidth]{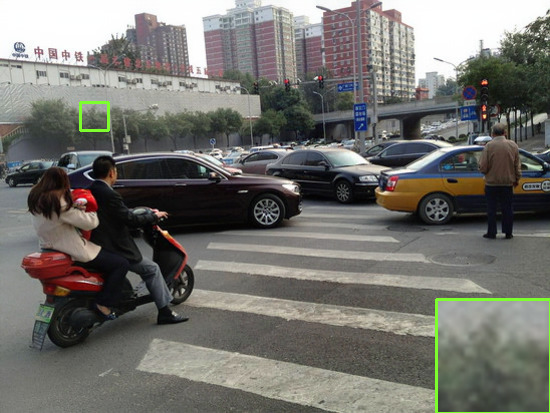}
	\end{minipage}
	\begin{minipage}[h]{0.118\linewidth}
		\centering
		\includegraphics[width=\linewidth]{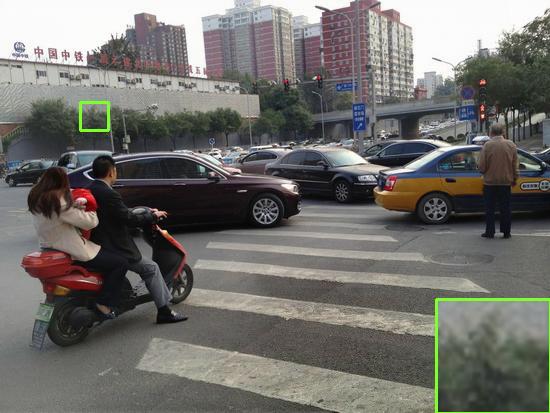}
	\end{minipage}
	\begin{minipage}[h]{0.118\linewidth}
		\centering
		\includegraphics[width=\linewidth]{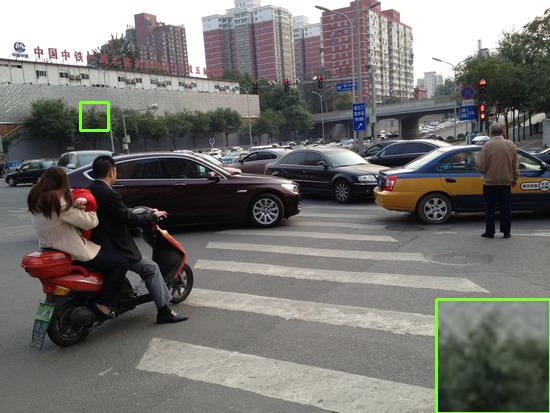}
	\end{minipage}
	\vspace{1mm}
	% Middlebury 1
	\begin{minipage}[h]{0.118\linewidth}
		\centering
		\includegraphics[width=\linewidth]{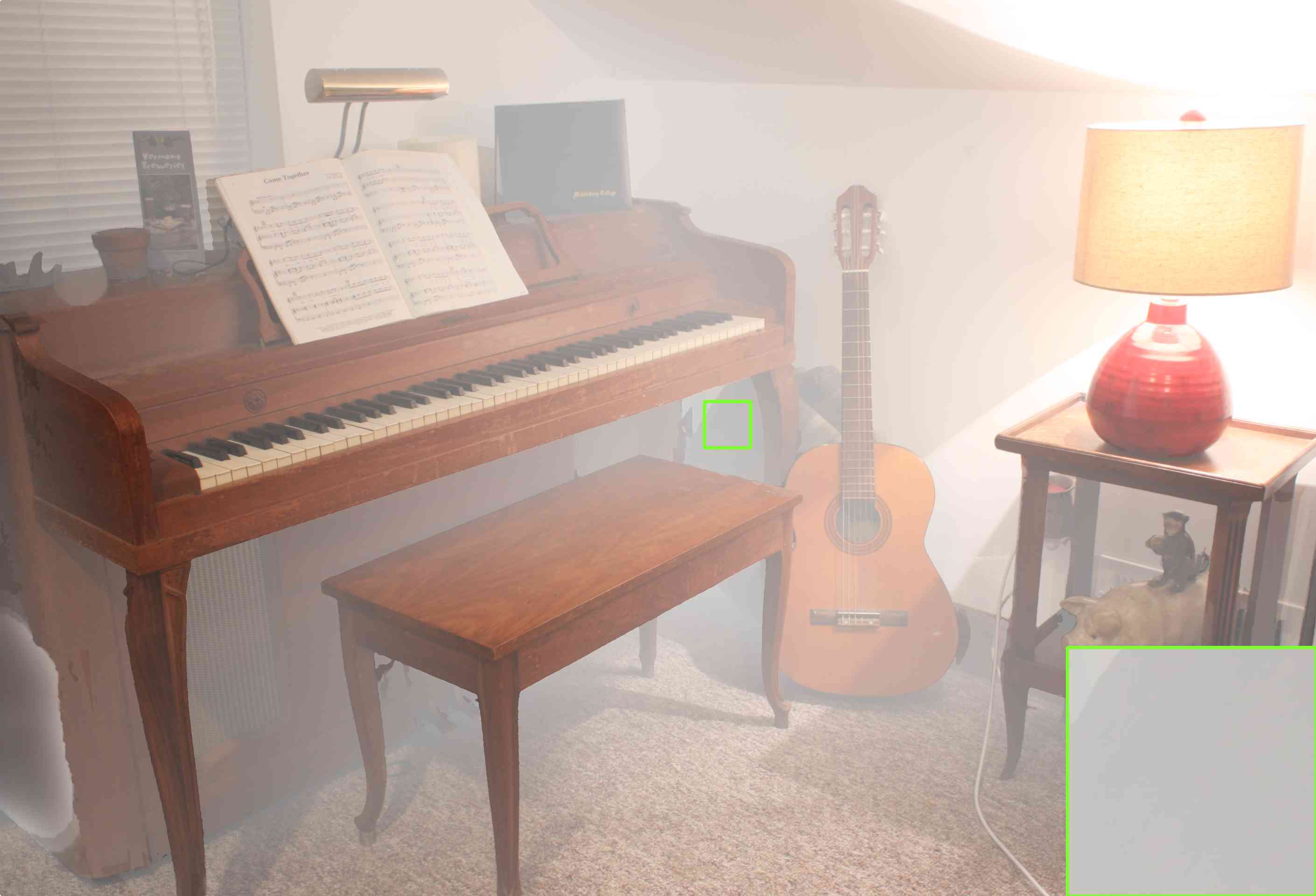}
	\end{minipage}
%	\begin{minipage}[h]{0.118\linewidth}
%		\centering
%		\includegraphics[width=\linewidth]{images/syn/dhaze/1/AOD_Piano_Hazy.jpg}
%	\end{minipage}
	\begin{minipage}[h]{0.118\linewidth}
		\centering
		\includegraphics[width=\linewidth]{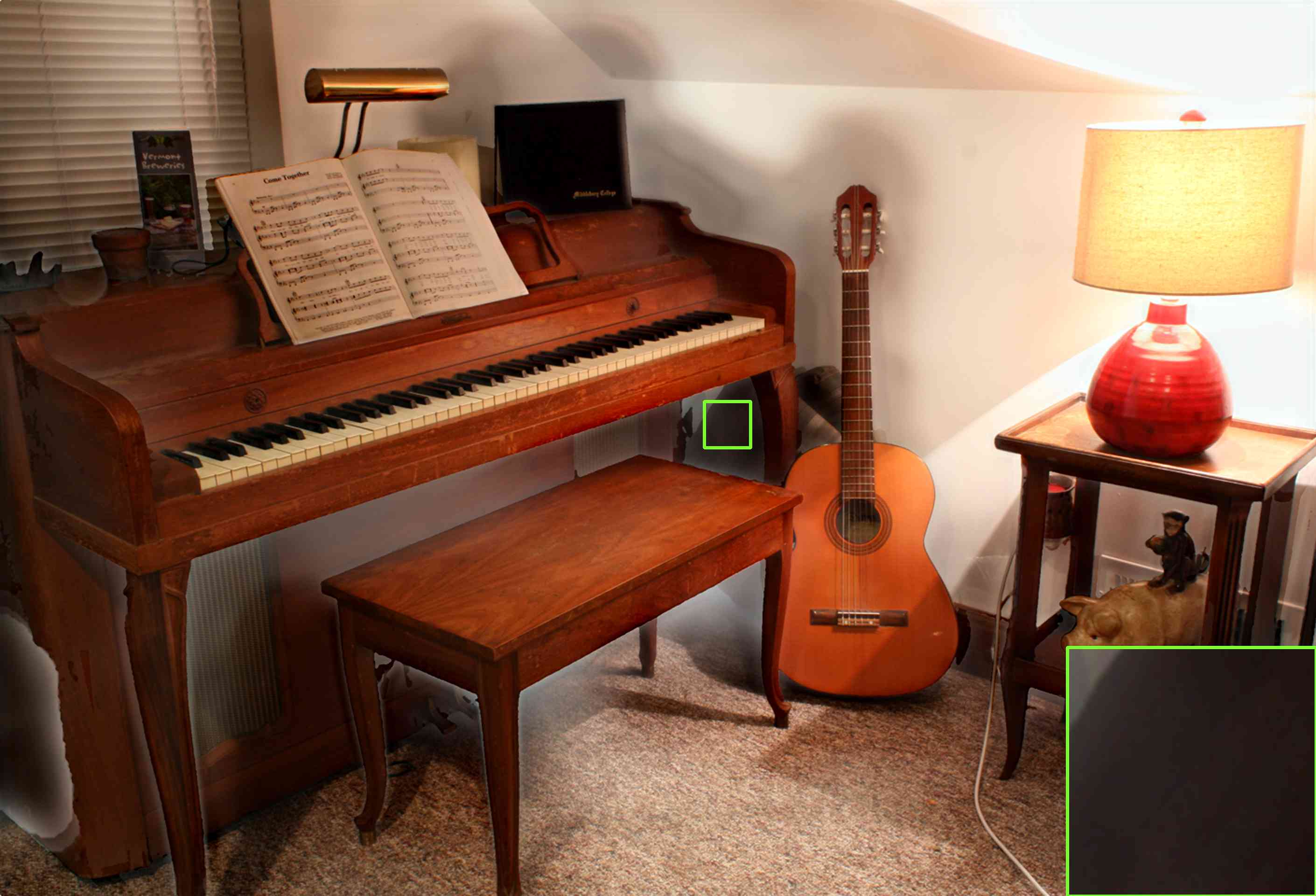}
	\end{minipage}
	\begin{minipage}[h]{0.118\linewidth}
		\centering
		\includegraphics[width=\linewidth]{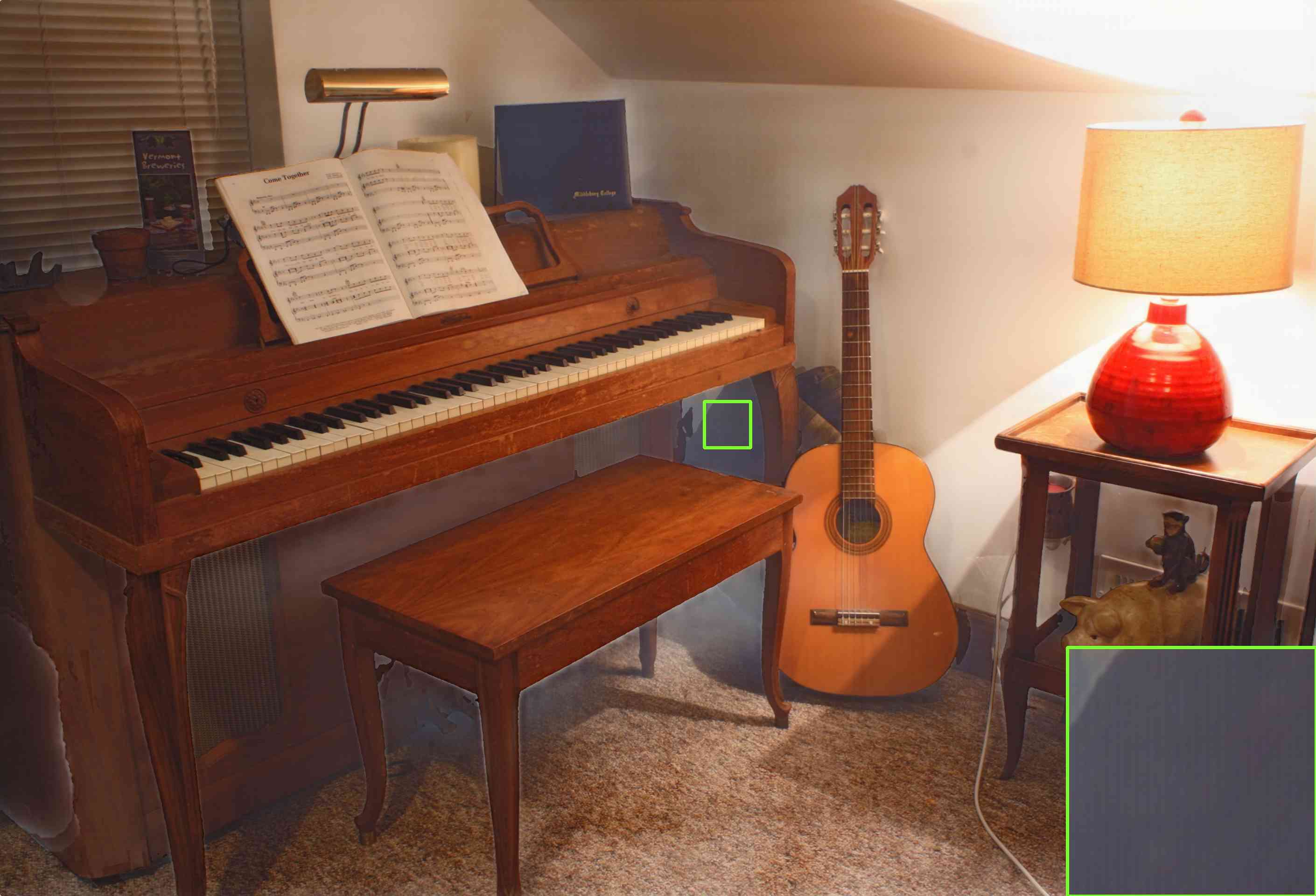}
	\end{minipage}
	\begin{minipage}[h]{0.118\linewidth}
		\centering
		\includegraphics[width=\linewidth]{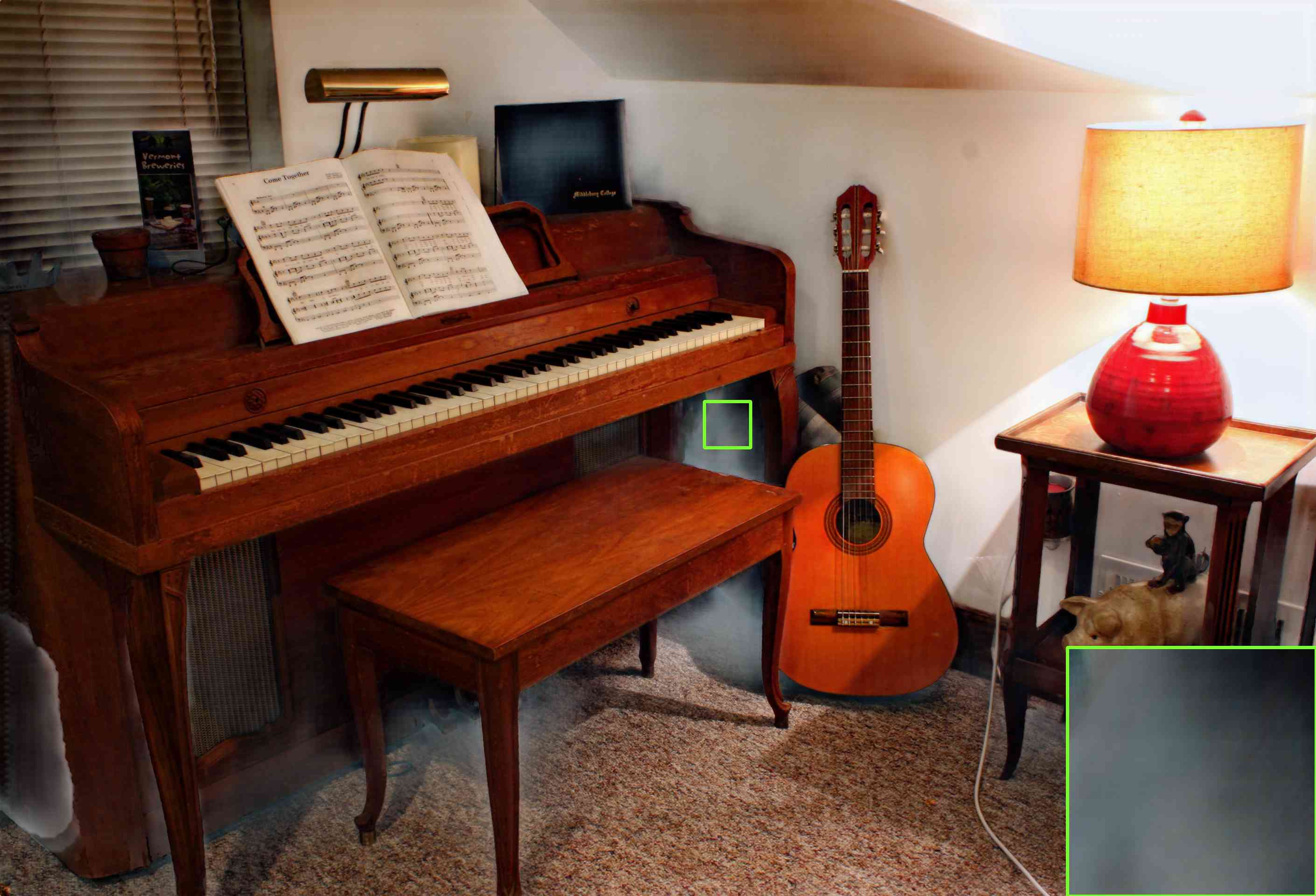}
	\end{minipage}
	\begin{minipage}[h]{0.118\linewidth}
		\centering
		\includegraphics[width=\linewidth]{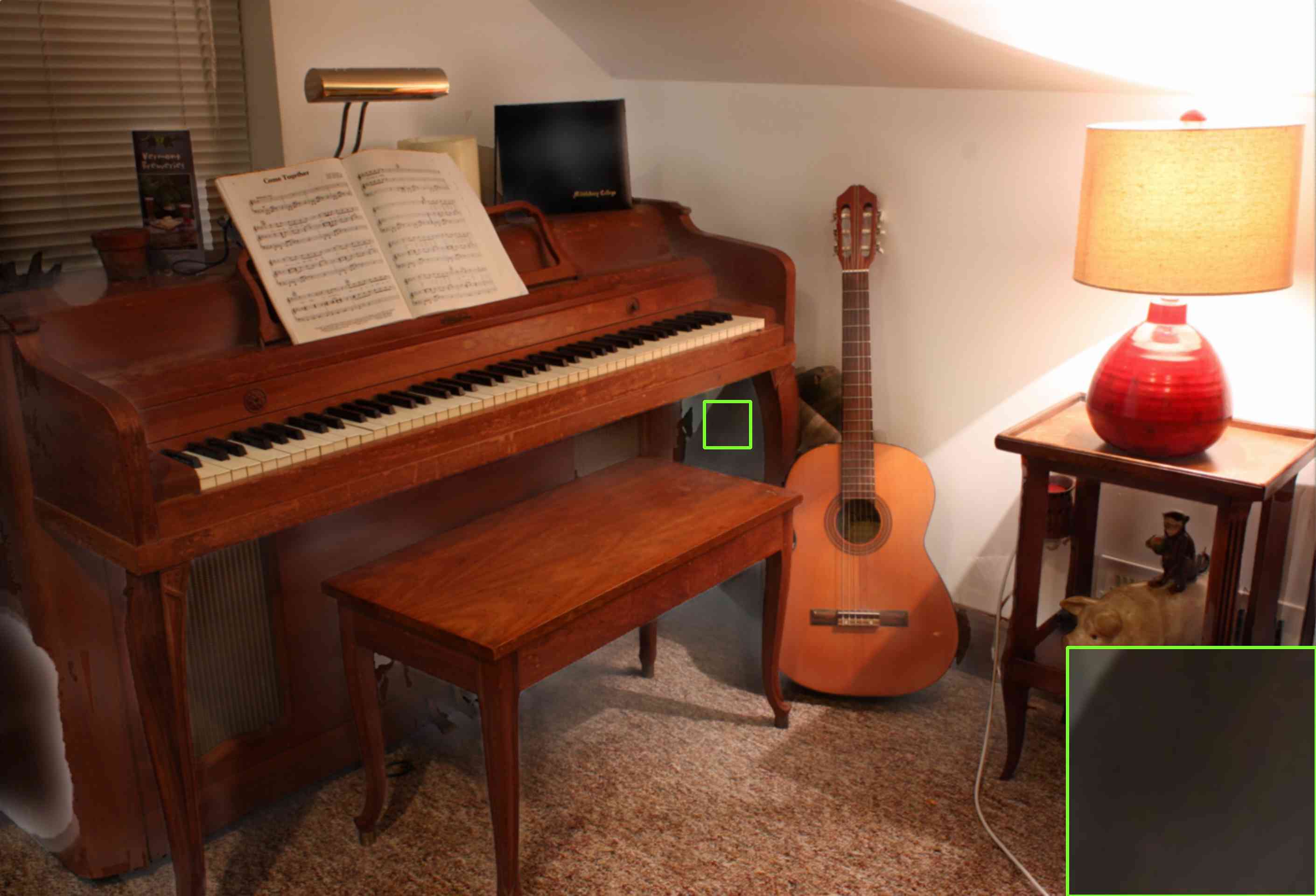}
	\end{minipage}
	\begin{minipage}[h]{0.118\linewidth}
		\centering
		\includegraphics[width=\linewidth]{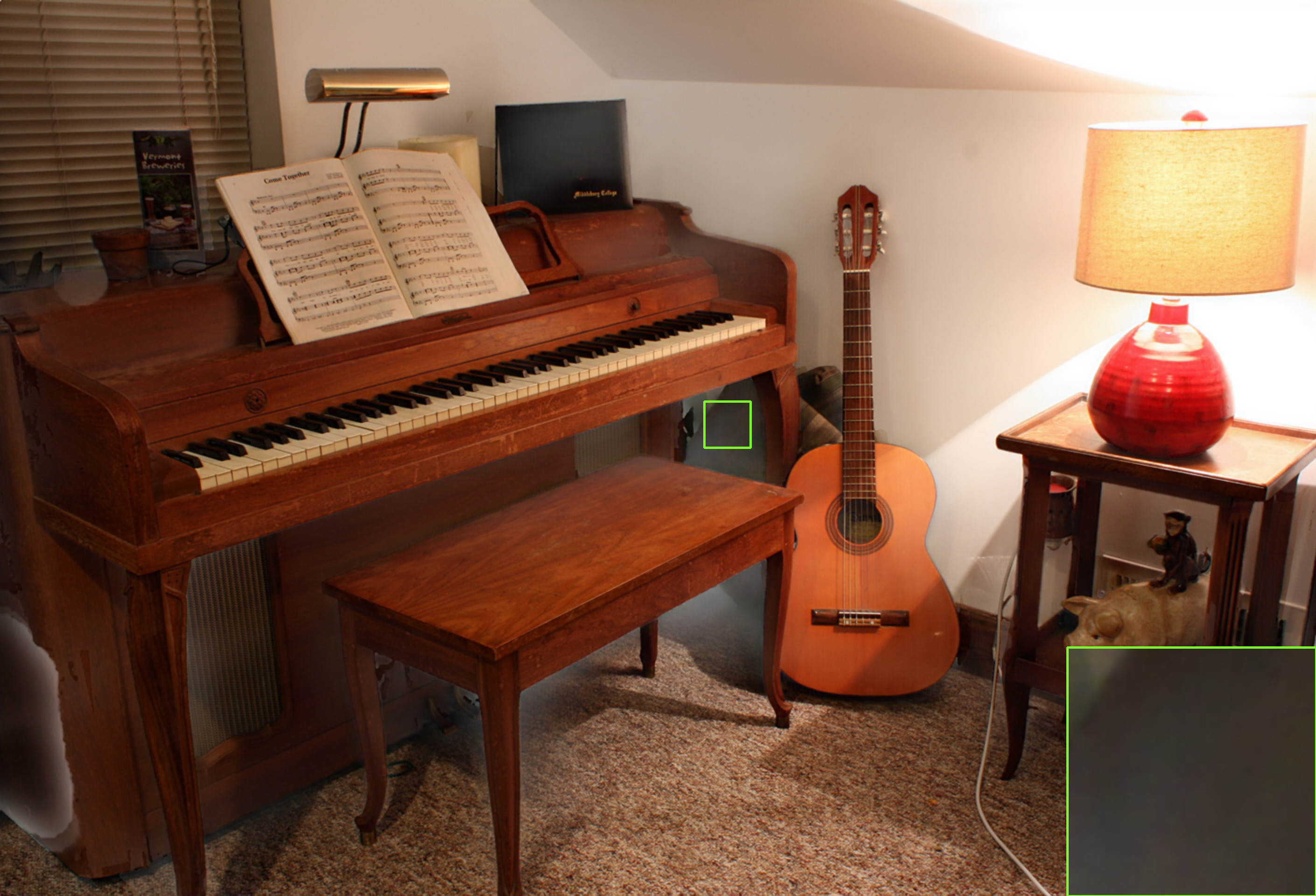}
	\end{minipage}
	\begin{minipage}[h]{0.118\linewidth}
		\centering
		\includegraphics[width=\linewidth]{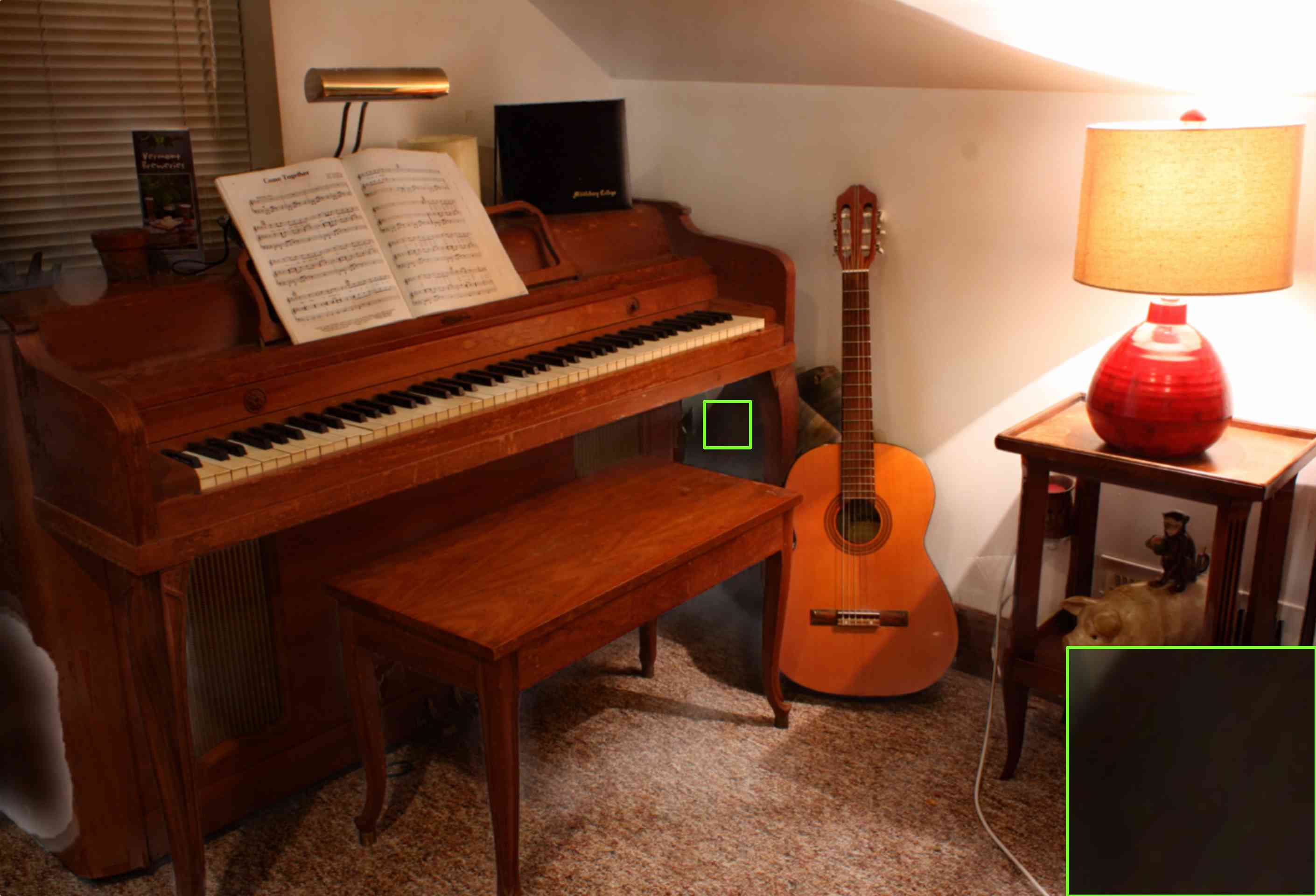}
	\end{minipage}
	\begin{minipage}[h]{0.118\linewidth}
		\centering
		\includegraphics[width=\linewidth]{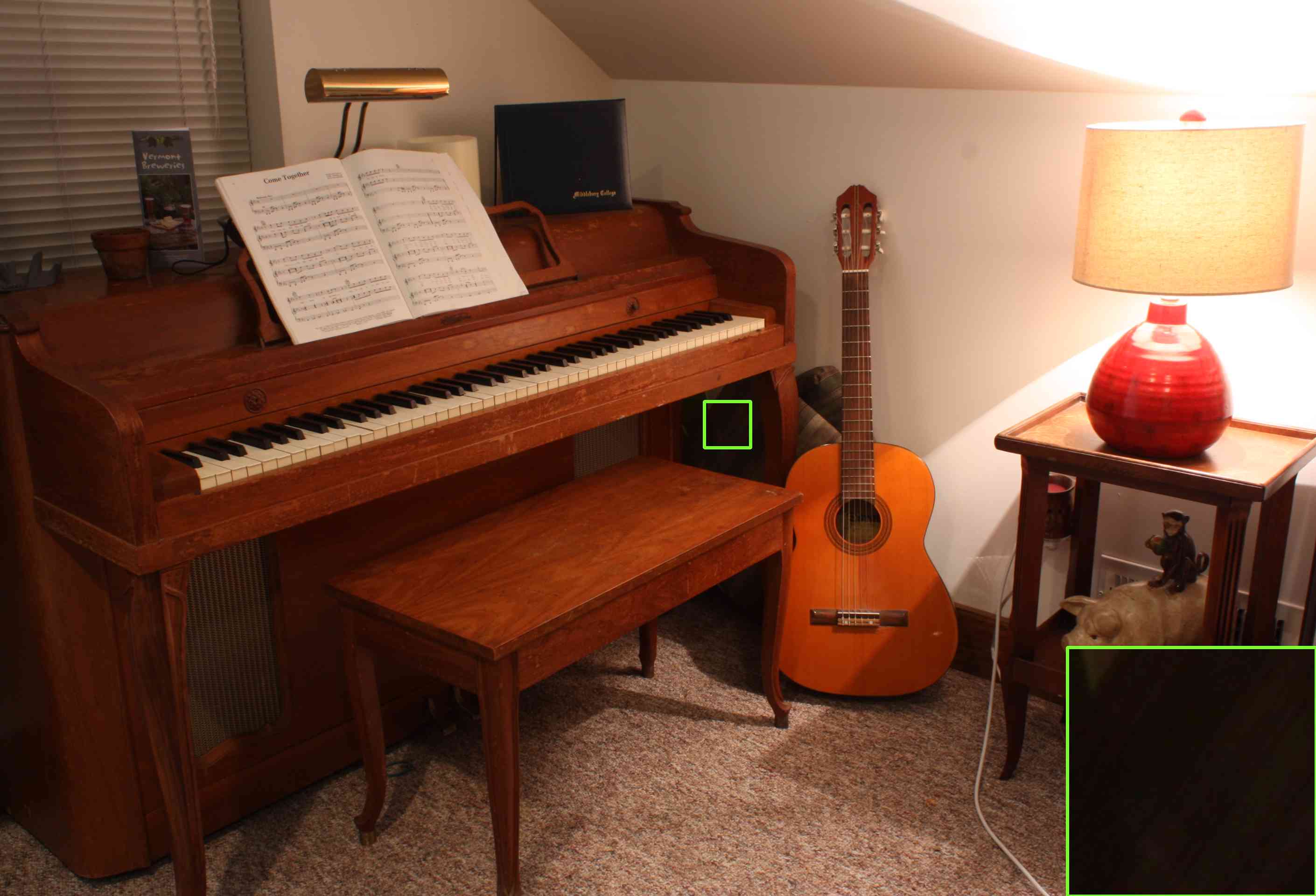}
	\end{minipage}
	\vspace{1mm}
	% Middlebury 2
	\begin{minipage}[h]{0.118\linewidth}
		\centering
		\includegraphics[width=\linewidth]{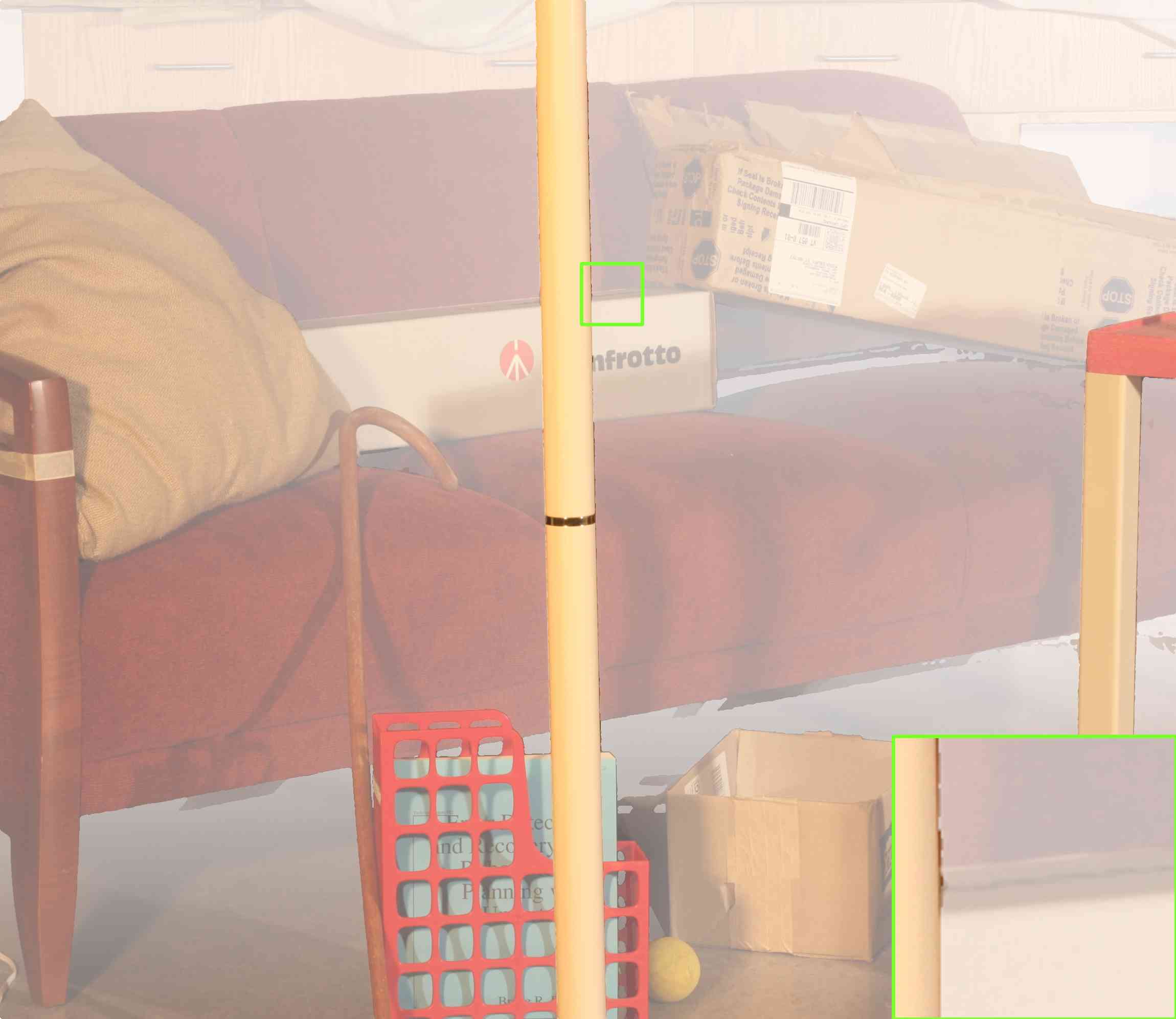}
	\end{minipage}
%	\begin{minipage}[h]{0.118\linewidth}
%		\centering
%		\includegraphics[width=\linewidth]{images/syn/dhaze/2/AOD_Couch_Hazy.jpg}
%	\end{minipage}
	\begin{minipage}[h]{0.118\linewidth}
		\centering
		\includegraphics[width=\linewidth]{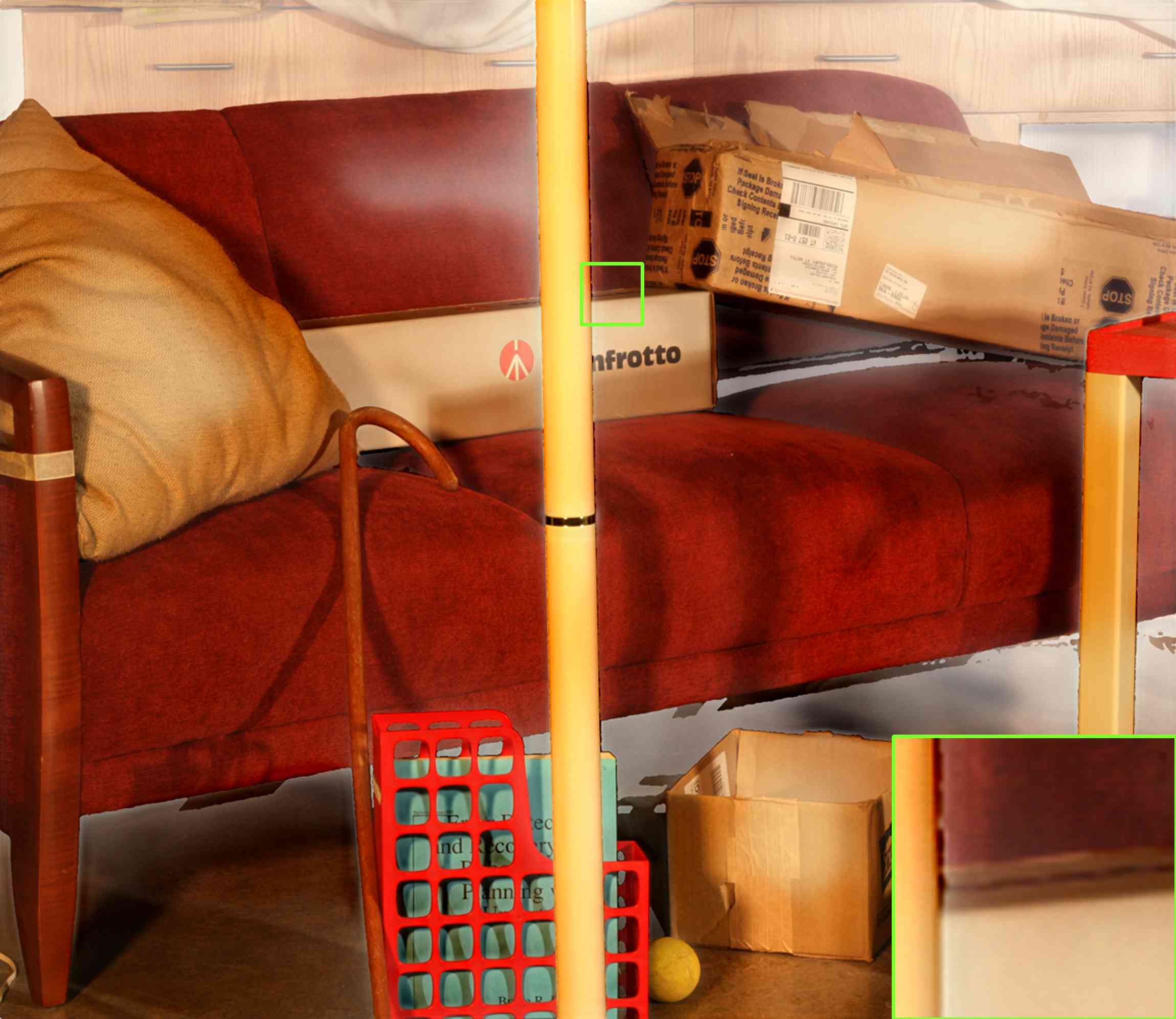}
	\end{minipage}
	\begin{minipage}[h]{0.118\linewidth}
		\centering
		\includegraphics[width=\linewidth]{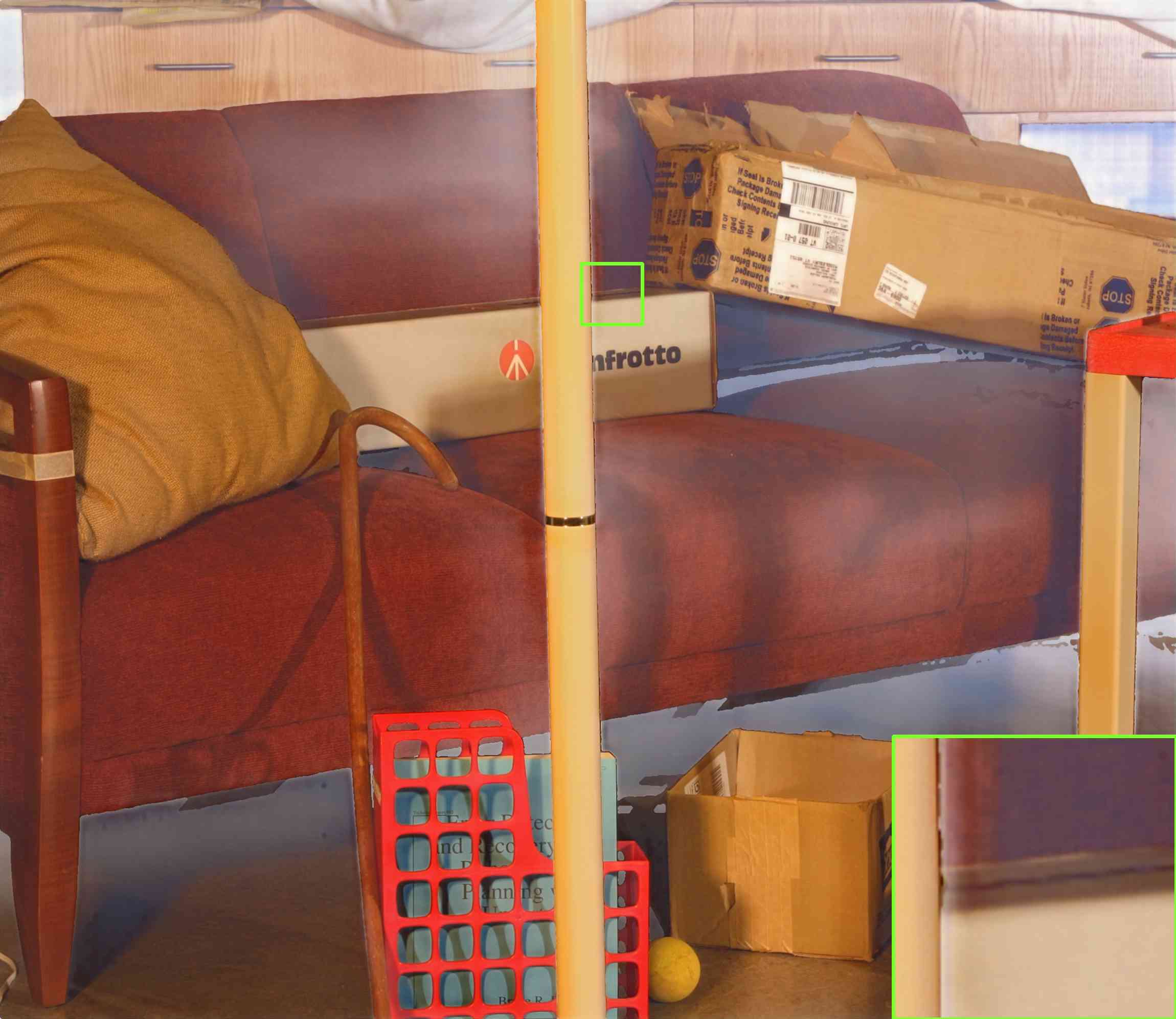}
	\end{minipage}
	\begin{minipage}[h]{0.118\linewidth}
		\centering
		\includegraphics[width=\linewidth]{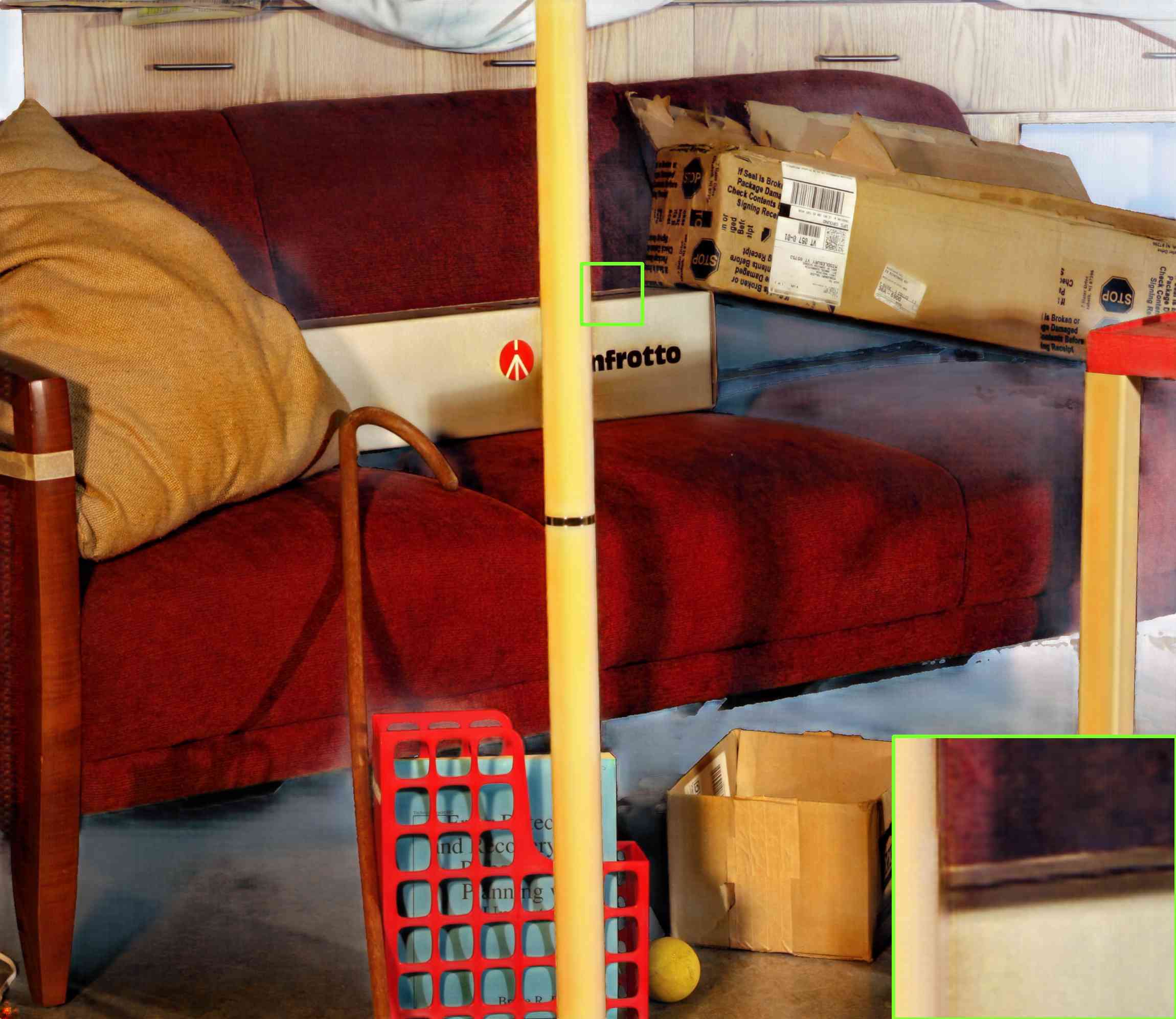}
	\end{minipage}
	\begin{minipage}[h]{0.118\linewidth}
		\centering
		\includegraphics[width=\linewidth]{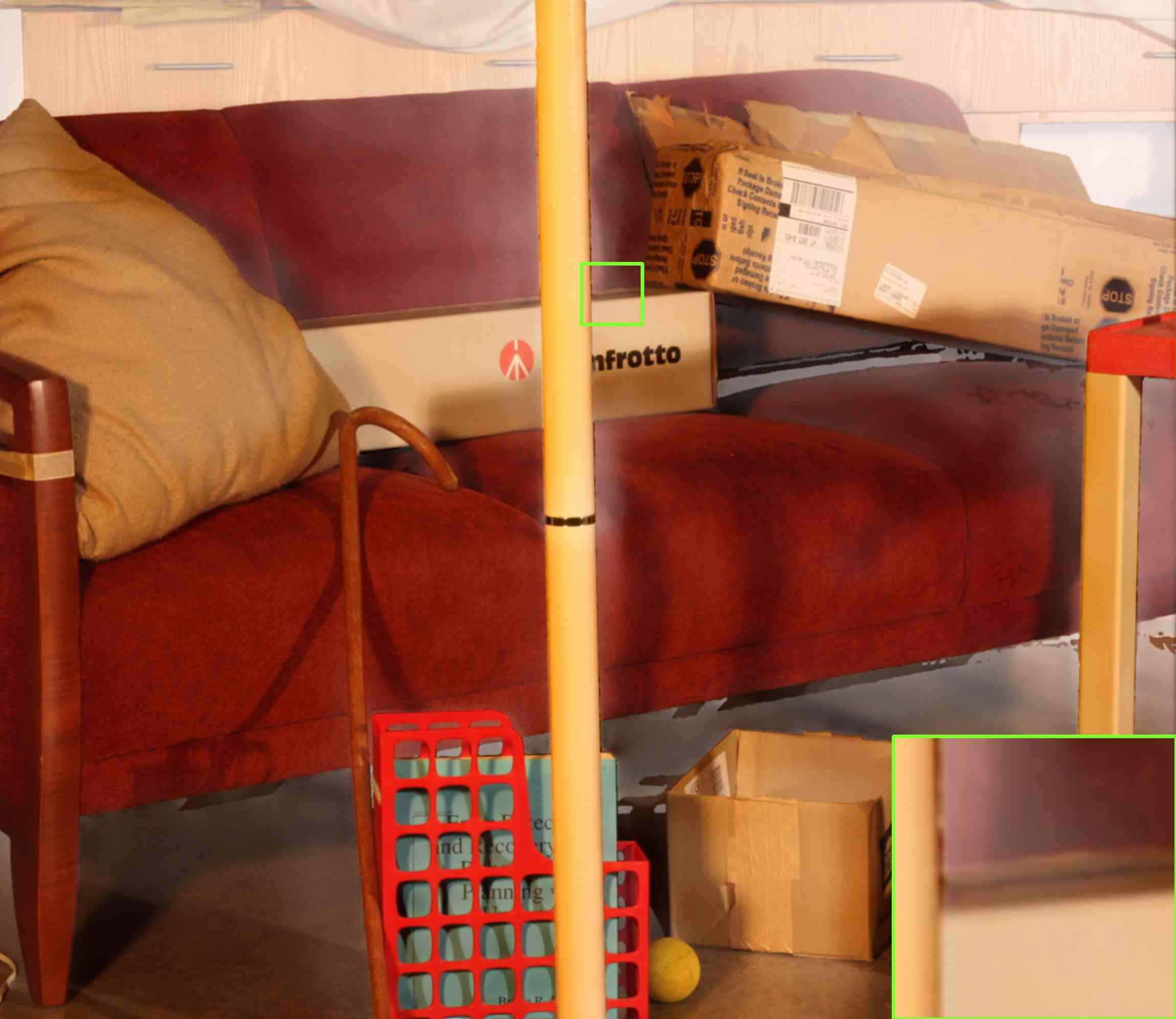}
	\end{minipage}
	\begin{minipage}[h]{0.118\linewidth}
		\centering
		\includegraphics[width=\linewidth]{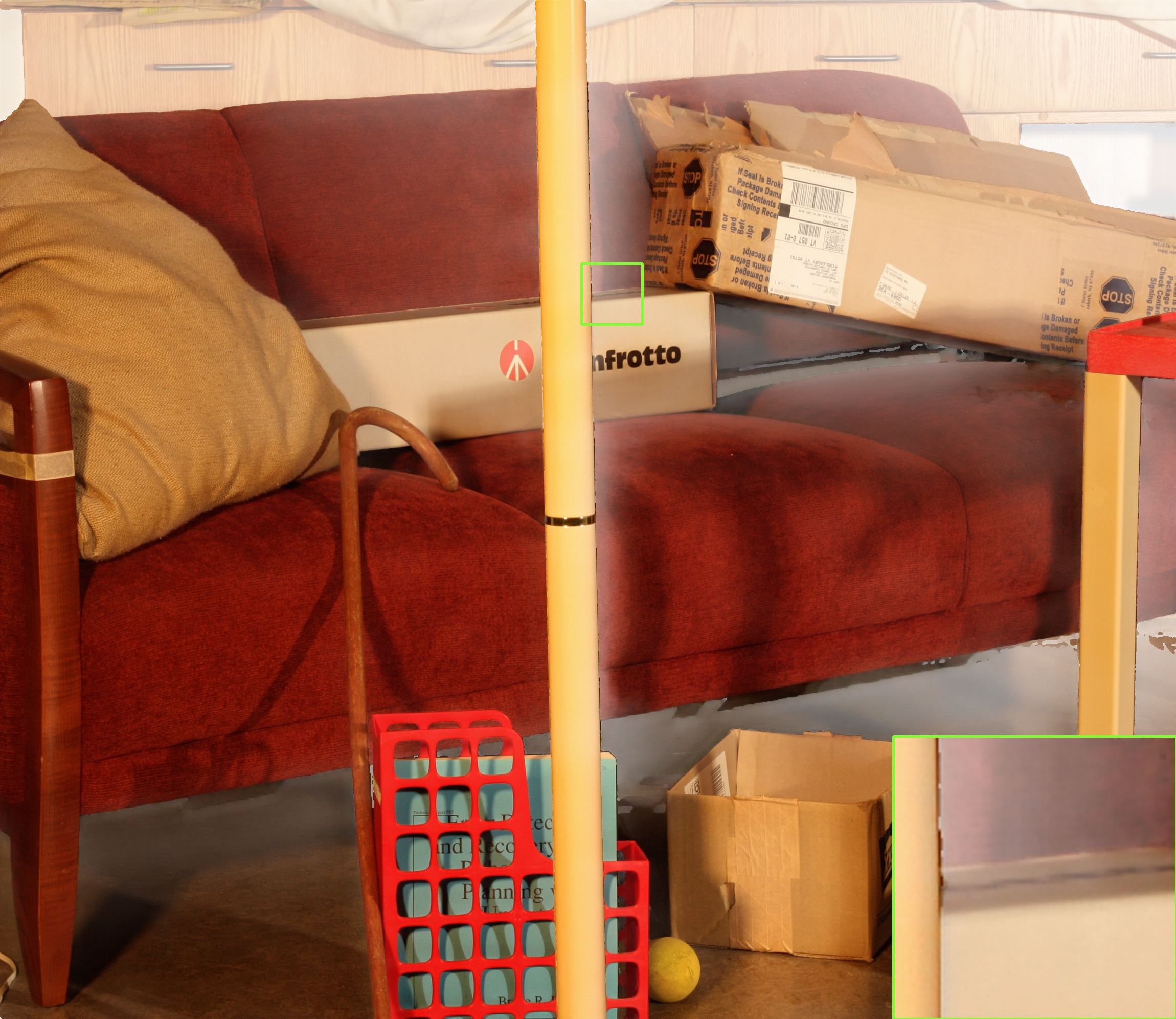}
	\end{minipage}
	\begin{minipage}[h]{0.118\linewidth}
		\centering
		\includegraphics[width=\linewidth]{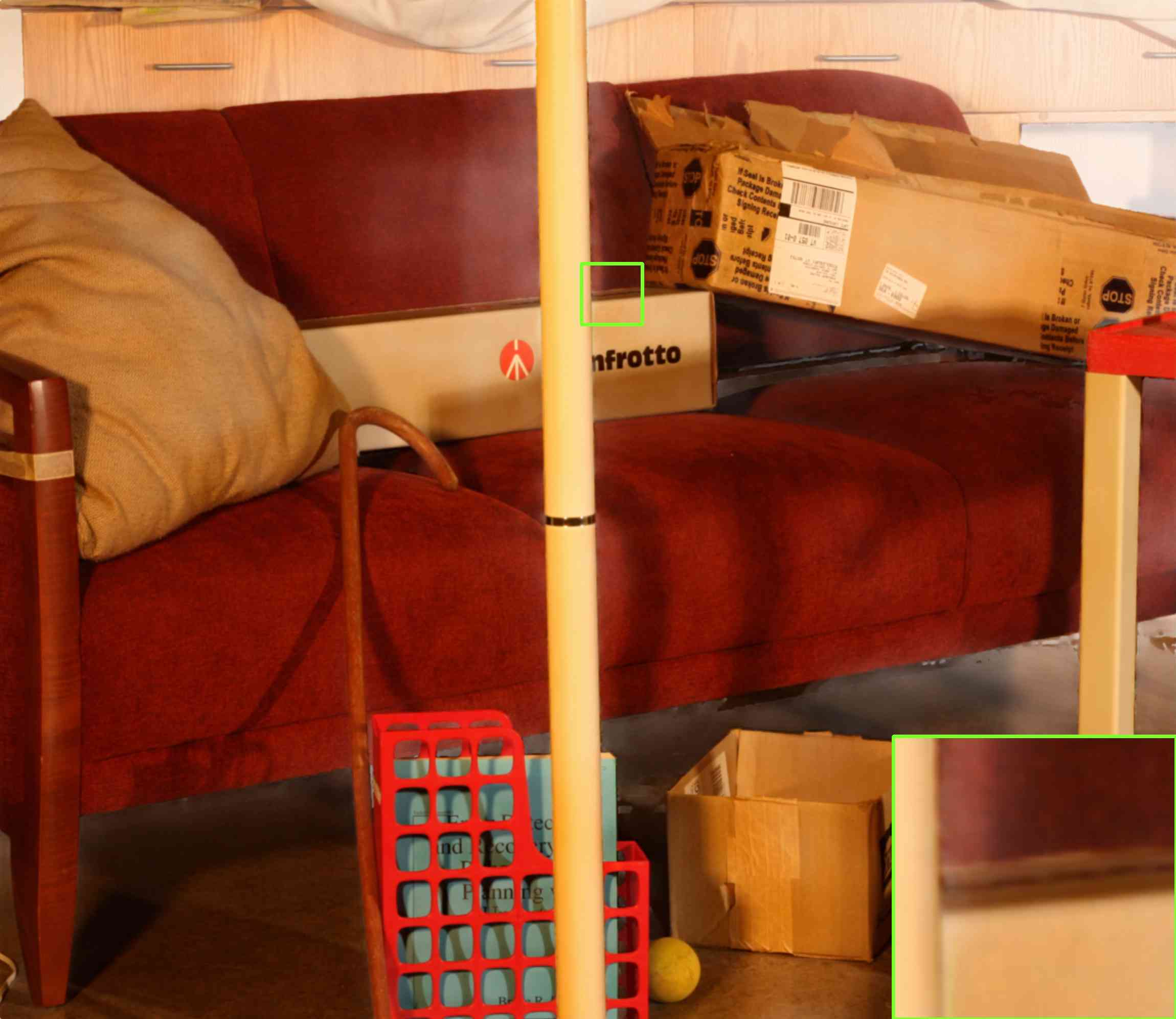}
	\end{minipage}
	\begin{minipage}[h]{0.118\linewidth}
		\centering
		\includegraphics[width=\linewidth]{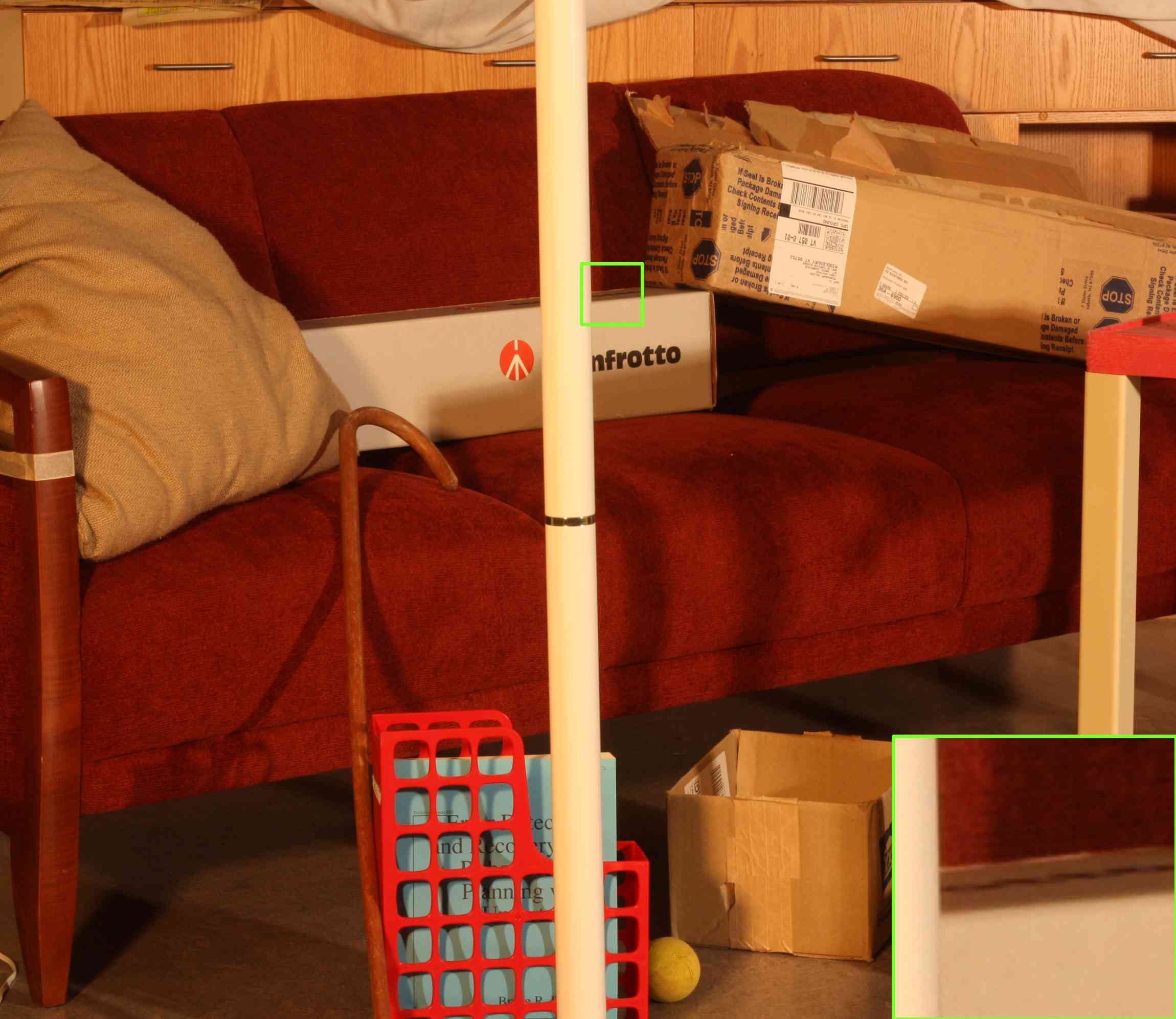}
	\end{minipage}
	\vspace{1mm}
	% HazeRD 1
	\begin{minipage}[h]{0.118\linewidth}
		\centering
		\includegraphics[width=\linewidth]{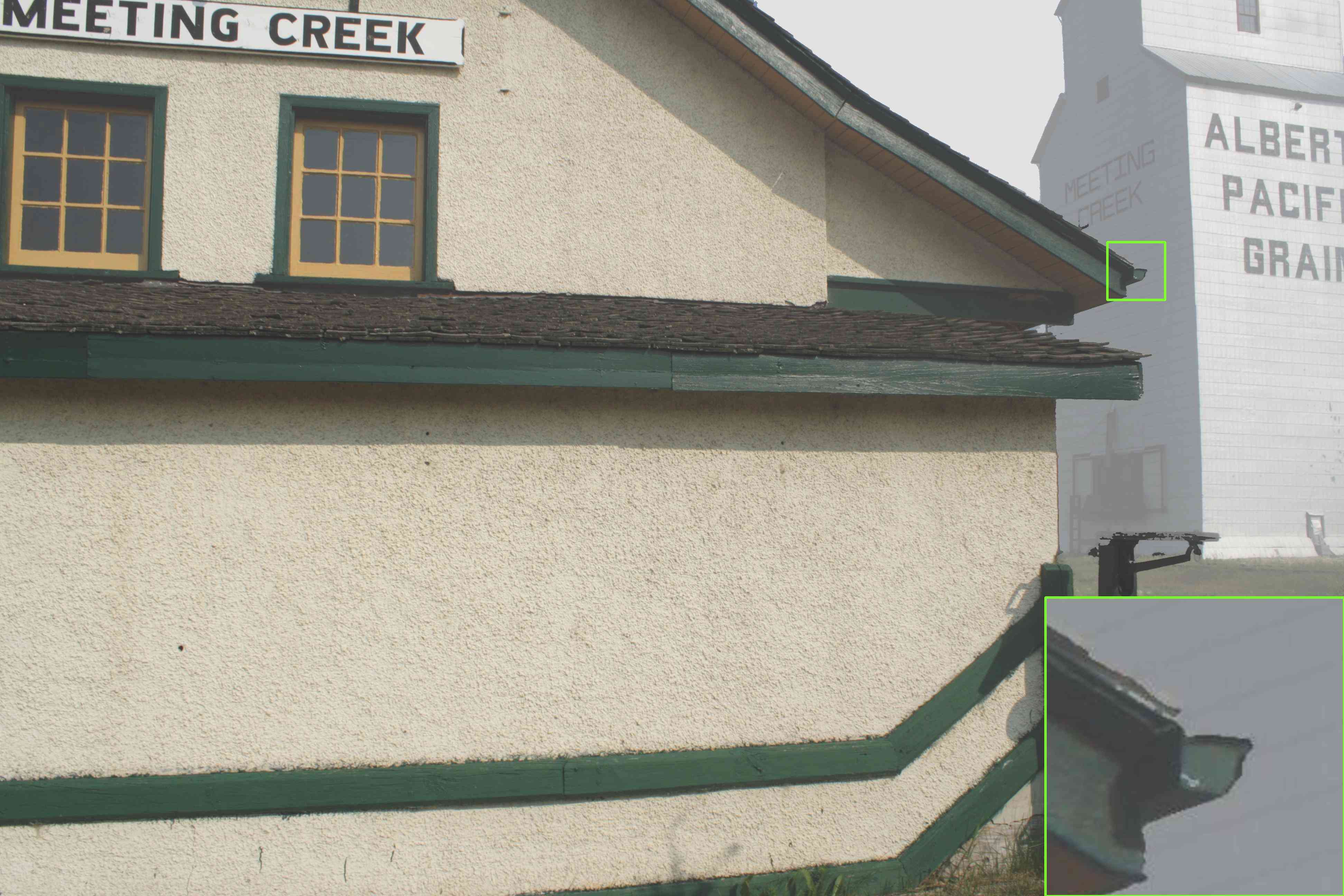}
	\end{minipage}
%	\begin{minipage}[h]{0.118\linewidth}
%		\centering
%		\includegraphics[width=\linewidth]{images/syn/HazeRD/1/AOD_IMG_8583_500.jpg}
%	\end{minipage}
	\begin{minipage}[h]{0.118\linewidth}
		\centering
		\includegraphics[width=\linewidth]{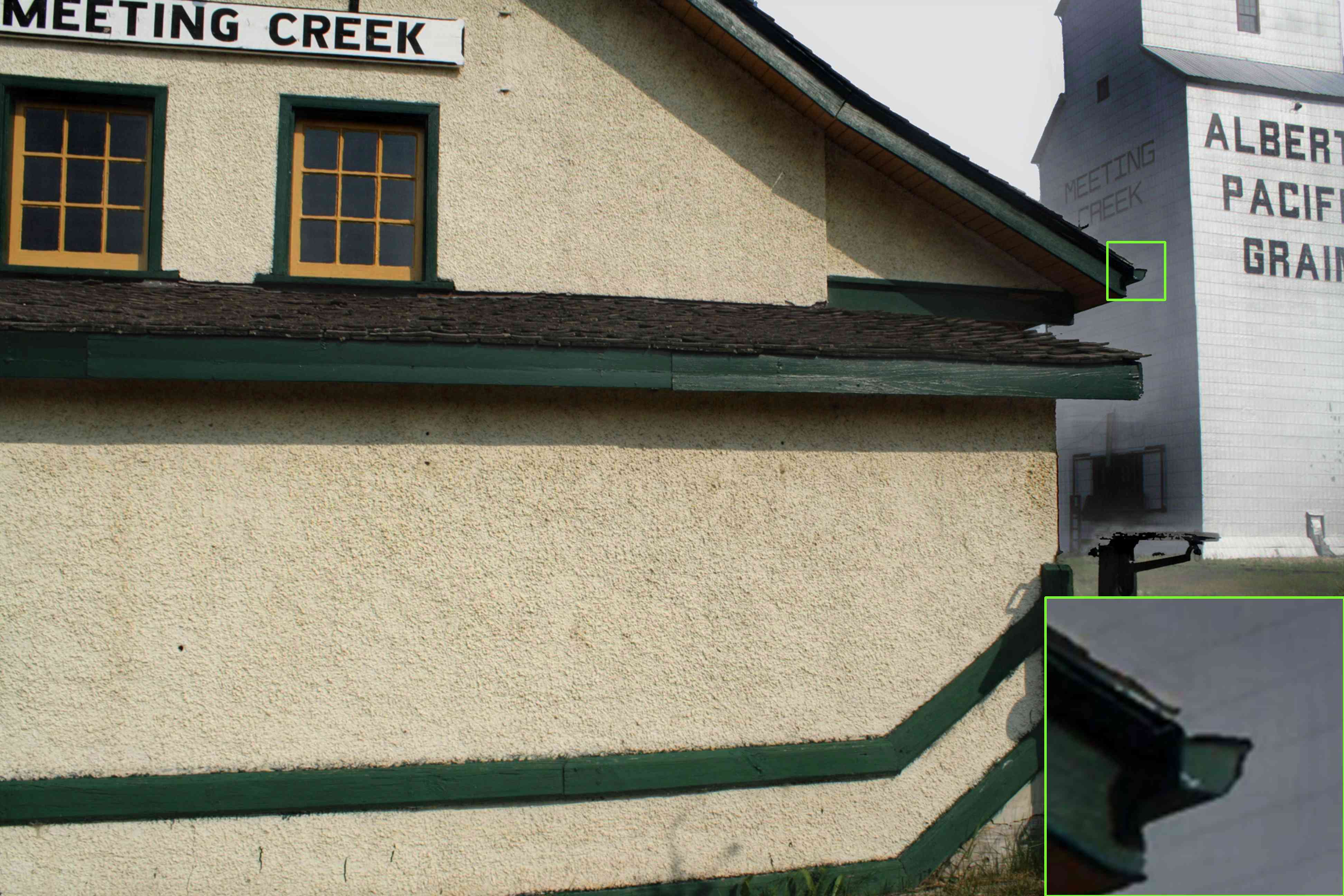}
	\end{minipage}
	\begin{minipage}[h]{0.118\linewidth}
		\centering
		\includegraphics[width=\linewidth]{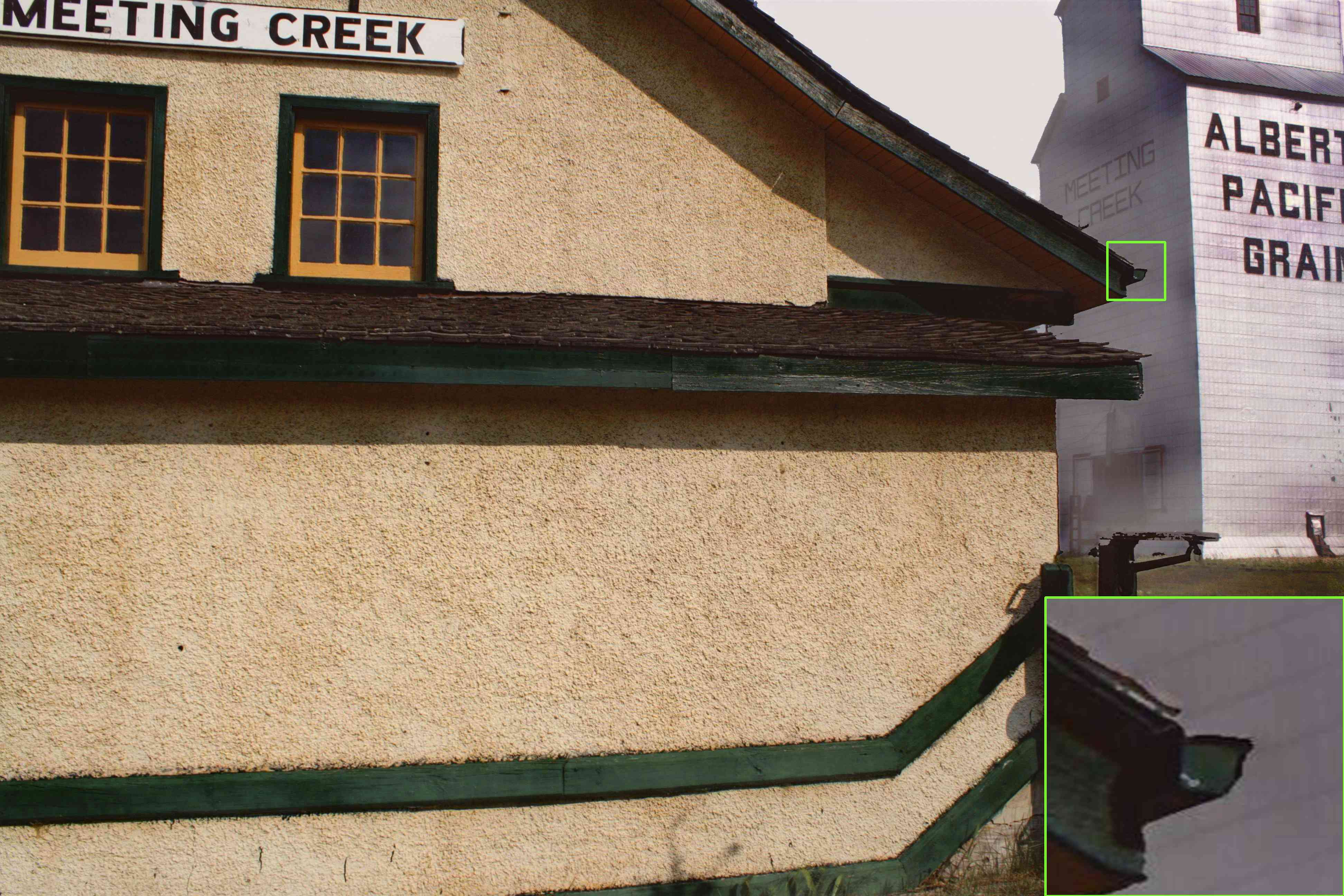}
	\end{minipage}
	\begin{minipage}[h]{0.118\linewidth}
		\centering
		\includegraphics[width=\linewidth]{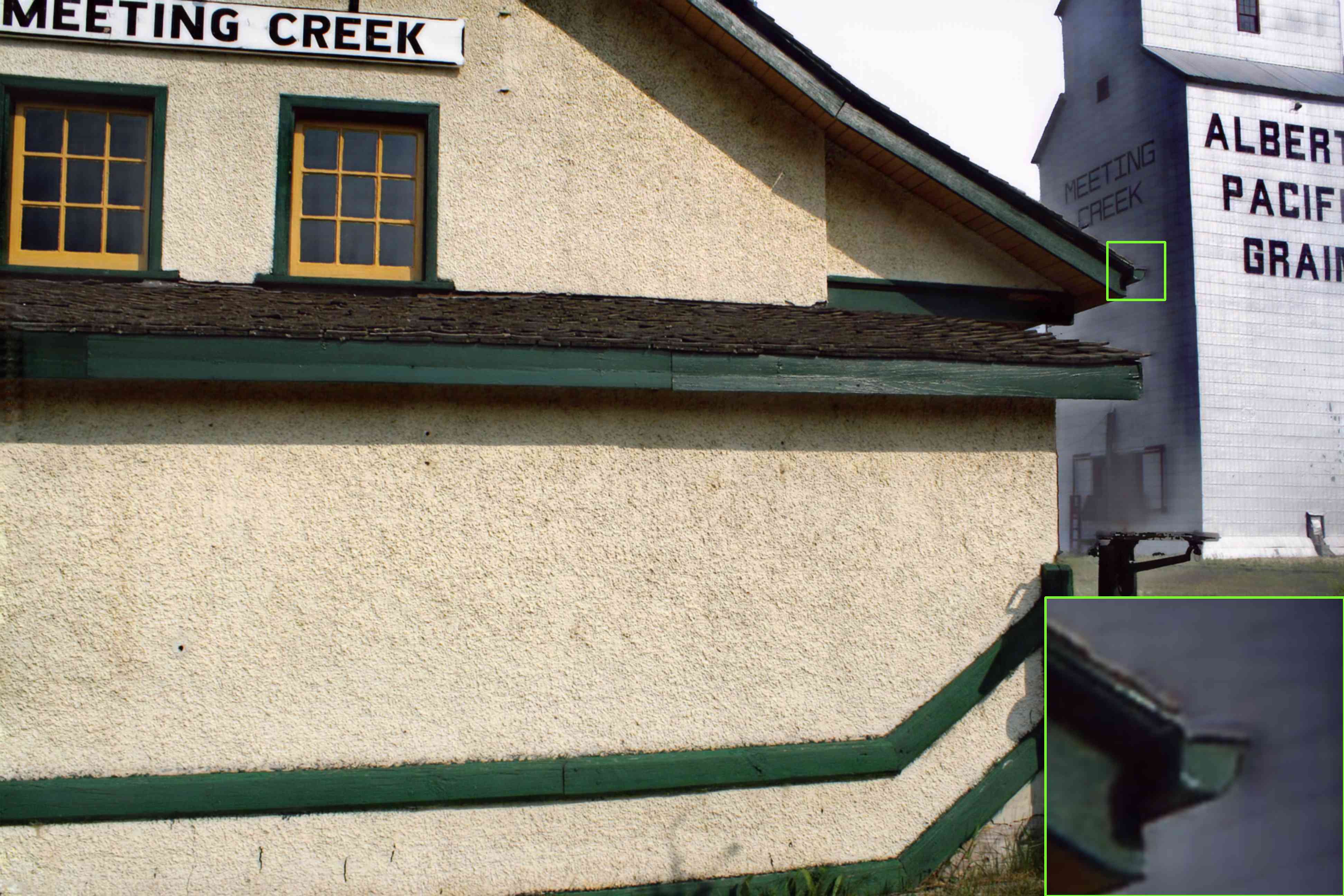}
	\end{minipage}
	\begin{minipage}[h]{0.118\linewidth}
		\centering
		\includegraphics[width=\linewidth]{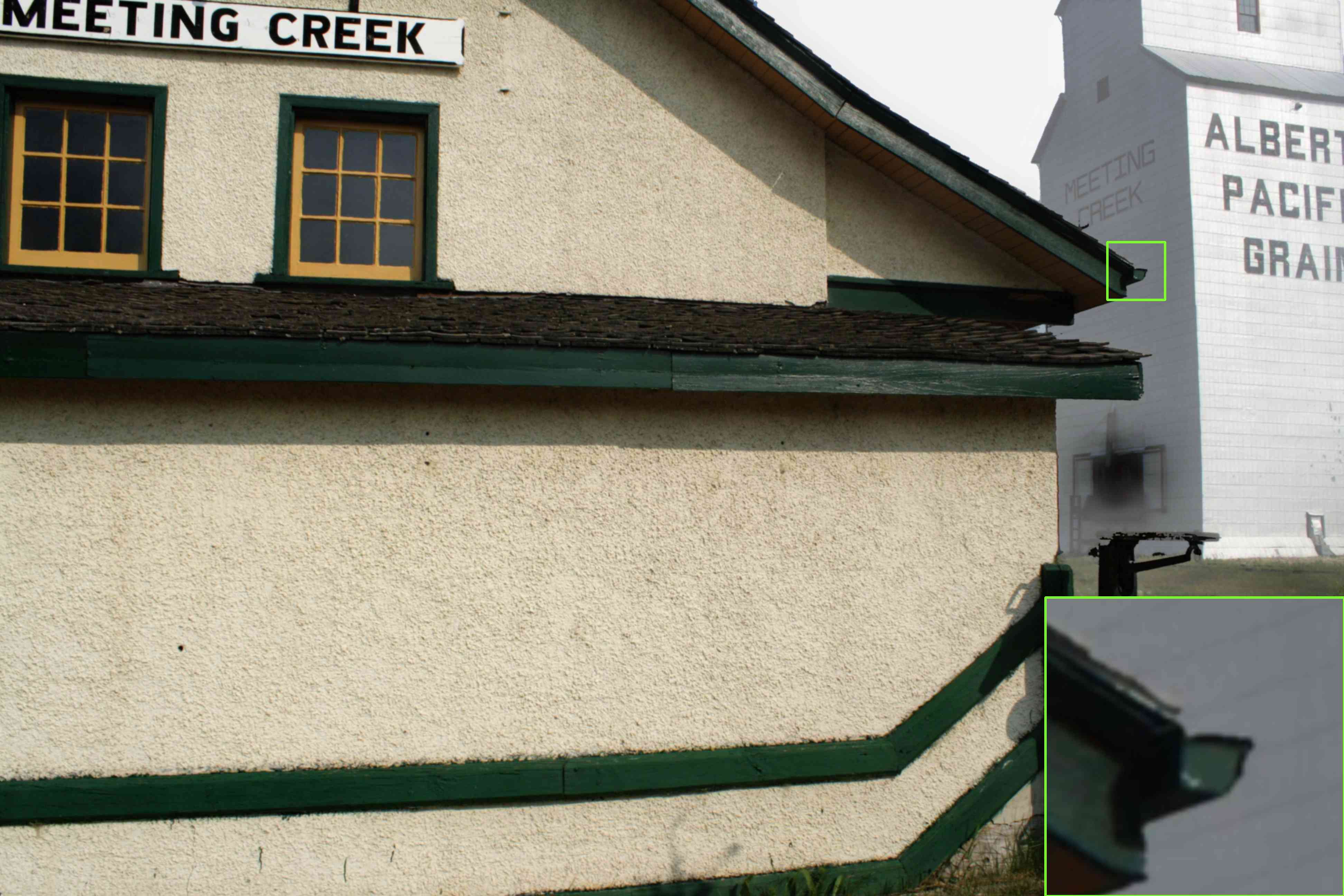}
	\end{minipage}
	\begin{minipage}[h]{0.118\linewidth}
		\centering
		\includegraphics[width=\linewidth]{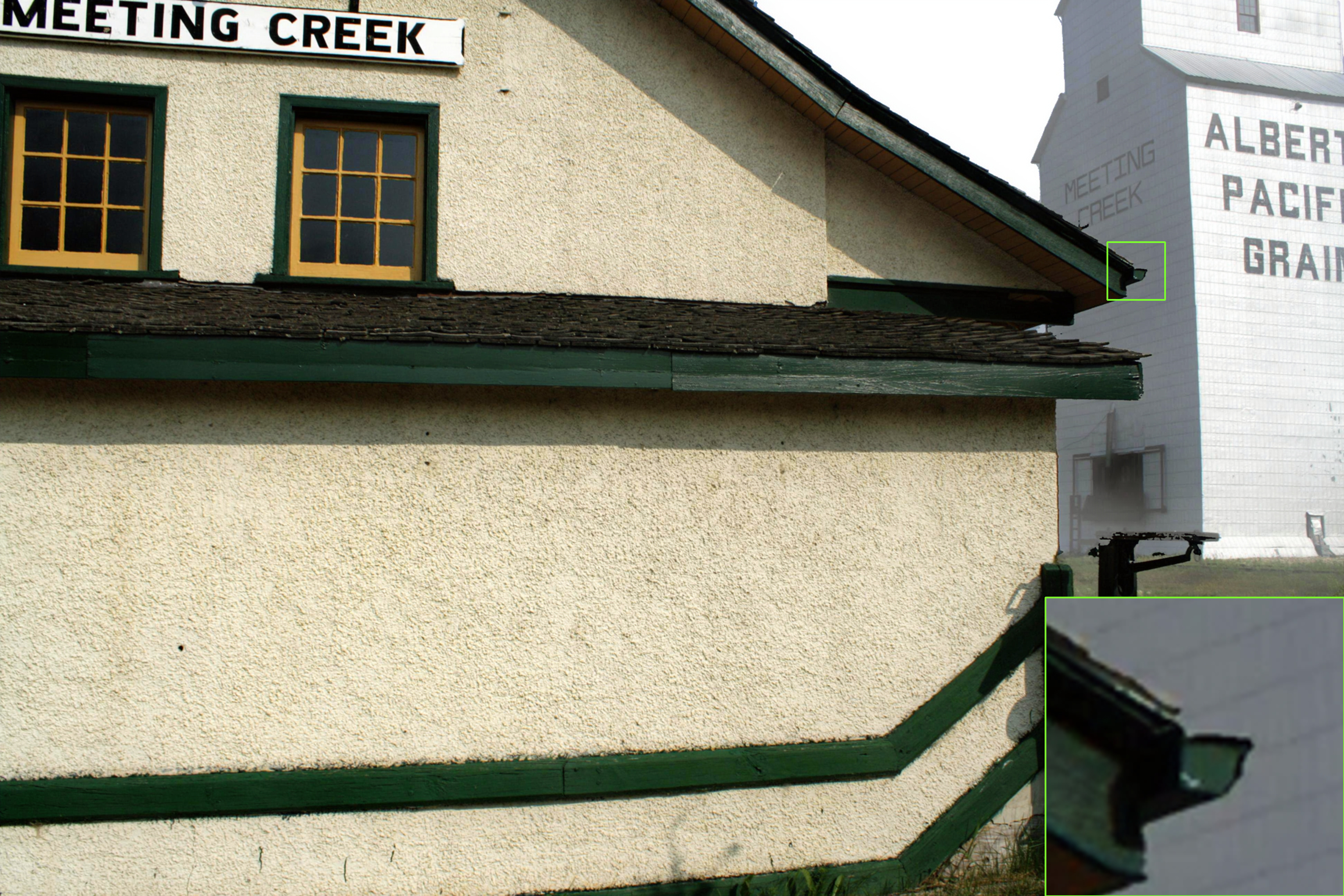}
	\end{minipage}
	\begin{minipage}[h]{0.118\linewidth}
		\centering
		\includegraphics[width=\linewidth]{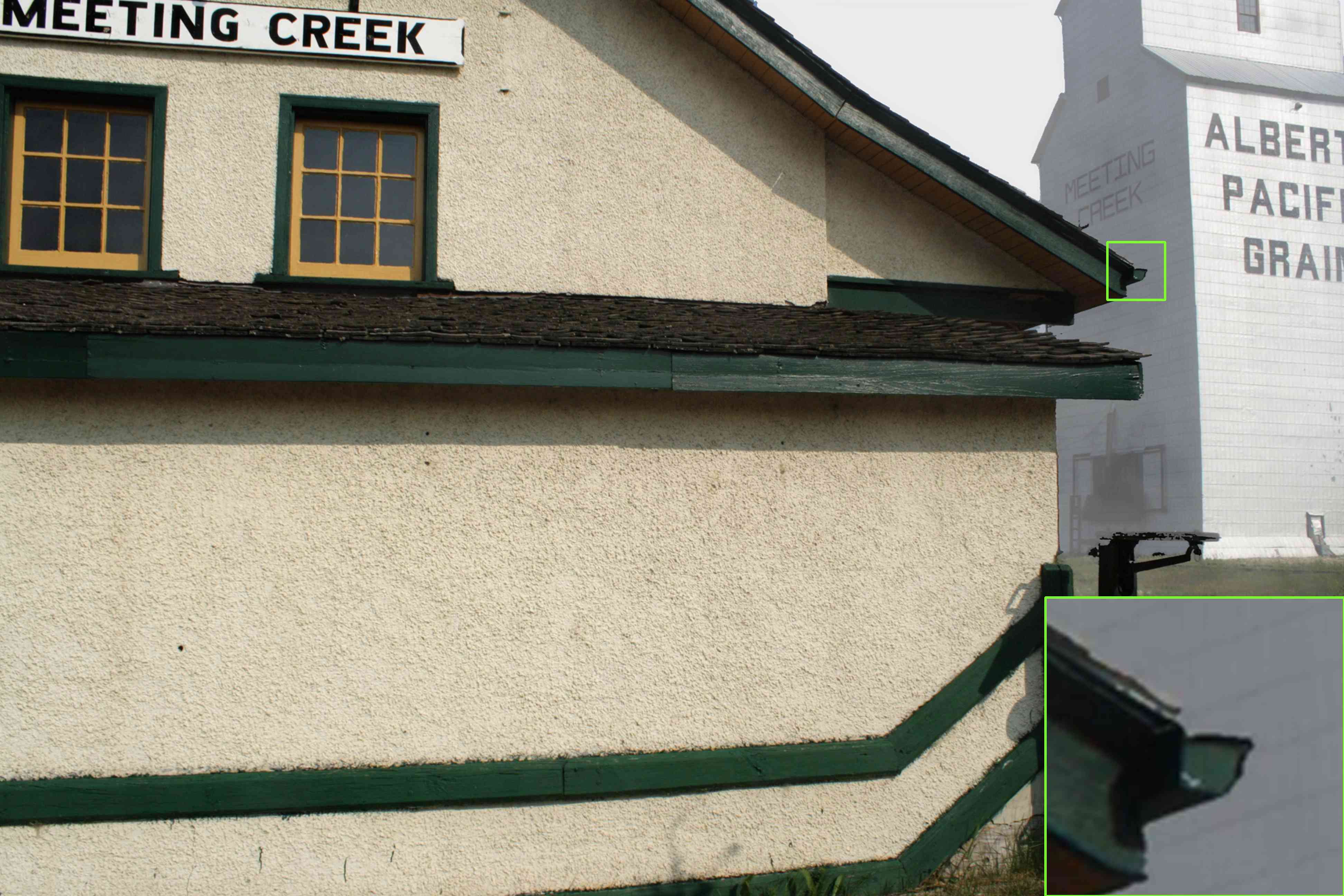}
	\end{minipage}
	\begin{minipage}[h]{0.118\linewidth}
		\centering
		\includegraphics[width=\linewidth]{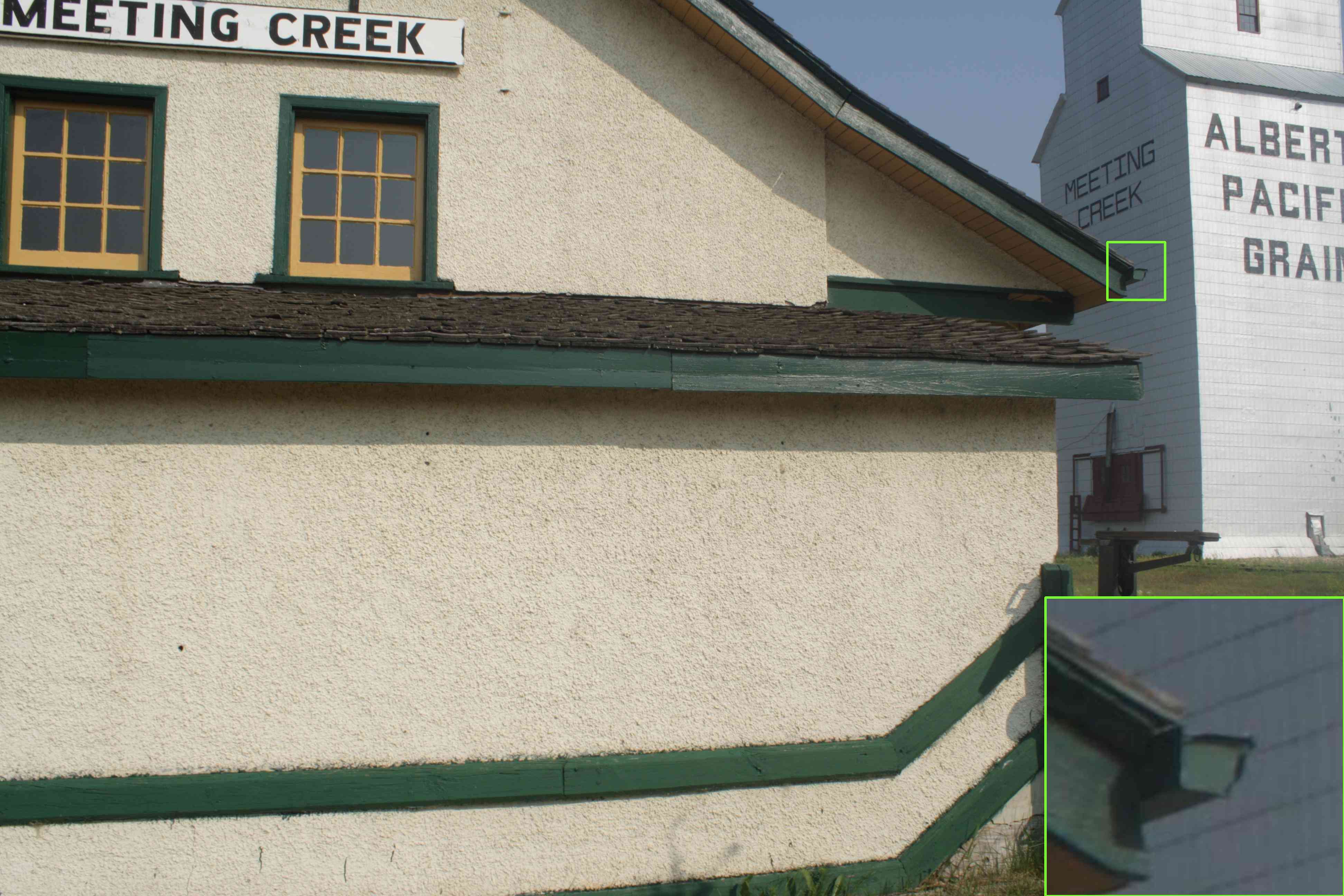}
	\end{minipage}
	\vspace{1mm}
	% HazeRD 2
	\begin{minipage}[h]{0.118\linewidth}
		\centering
		\includegraphics[width=\linewidth]{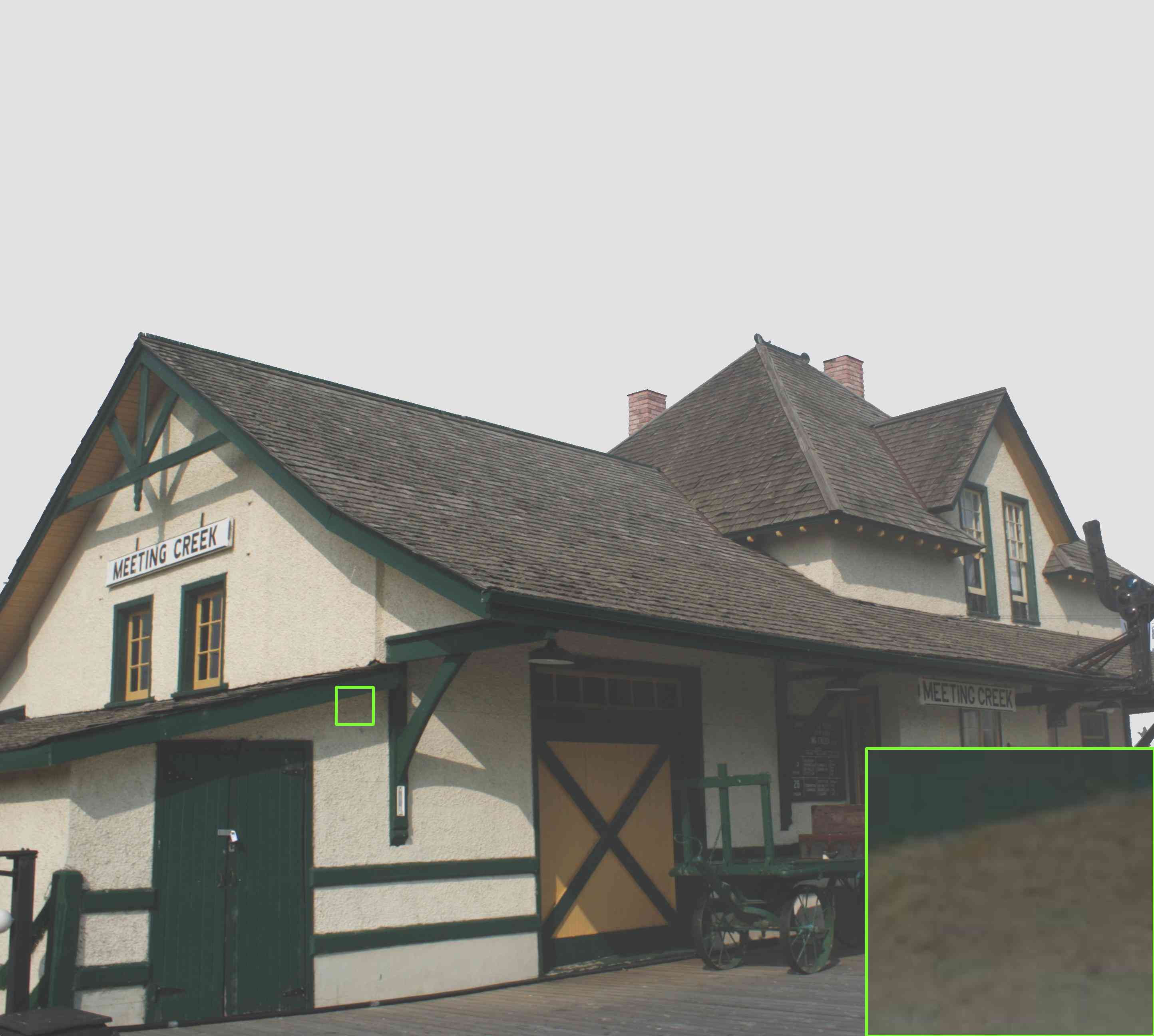}
	\end{minipage}
%	\begin{minipage}[h]{0.118\linewidth}
%		\centering
%		\includegraphics[width=\linewidth]{images/syn/HazeRD/2/AOD_IMG_8612_1000.jpg}
%	\end{minipage}
	\begin{minipage}[h]{0.118\linewidth}
		\centering
		\includegraphics[width=\linewidth]{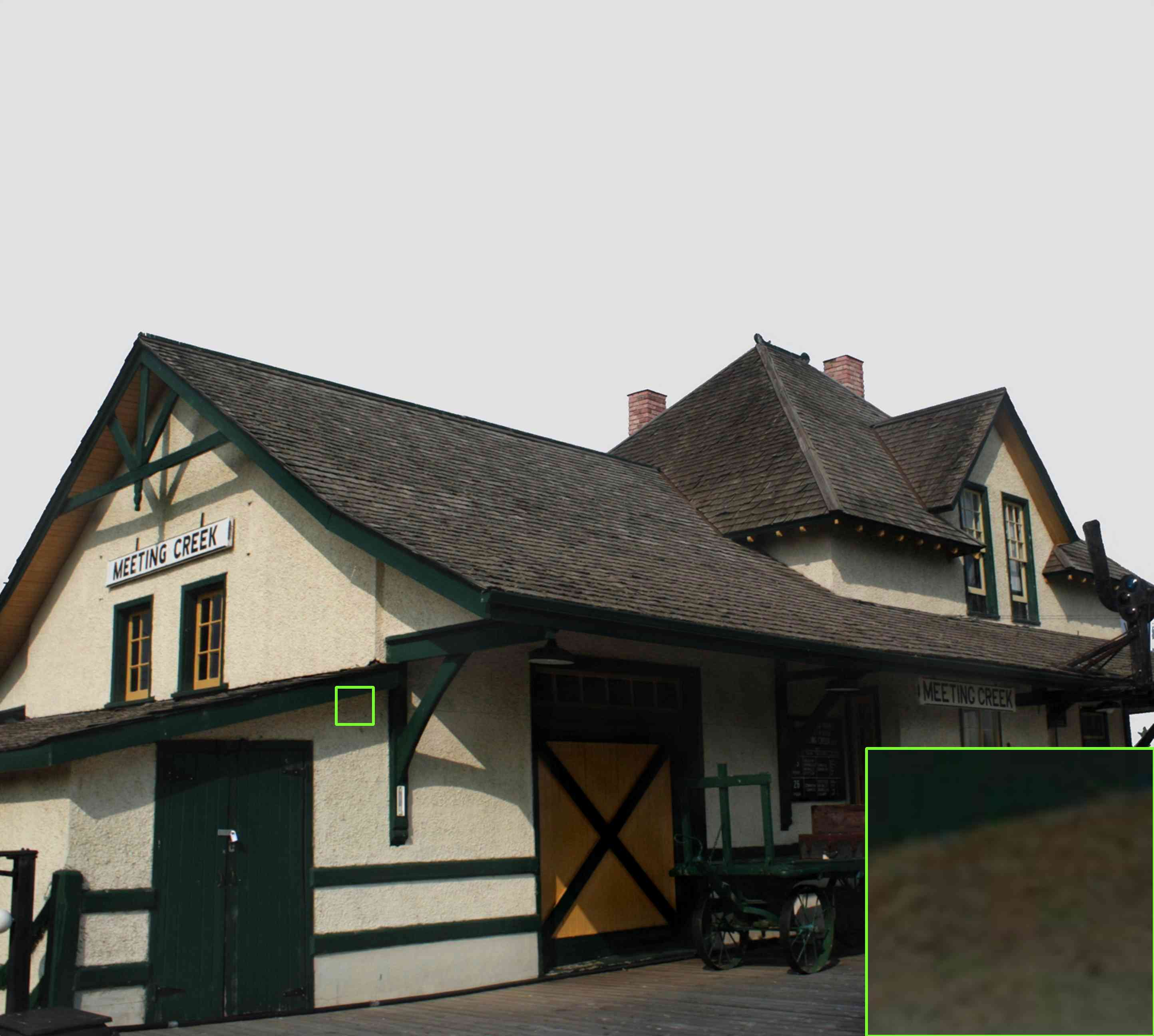}
	\end{minipage}
	\begin{minipage}[h]{0.118\linewidth}
		\centering
		\includegraphics[width=\linewidth]{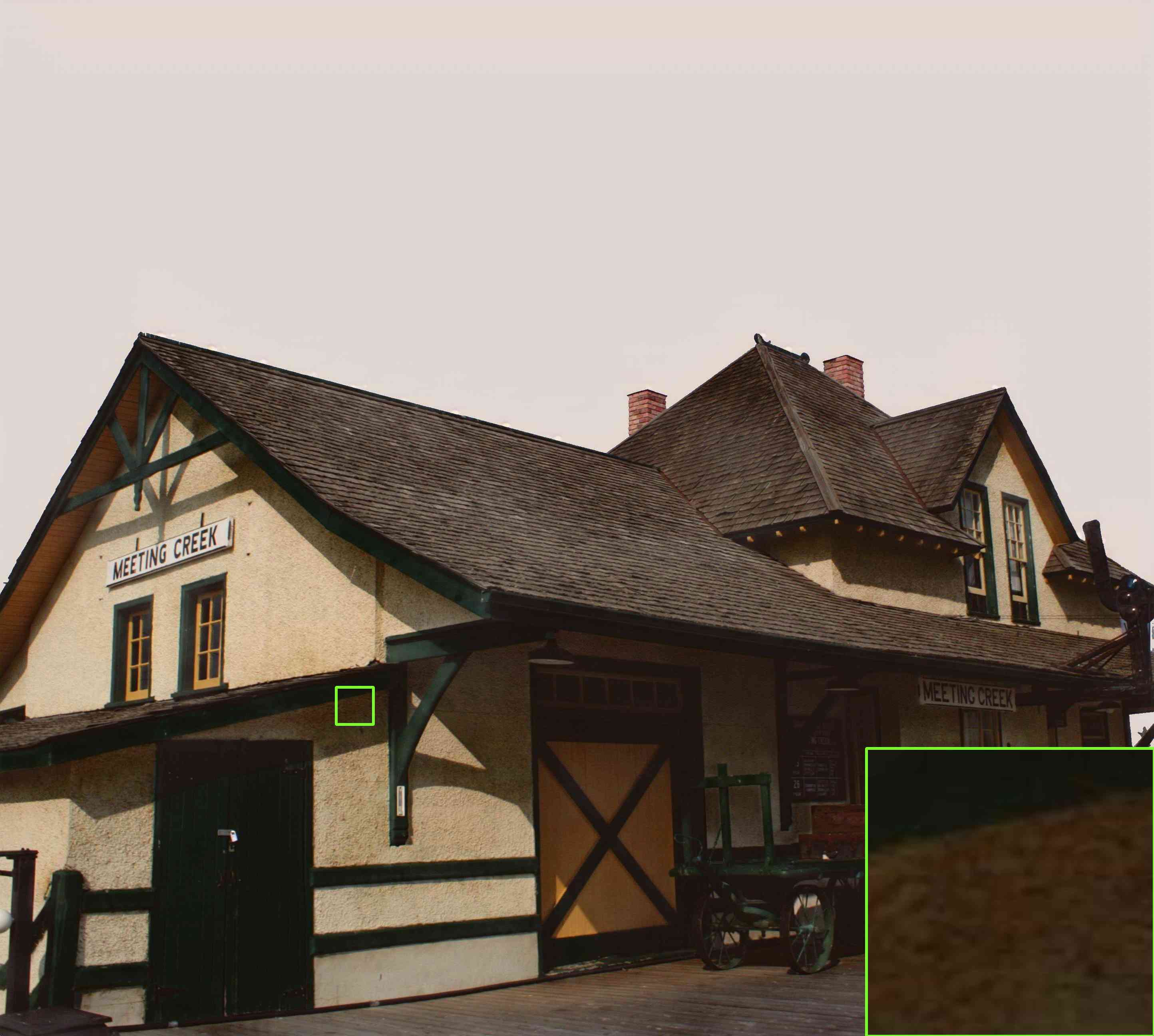}
	\end{minipage}
	\begin{minipage}[h]{0.118\linewidth}
		\centering
		\includegraphics[width=\linewidth]{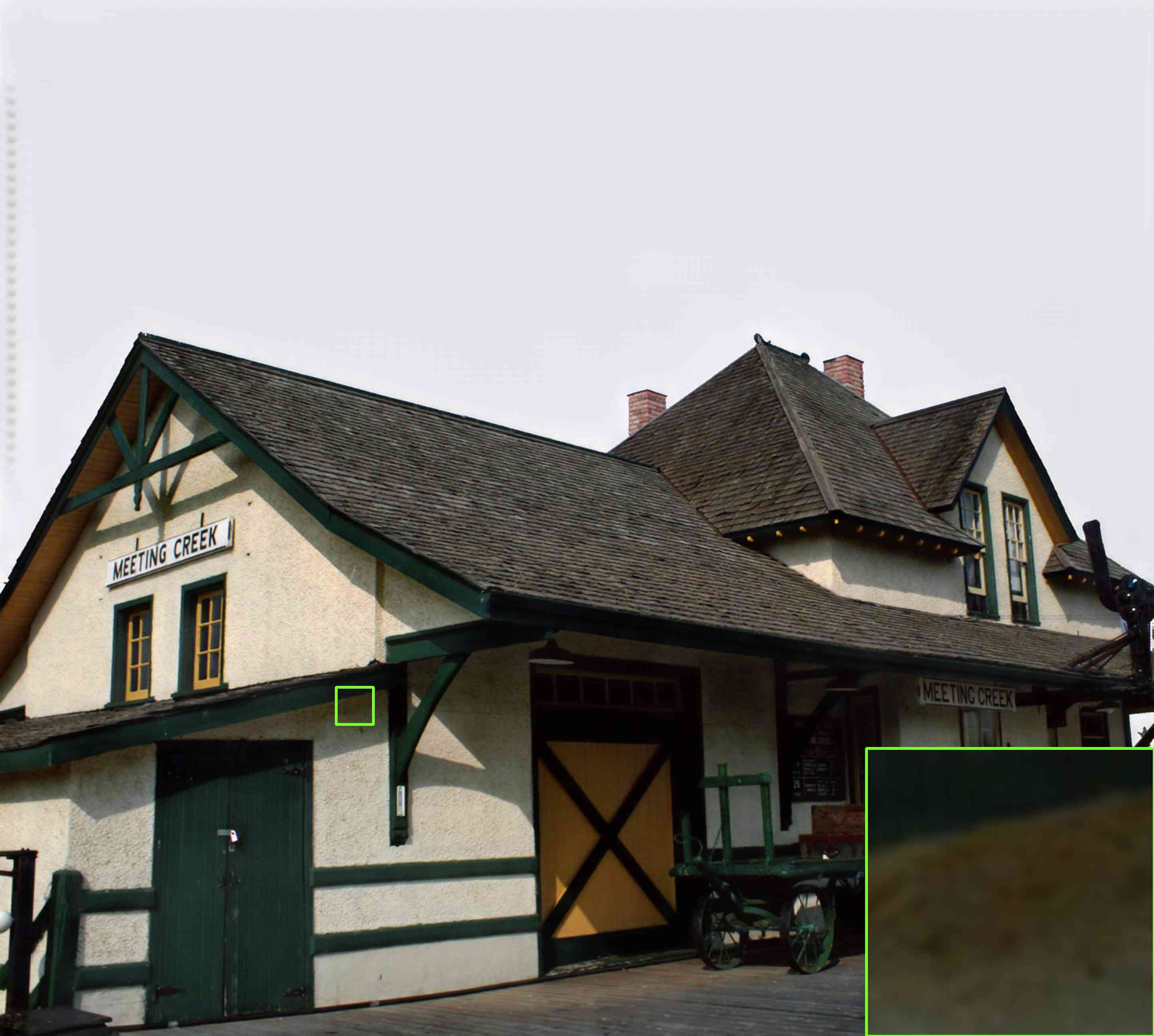}
	\end{minipage}
	\begin{minipage}[h]{0.118\linewidth}
		\centering
		\includegraphics[width=\linewidth]{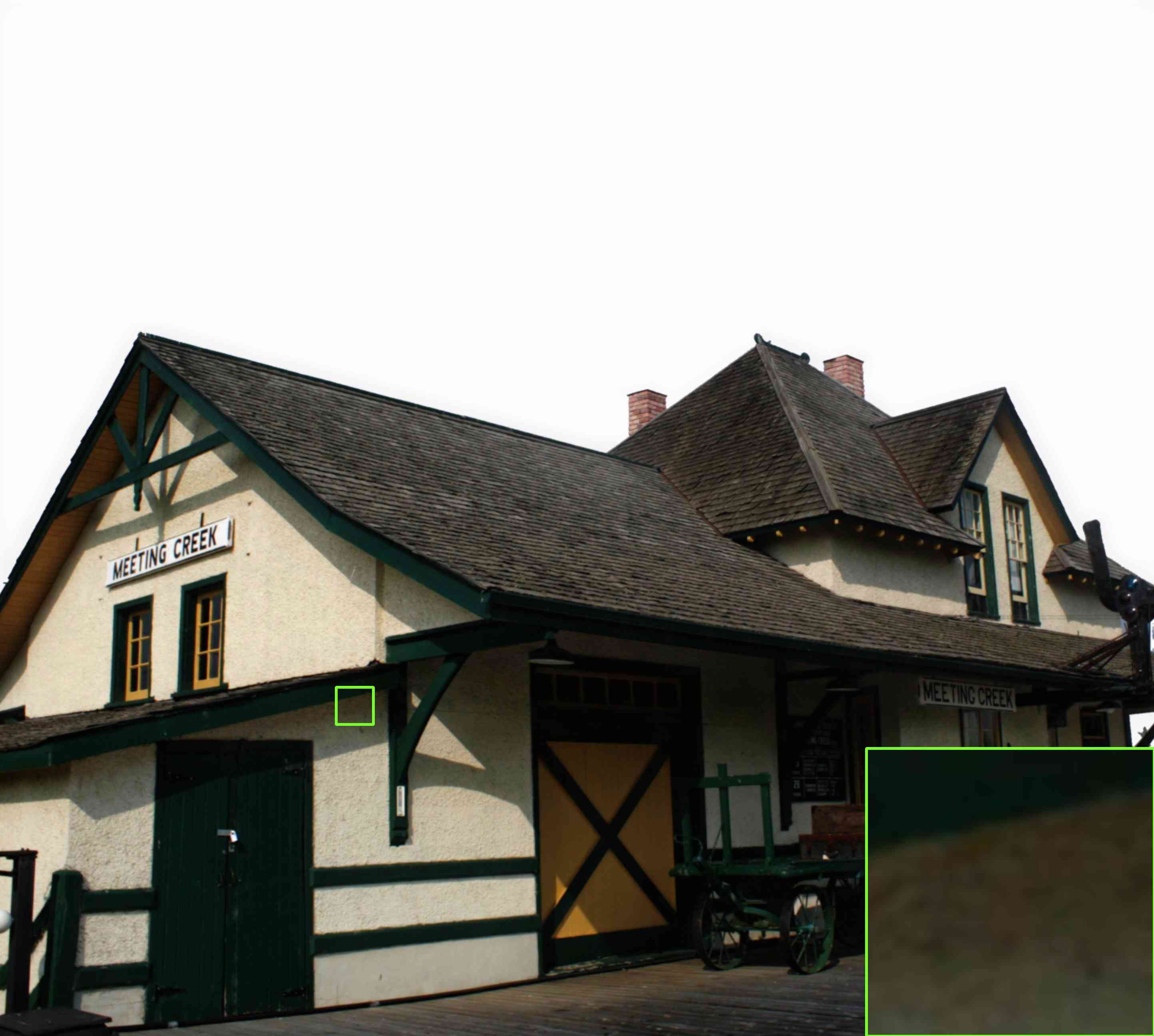}
	\end{minipage}
	\begin{minipage}[h]{0.118\linewidth}
		\centering
		\includegraphics[width=\linewidth]{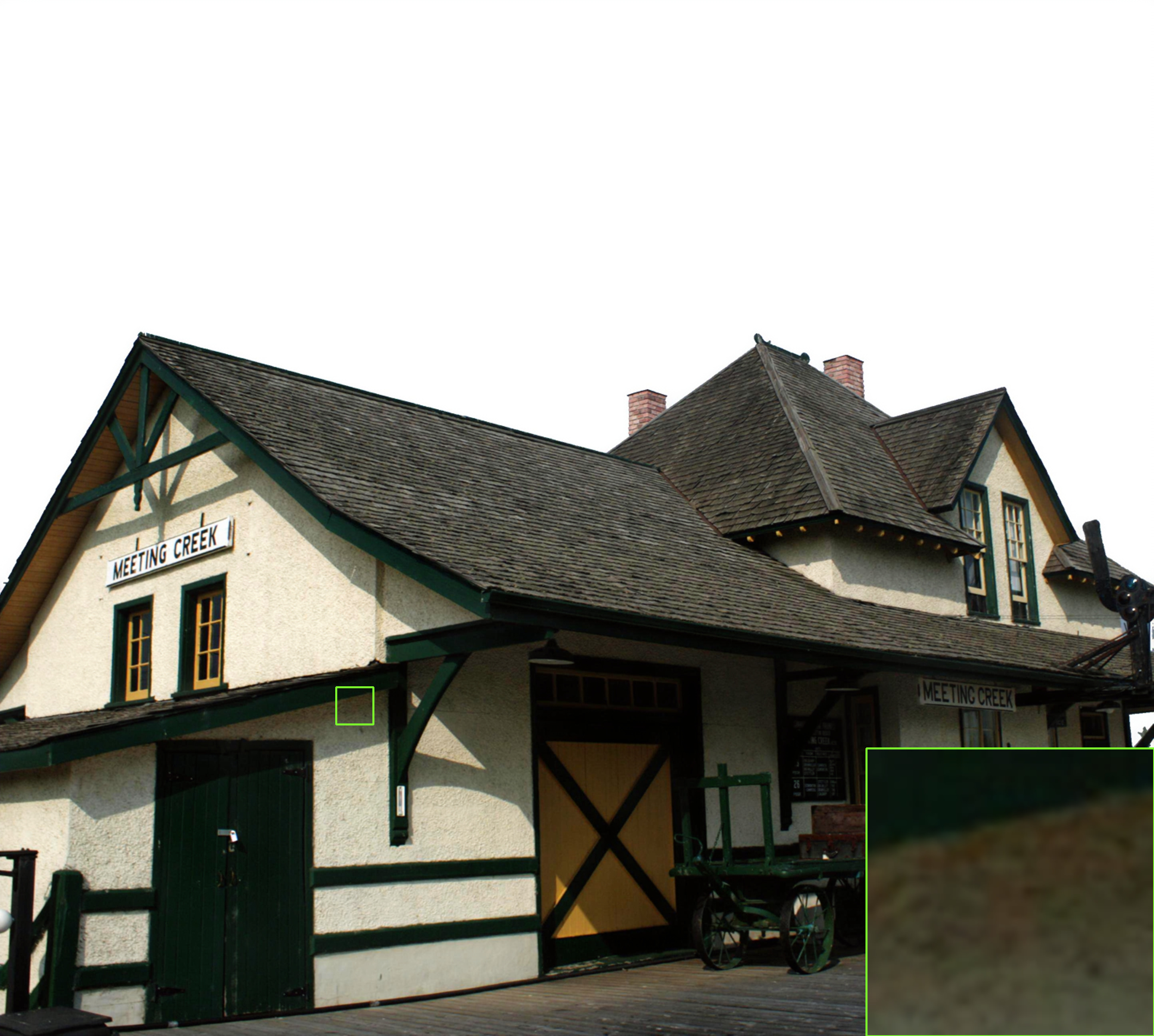}
	\end{minipage}
	\begin{minipage}[h]{0.118\linewidth}
		\centering
		\includegraphics[width=\linewidth]{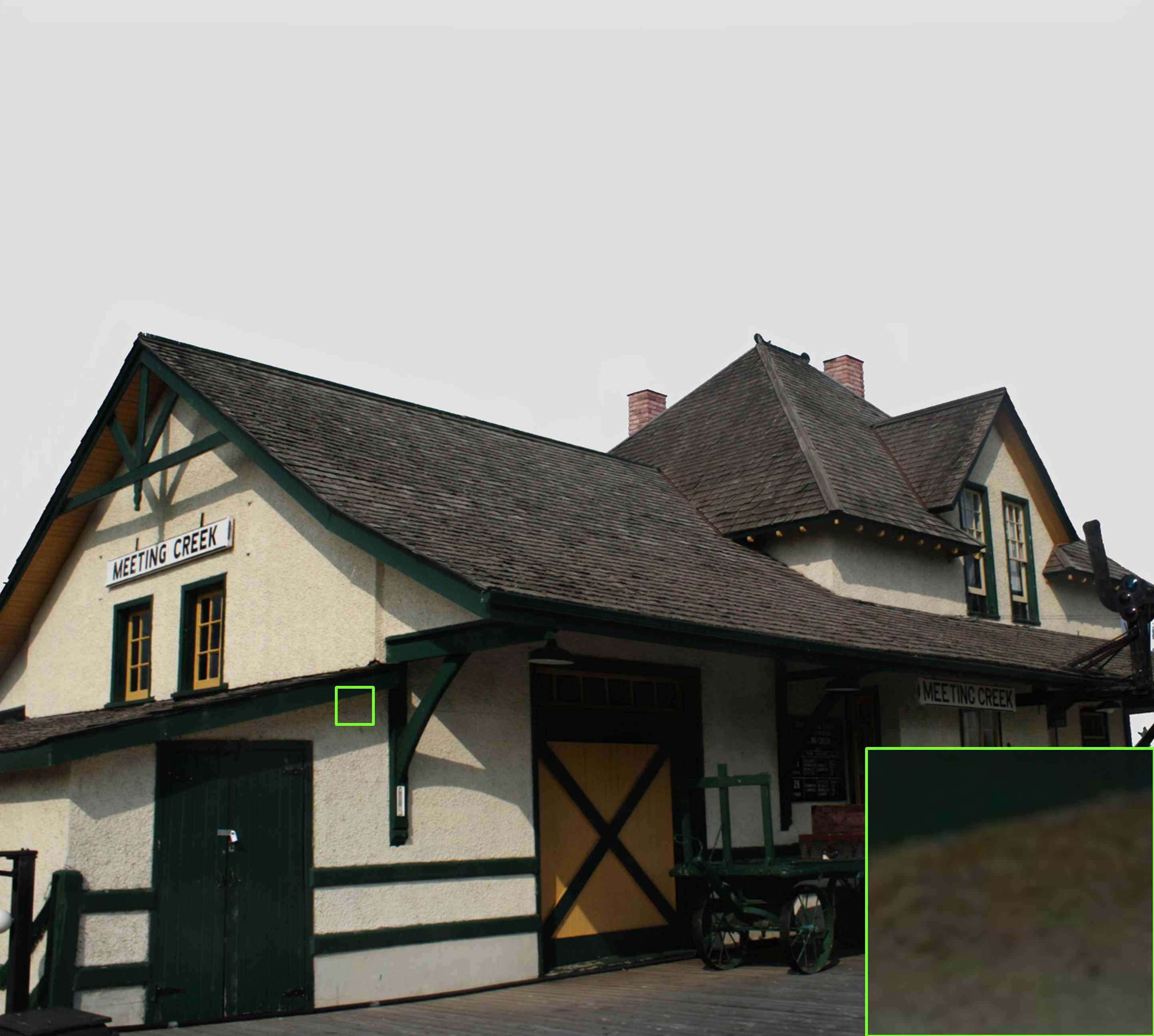}
	\end{minipage}
	\begin{minipage}[h]{0.118\linewidth}
		\centering
		\includegraphics[width=\linewidth]{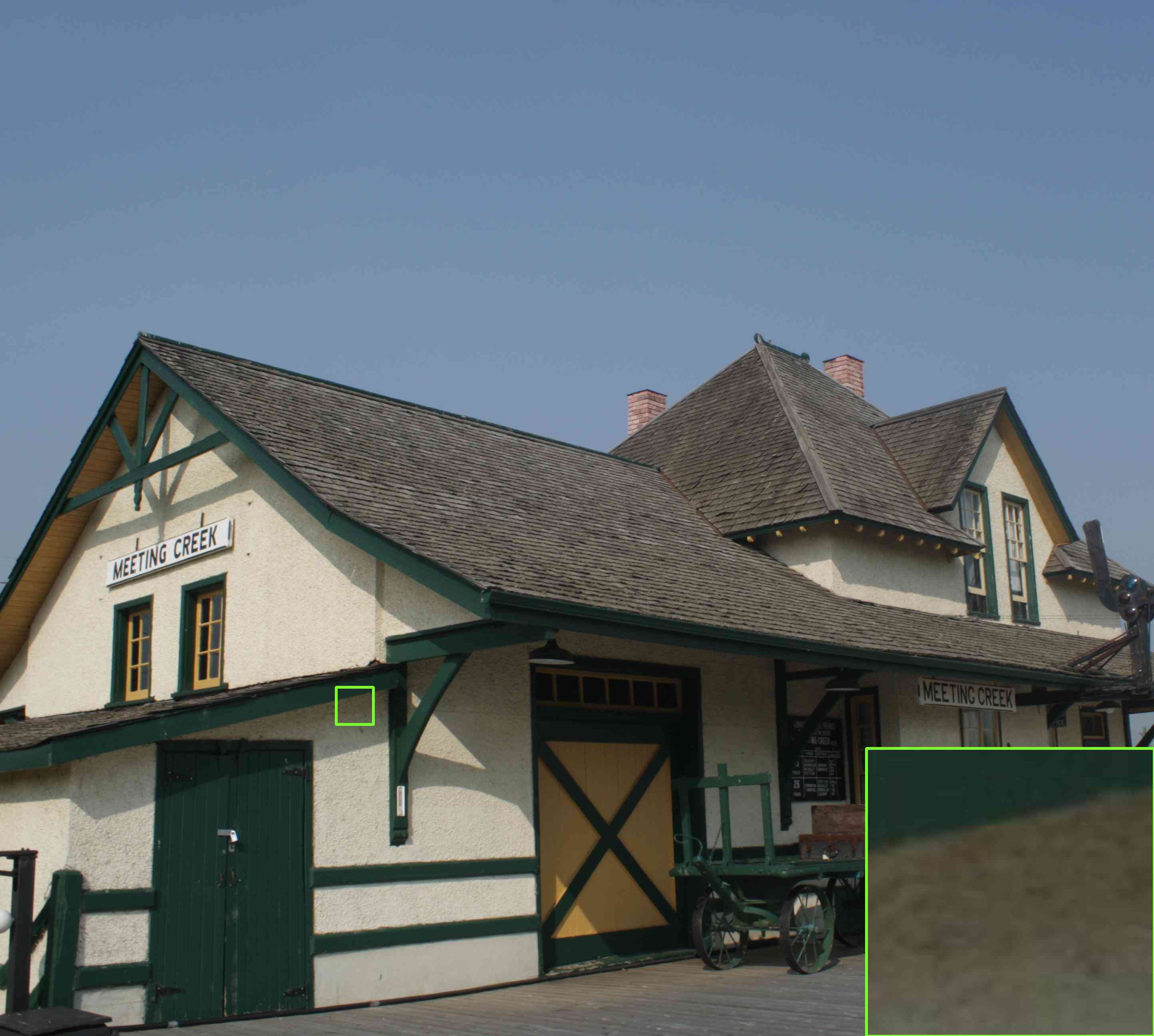}
	\end{minipage}
	\vspace{1mm}
	% O-HAZE 1
	\begin{minipage}[h]{0.118\linewidth}
		\centering
		\includegraphics[width=\linewidth]{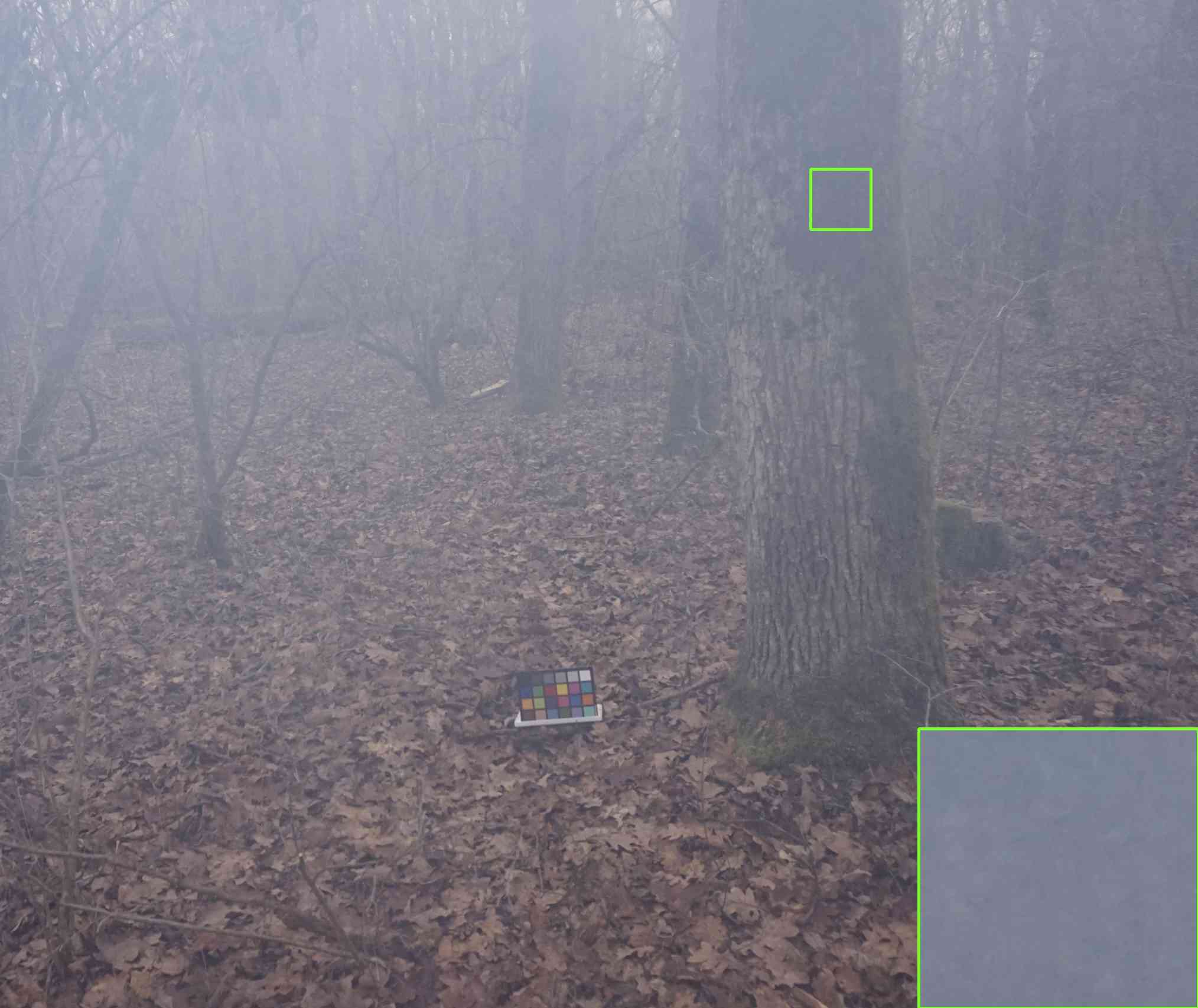}
	\end{minipage}
%	\begin{minipage}[h]{0.118\linewidth}
%		\centering
%		\includegraphics[width=\linewidth]{images/syn/ohaze/1/AOD_42_outdoor.jpg}
%	\end{minipage}
	\begin{minipage}[h]{0.118\linewidth}
		\centering
		\includegraphics[width=\linewidth]{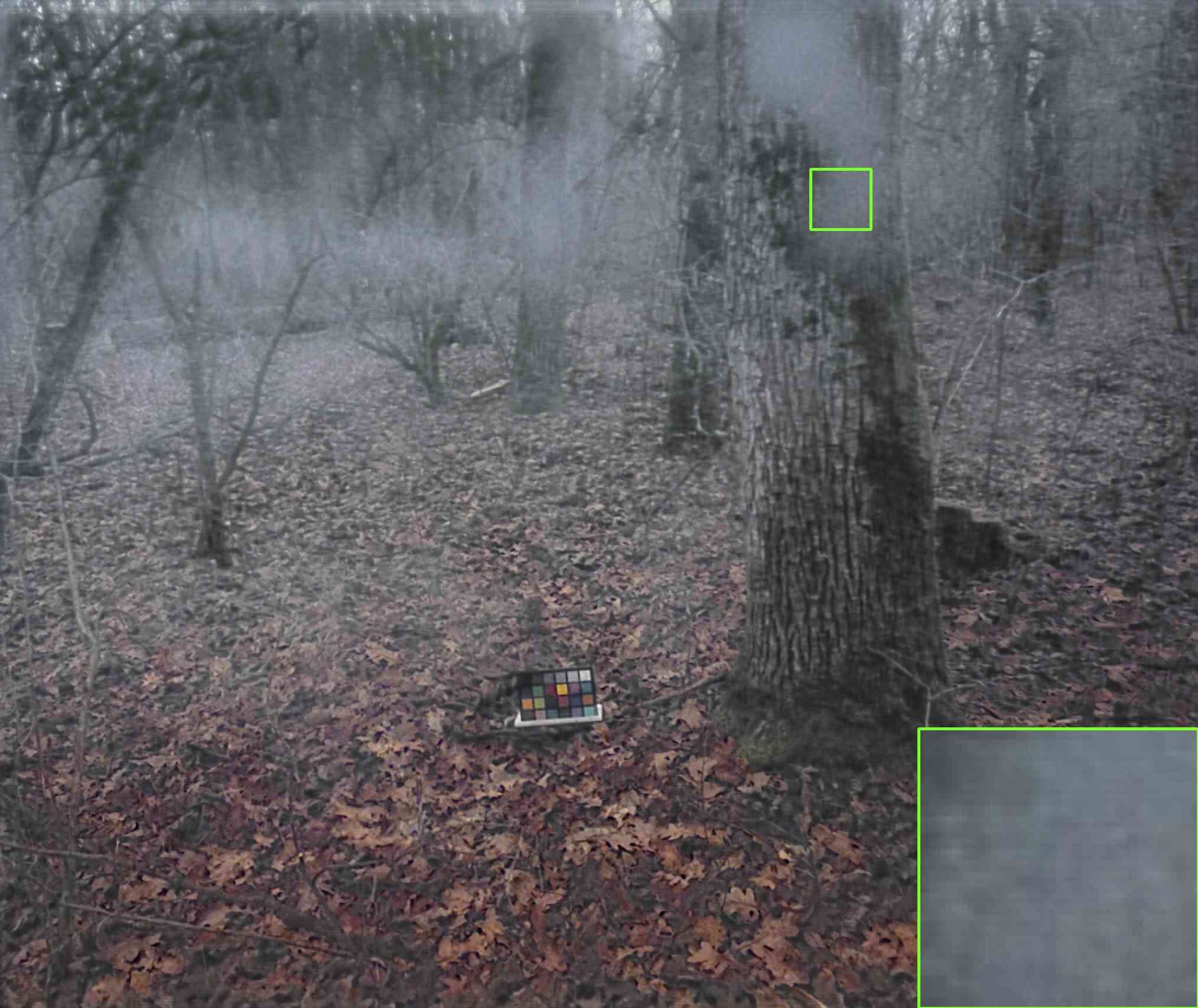}
	\end{minipage}
	\begin{minipage}[h]{0.118\linewidth}
		\centering
		\includegraphics[width=\linewidth]{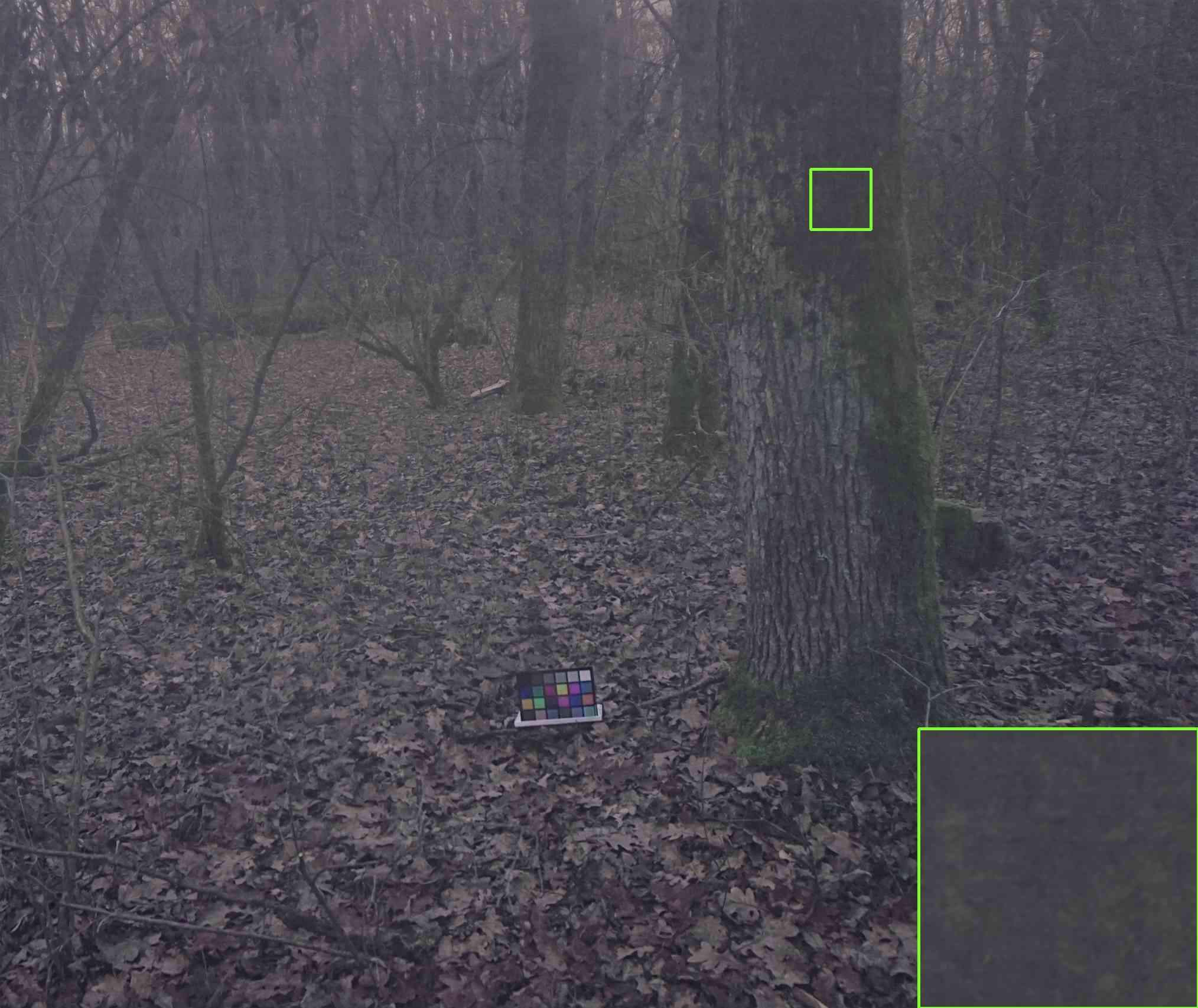}
	\end{minipage}
	\begin{minipage}[h]{0.118\linewidth}
		\centering
		\includegraphics[width=\linewidth]{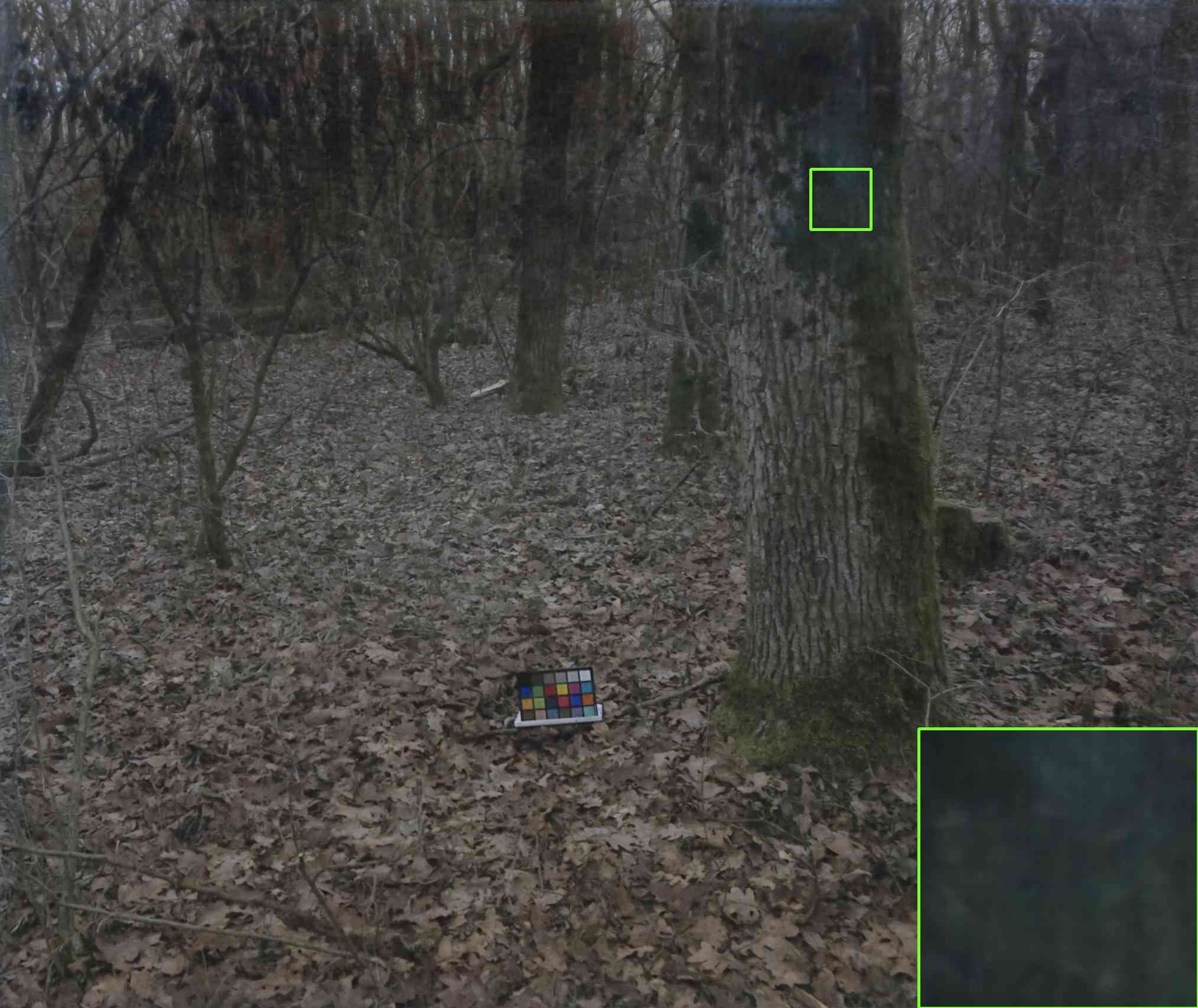}
	\end{minipage}
	\begin{minipage}[h]{0.118\linewidth}
		\centering
		\includegraphics[width=\linewidth]{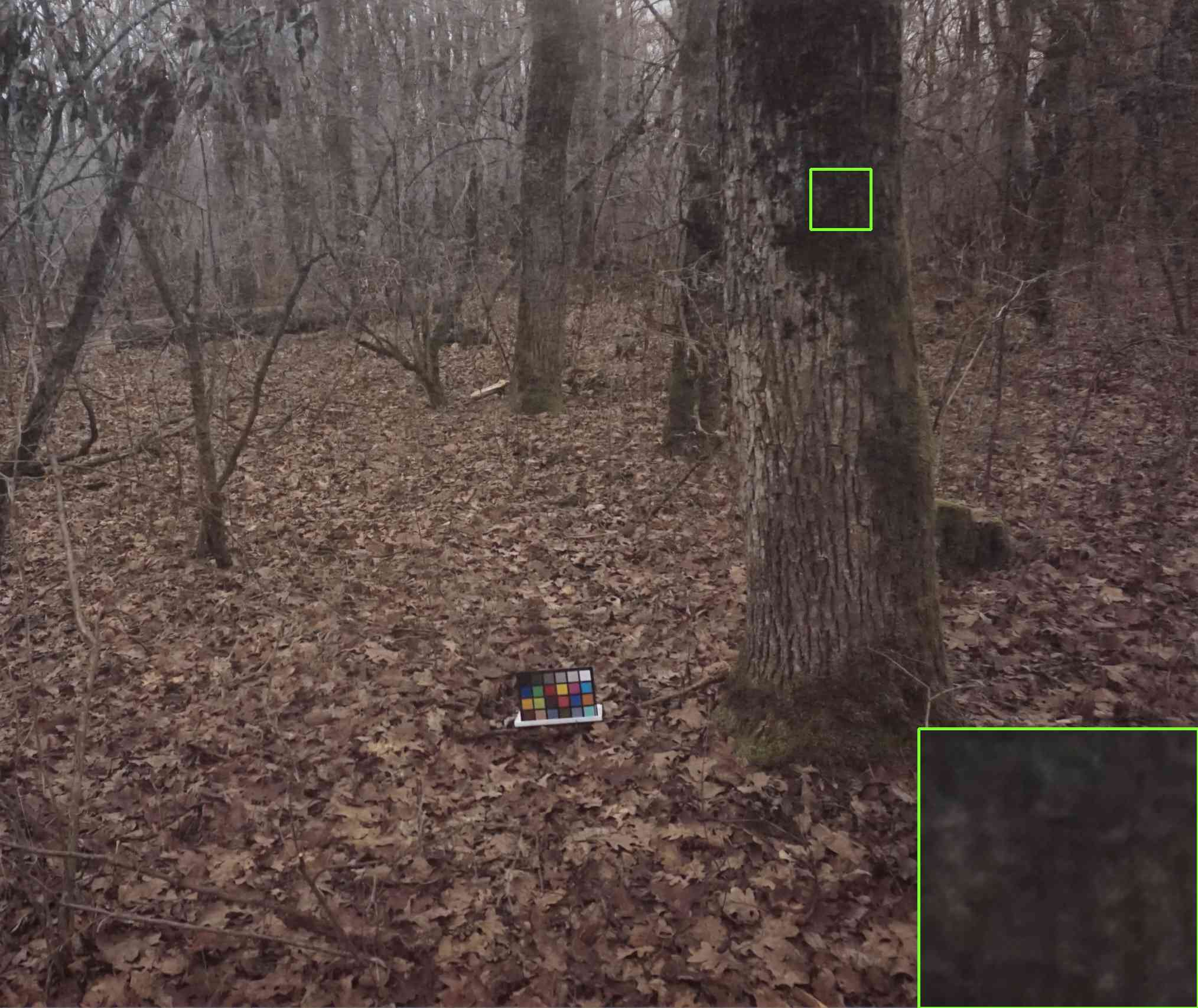}
	\end{minipage}
	\begin{minipage}[h]{0.118\linewidth}
		\centering
		\includegraphics[width=\linewidth]{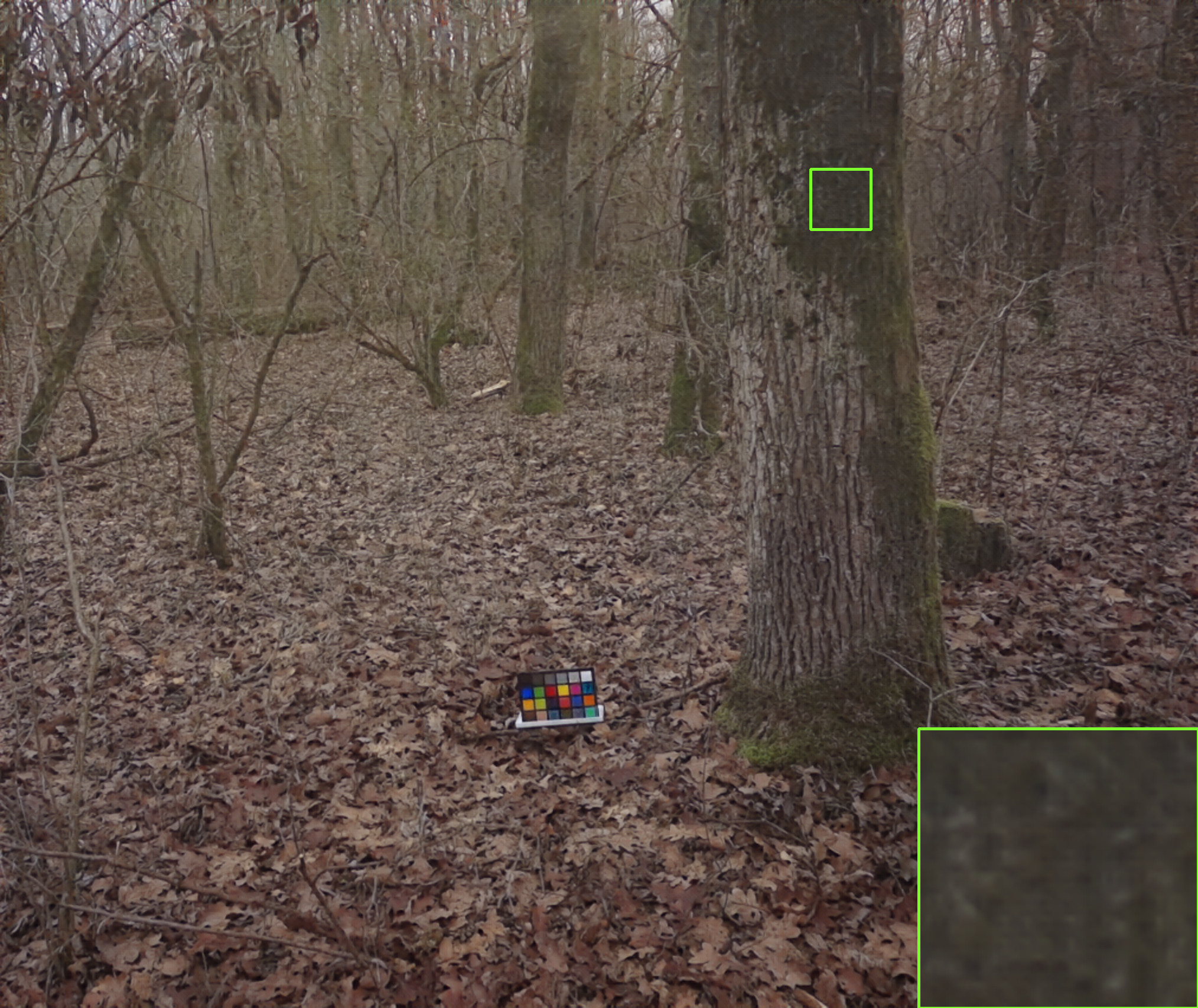}
	\end{minipage}
	\begin{minipage}[h]{0.118\linewidth}
		\centering
		\includegraphics[width=\linewidth]{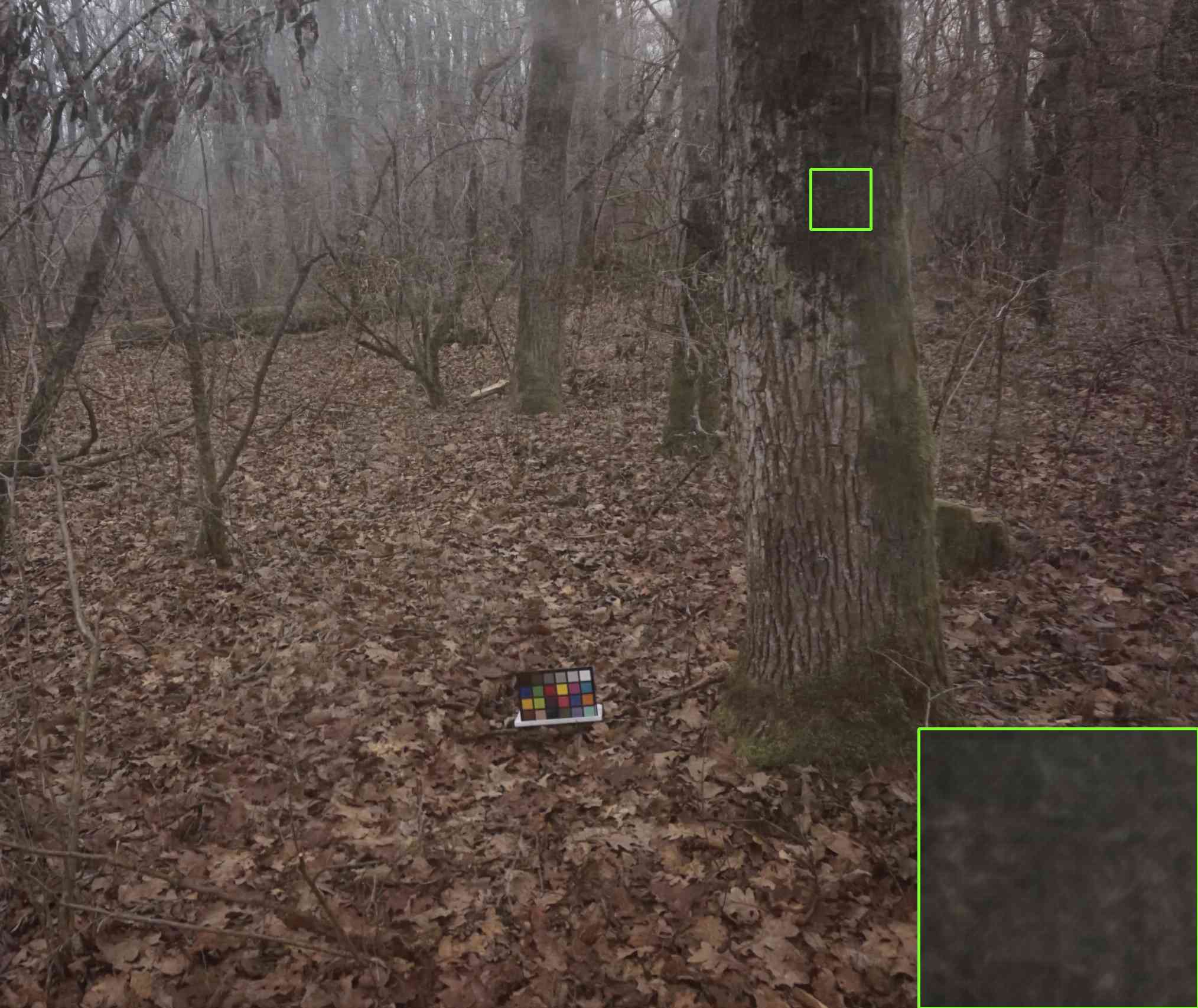}
	\end{minipage}
	\begin{minipage}[h]{0.118\linewidth}
		\centering
		\includegraphics[width=\linewidth]{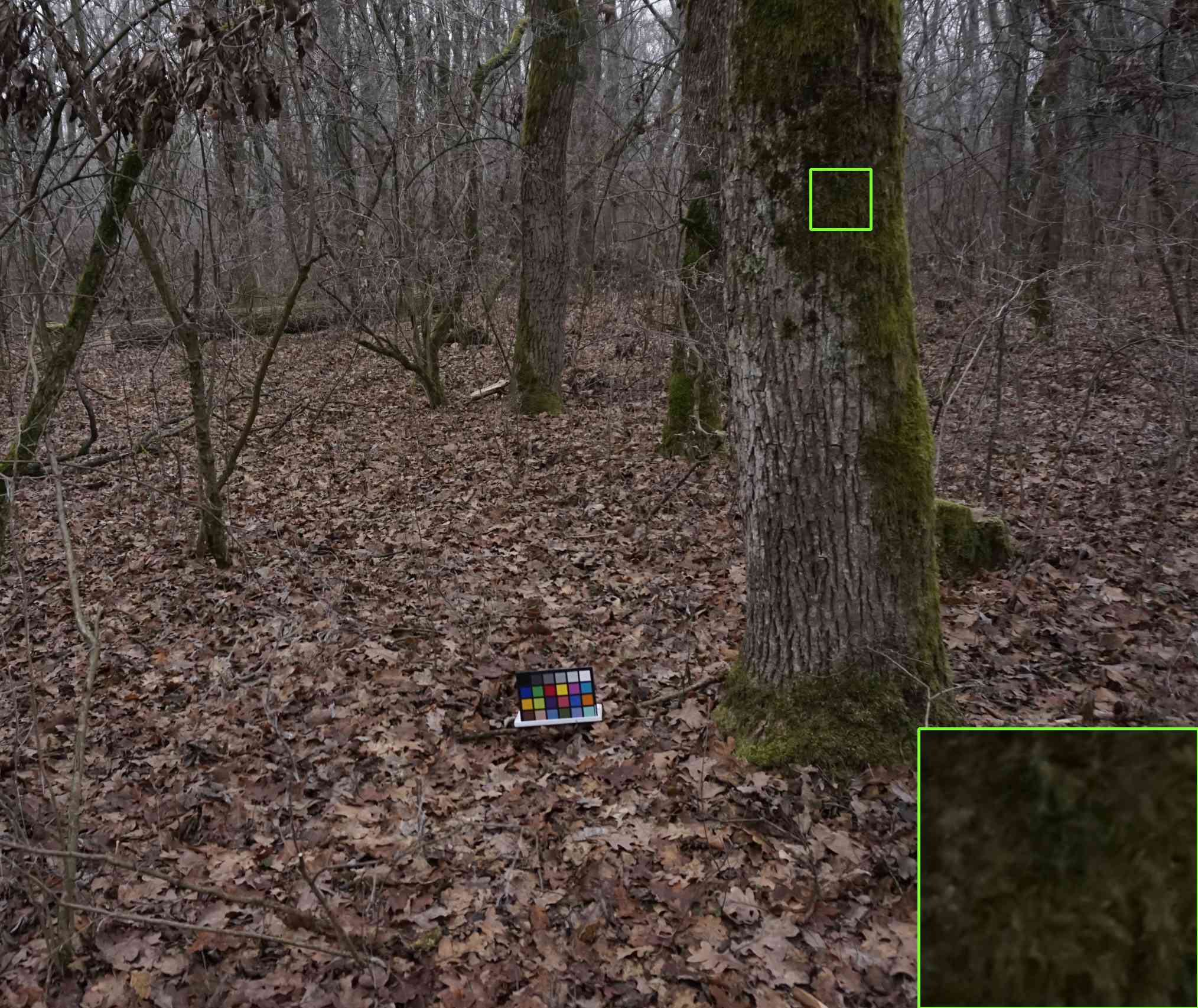}
	\end{minipage}
	\vspace{1mm}
	% O-HAZE 2
	\begin{minipage}[h]{0.118\linewidth}
		\centering
		\includegraphics[width=\linewidth]{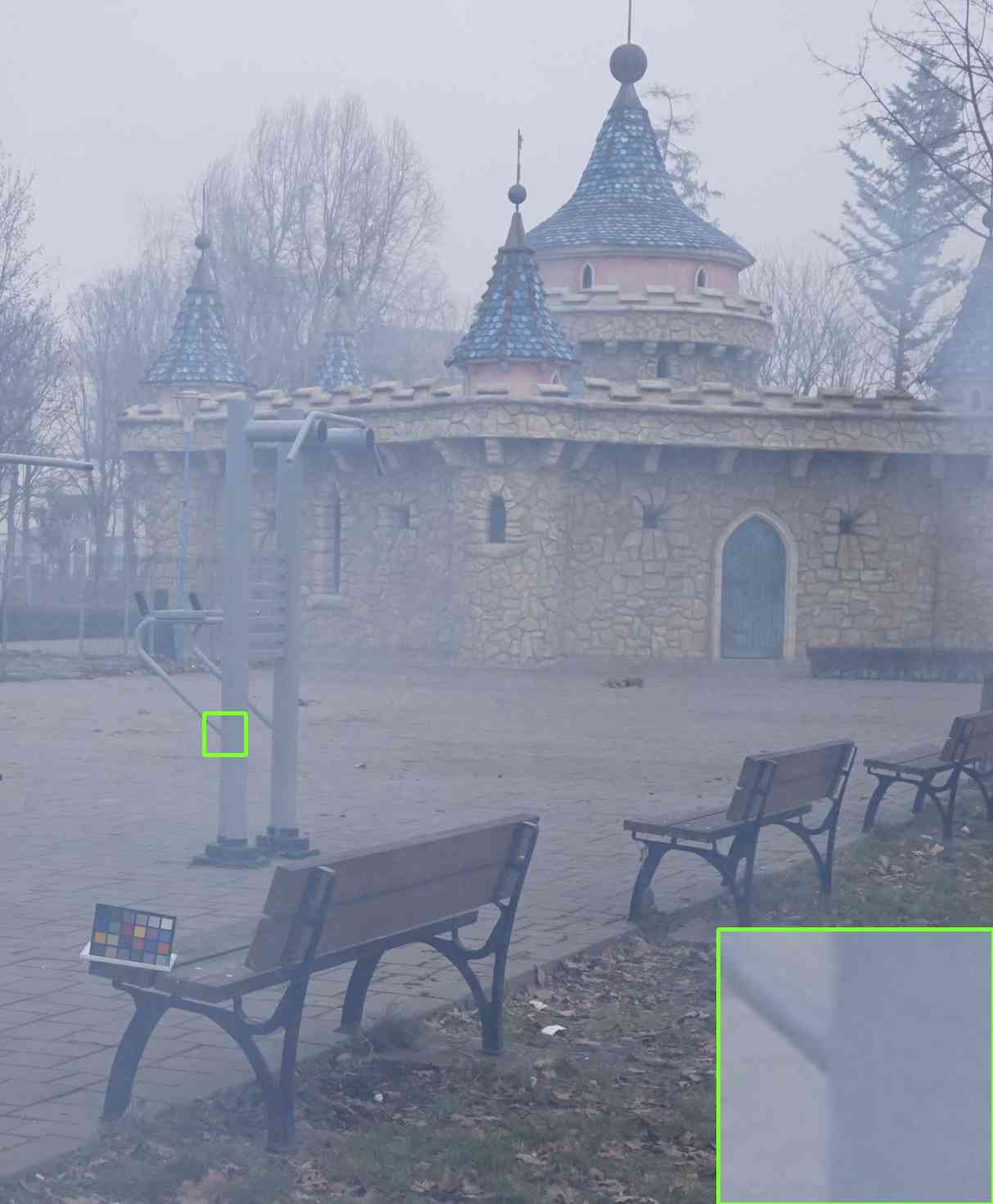}
		\scriptsize{(a) Hazy inputs}
	\end{minipage}
%	\begin{minipage}[h]{0.118\linewidth}
%		\centering
%		\includegraphics[width=\linewidth]{images/syn/ohaze/2/AOD_41_outdoor.jpg}
%		\scriptsize{(b) AOD-Net}
%	\end{minipage}
	\begin{minipage}[h]{0.118\linewidth}
		\centering
		\includegraphics[width=\linewidth]{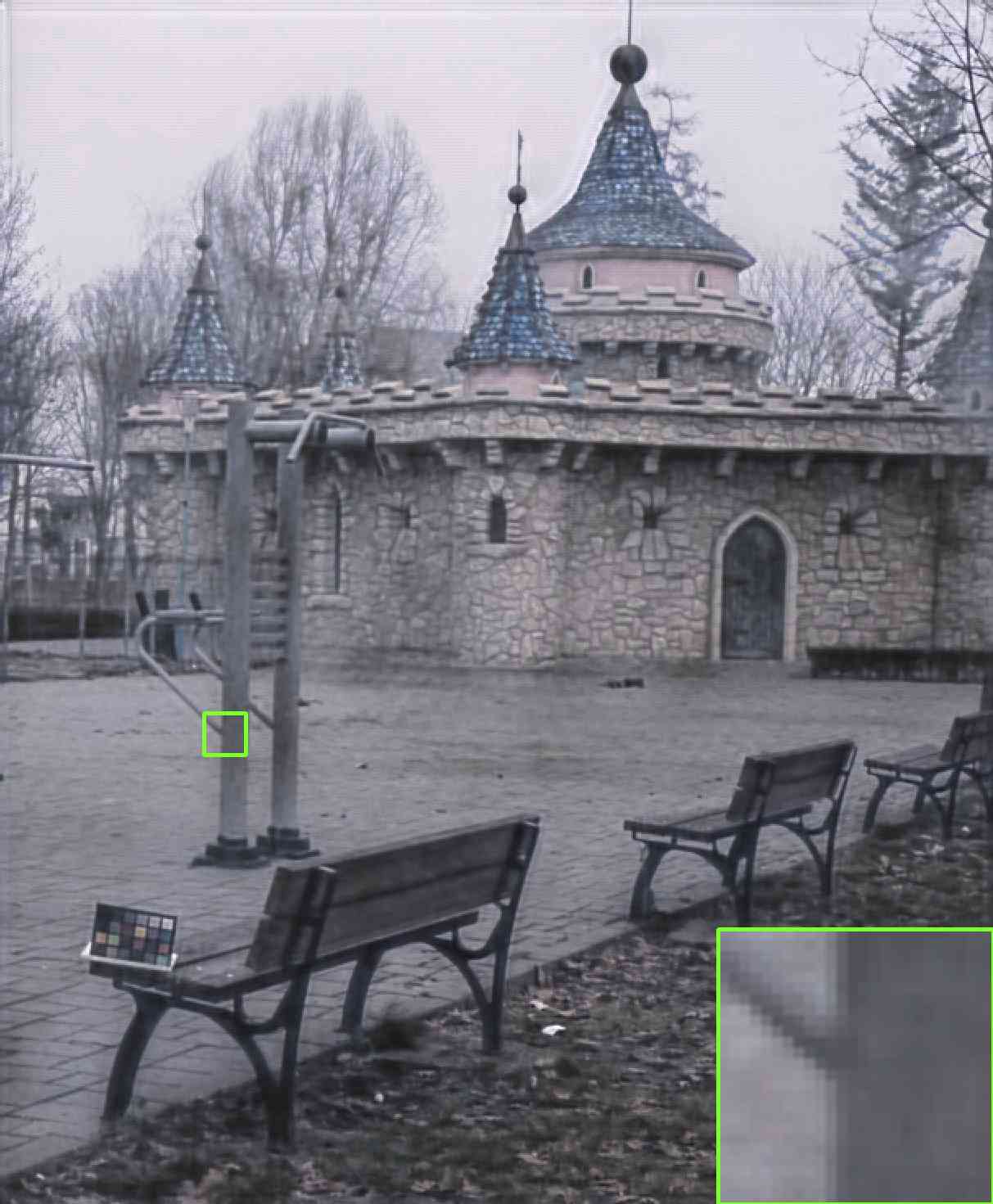}
		\scriptsize{(b) GFN}
	\end{minipage}
	\begin{minipage}[h]{0.118\linewidth}
		\centering
		\includegraphics[width=\linewidth]{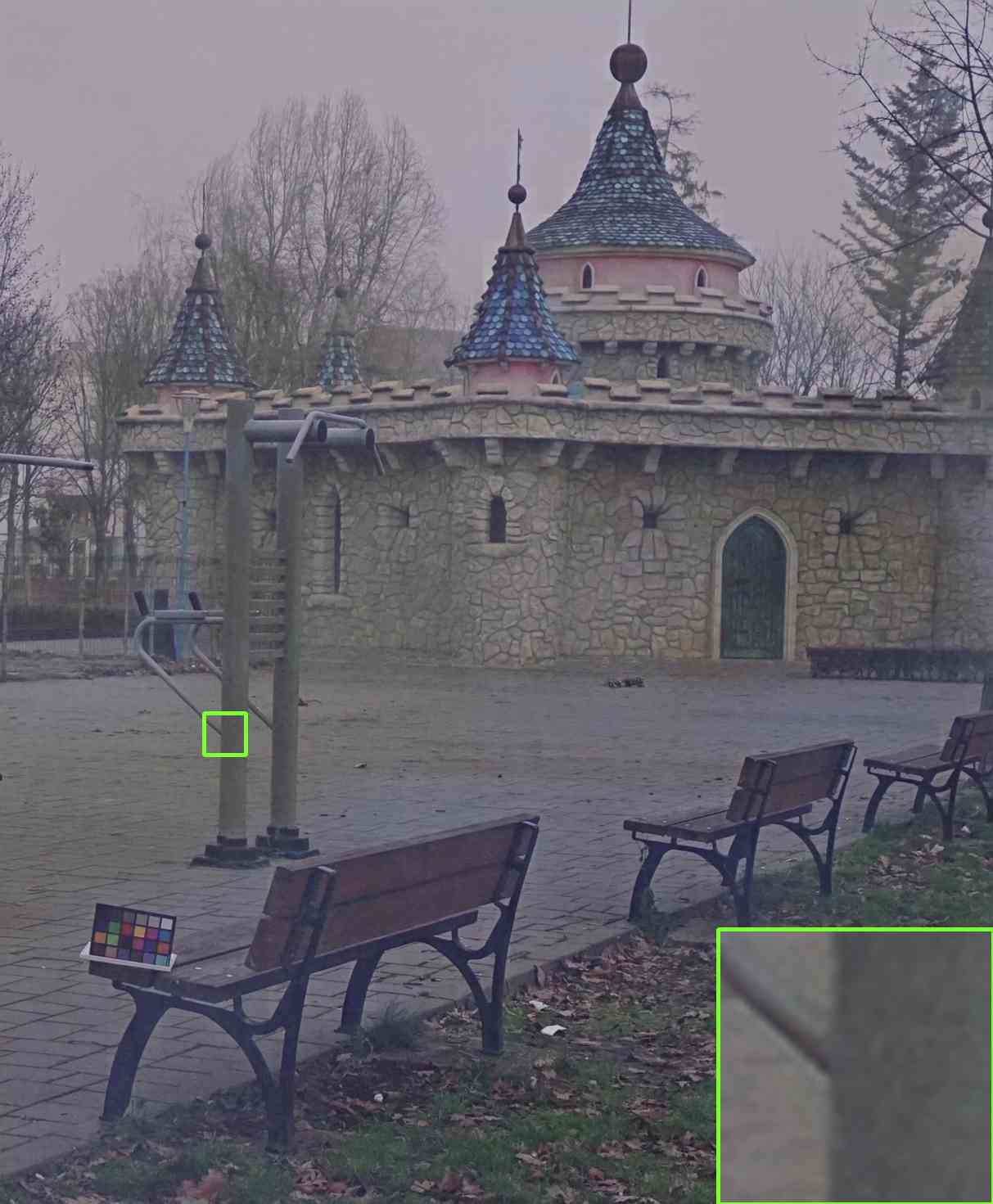}
		\scriptsize{(c) EPDN}
	\end{minipage}
	\begin{minipage}[h]{0.118\linewidth}
		\centering
		\includegraphics[width=\linewidth]{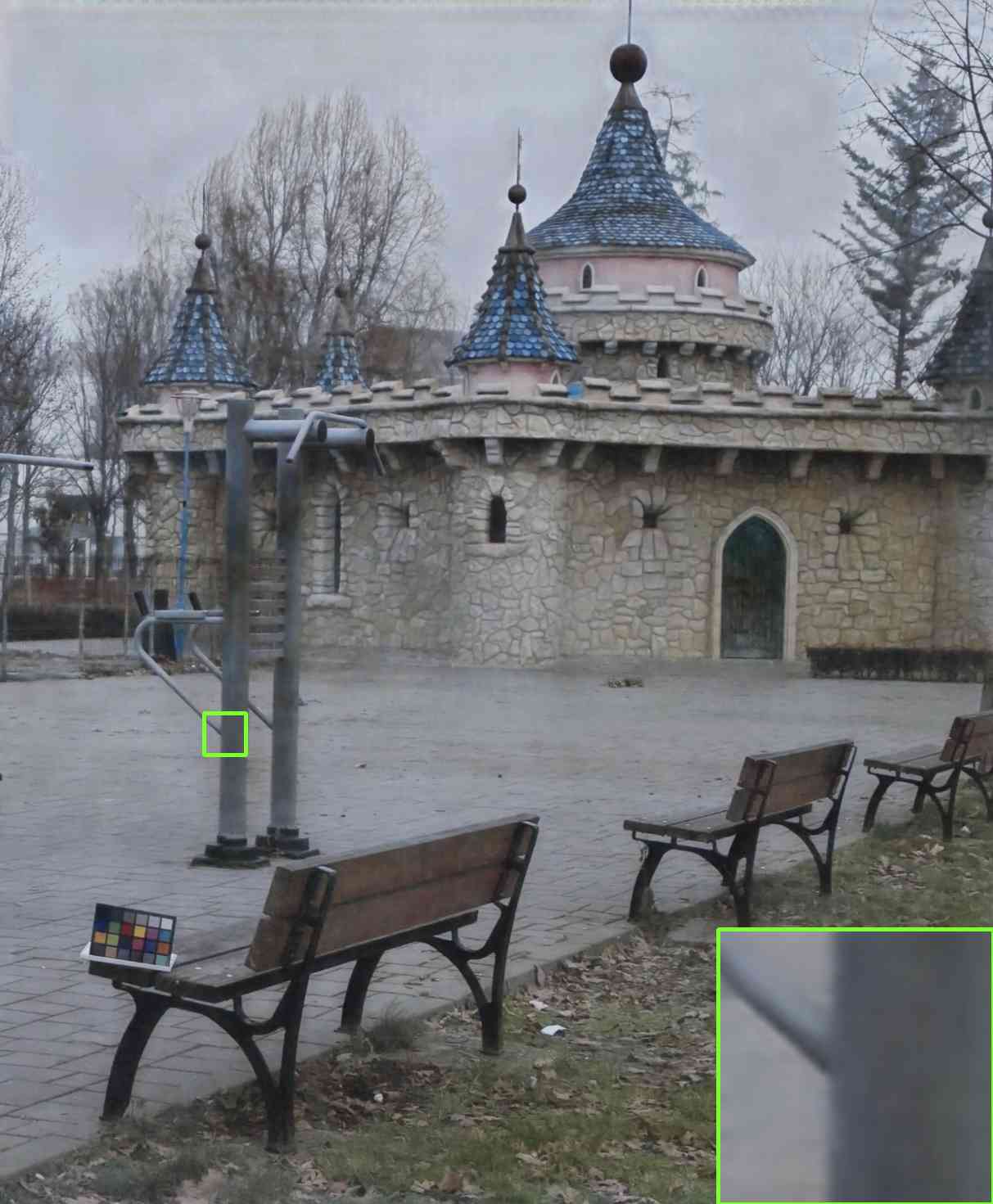}
		\scriptsize{(d) DADN}
	\end{minipage}
	\begin{minipage}[h]{0.118\linewidth}
		\centering
		\includegraphics[width=\linewidth]{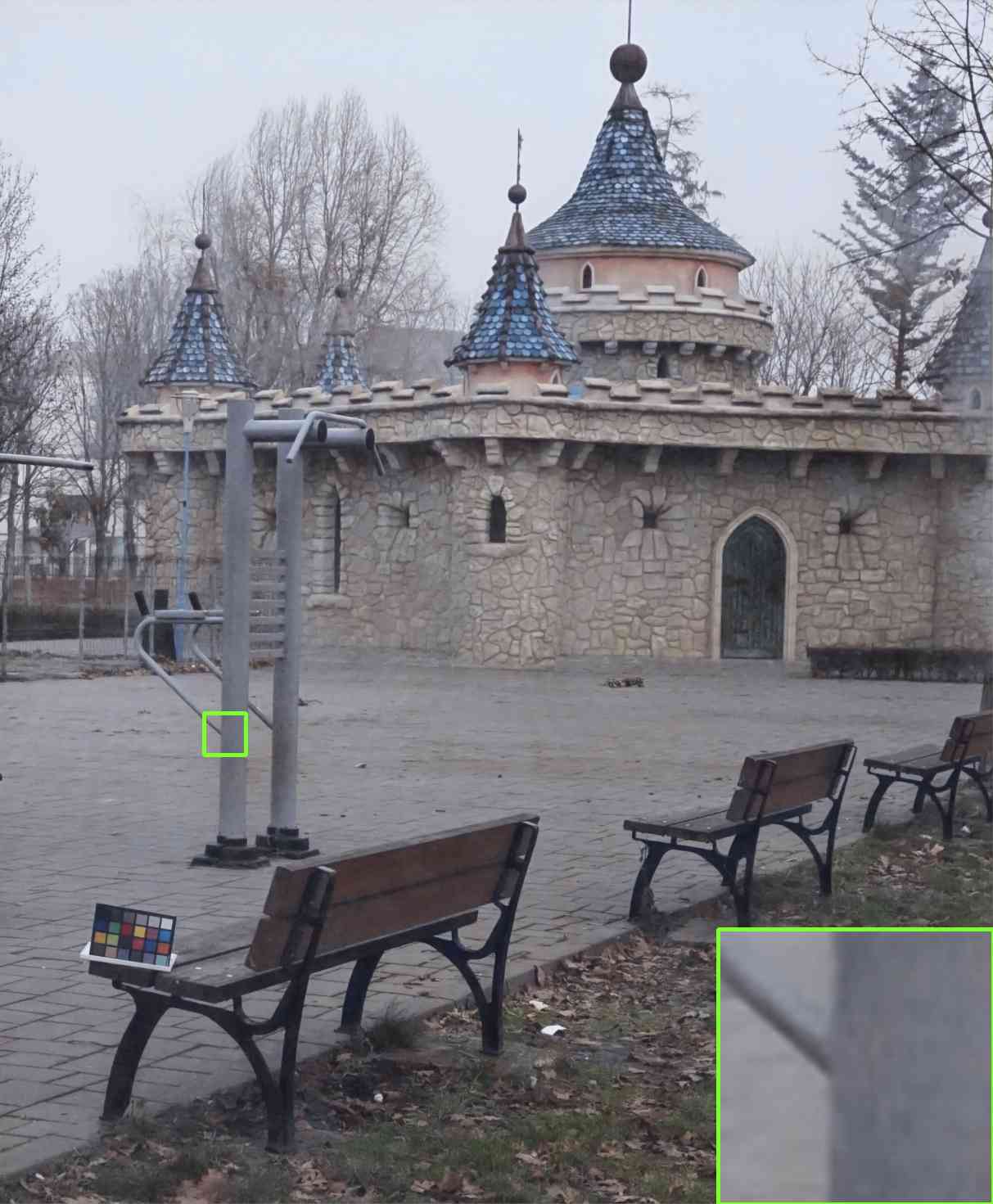}
		\scriptsize{(e) KDDN}
	\end{minipage}
	\begin{minipage}[h]{0.118\linewidth}
		\centering
		\includegraphics[width=\linewidth]{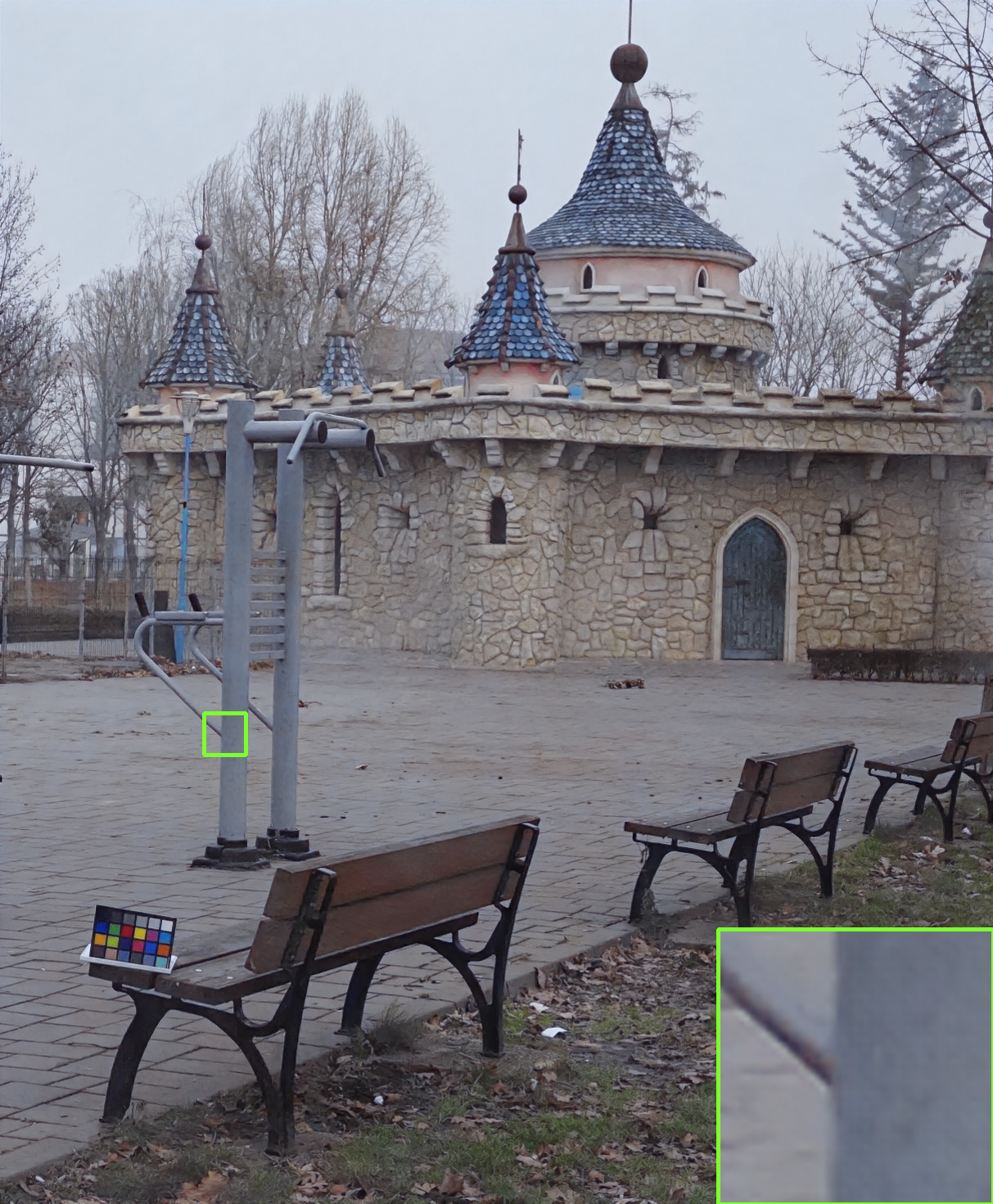}
		\scriptsize{(f) ACER-Net}
	\end{minipage}
	\begin{minipage}[h]{0.118\linewidth}
		\centering
		\includegraphics[width=\linewidth]{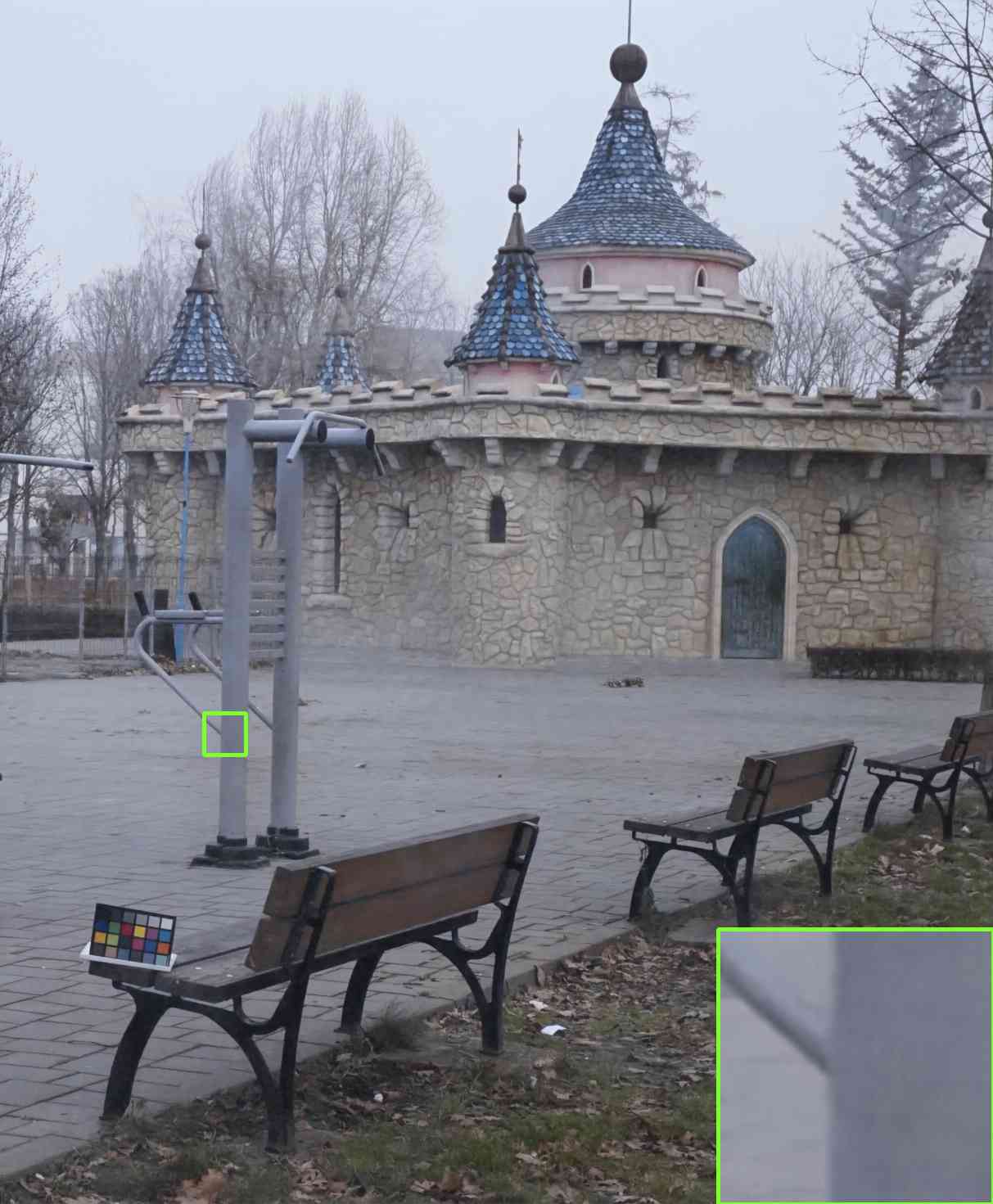}
		\scriptsize{(g) GDN+}
	\end{minipage}
	\begin{minipage}[h]{0.118\linewidth}
		\centering
		\includegraphics[width=\linewidth]{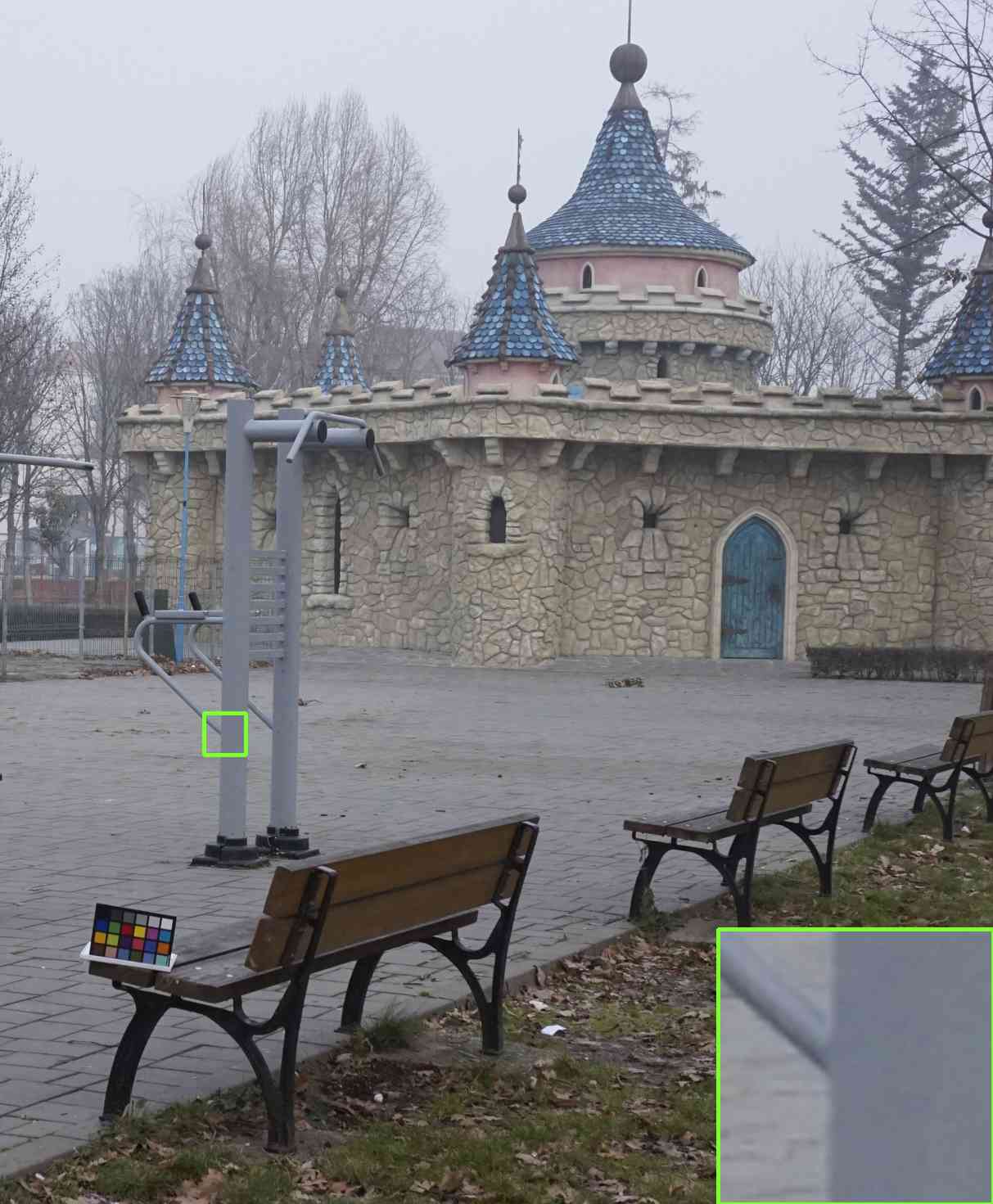}
		\scriptsize{(h) Ground truth}
	\end{minipage}
	\caption{\label{fig:synthetic} Visual comparisons of different methods on synthetic datasets: SOTS, Middlebury, HazeRD, and O-HAZE (from top to bottom with two rows for each dataset). Zoom in for details.}
\end{figure*}

\begin{figure*}[t]
	\centering
	% 37real 1
	\begin{minipage}[h]{0.118\linewidth}
		\centering
		\includegraphics[width=\linewidth]{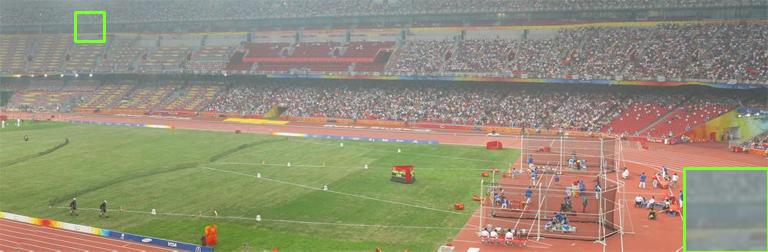}
	\end{minipage}
%	\begin{minipage}[h]{0.118\linewidth}
%		\centering
%		\includegraphics[width=\linewidth]{images/real/37real/1/AOD_stadium.jpg}
%	\end{minipage}
	\begin{minipage}[h]{0.118\linewidth}
		\centering
		\includegraphics[width=\linewidth]{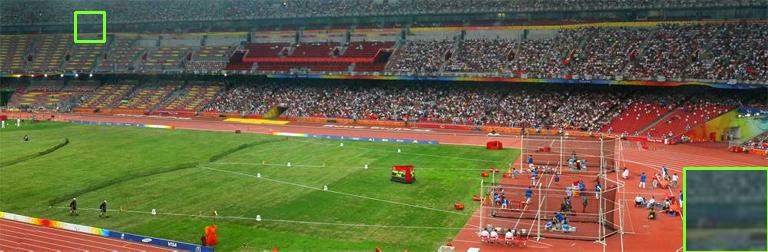}
	\end{minipage}
	\begin{minipage}[h]{0.118\linewidth}
		\centering
		\includegraphics[width=\linewidth]{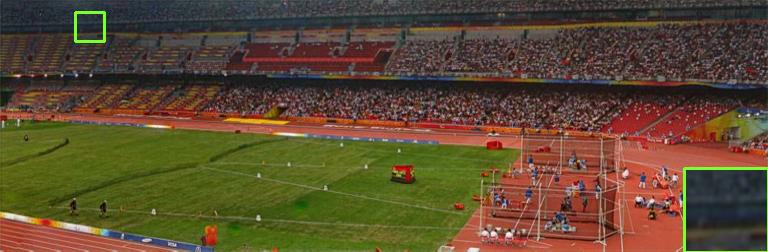}
	\end{minipage}
	\begin{minipage}[h]{0.118\linewidth}
		\centering
		\includegraphics[width=\linewidth]{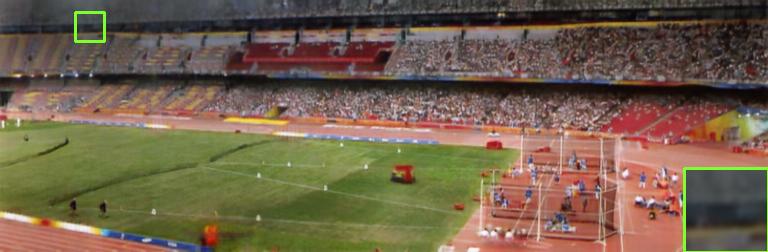}
	\end{minipage}
	\begin{minipage}[h]{0.118\linewidth}
		\centering
		\includegraphics[width=\linewidth]{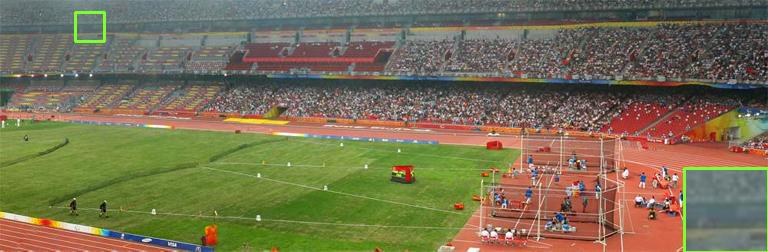}
	\end{minipage}
	\begin{minipage}[h]{0.118\linewidth}
		\centering
		\includegraphics[width=\linewidth]{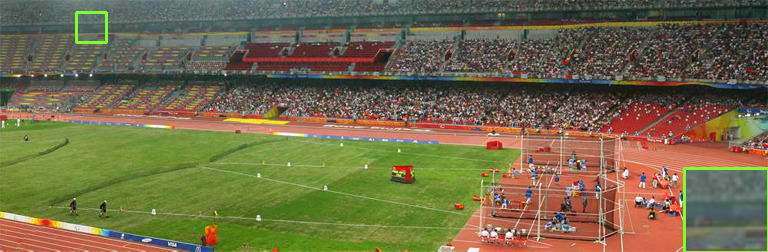}
	\end{minipage}
	\begin{minipage}[h]{0.118\linewidth}
		\centering
		\includegraphics[width=\linewidth]{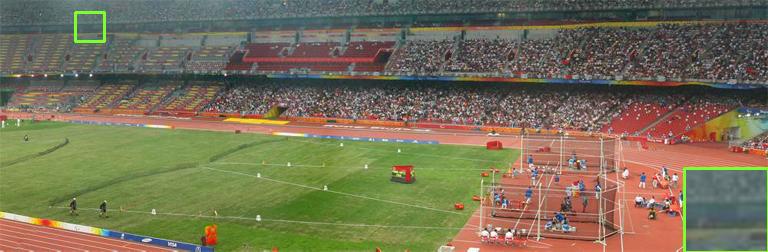}
	\end{minipage}
	\vspace{1mm}
	\begin{minipage}[h]{0.118\linewidth}
		\centering
		\includegraphics[width=\linewidth]{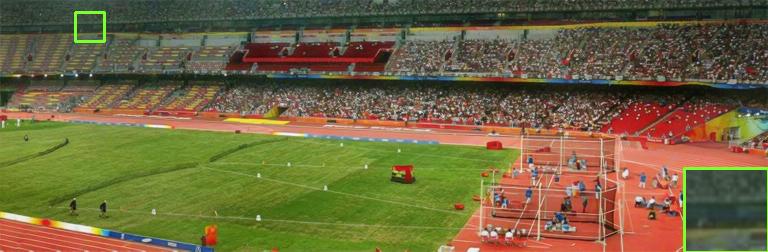}
	\end{minipage}
	\vspace{1mm}
	% 37real 3
	\begin{minipage}[h]{0.118\linewidth}
		\centering
		\includegraphics[width=\linewidth]{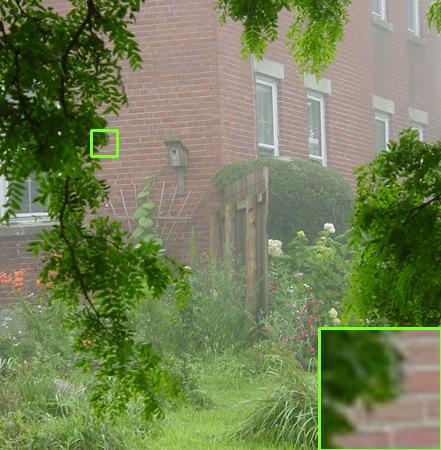}
	\end{minipage}
%	\begin{minipage}[h]{0.118\linewidth}
%		\centering
%		\includegraphics[width=\linewidth]{images/real/37real/3/AOD_trees2.jpg}
%	\end{minipage}
	\begin{minipage}[h]{0.118\linewidth}
		\centering
		\includegraphics[width=\linewidth]{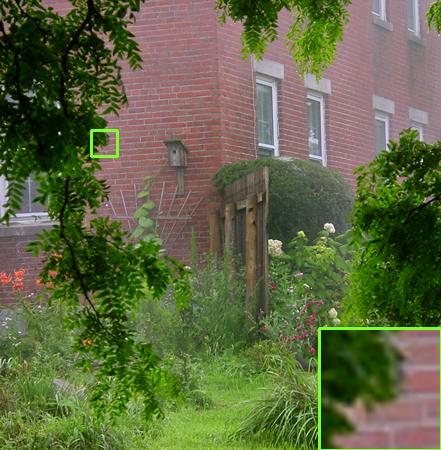}
	\end{minipage}
	\begin{minipage}[h]{0.118\linewidth}
		\centering
		\includegraphics[width=\linewidth]{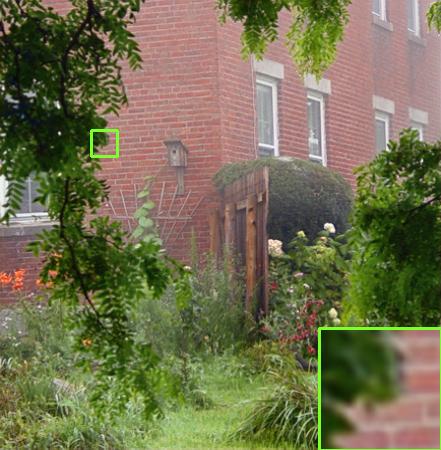}
	\end{minipage}
	\begin{minipage}[h]{0.118\linewidth}
		\centering
		\includegraphics[width=\linewidth]{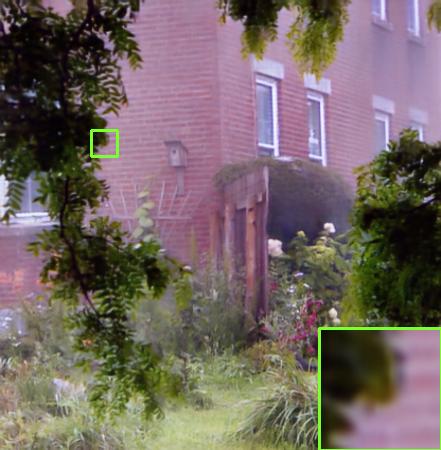}
	\end{minipage}
	\begin{minipage}[h]{0.118\linewidth}
		\centering
		\includegraphics[width=\linewidth]{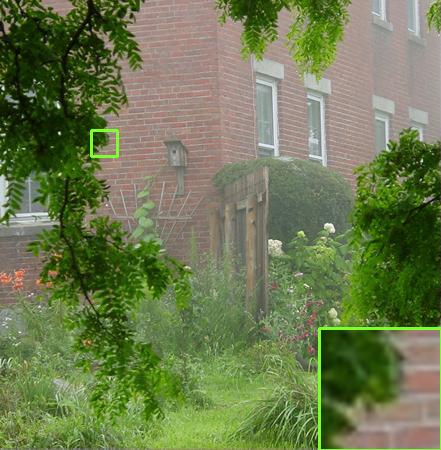}
	\end{minipage}
	\begin{minipage}[h]{0.118\linewidth}
		\centering
		\includegraphics[width=\linewidth]{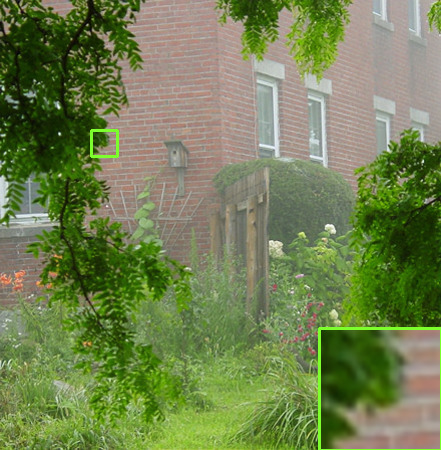}
	\end{minipage}
	\begin{minipage}[h]{0.118\linewidth}
		\centering
		\includegraphics[width=\linewidth]{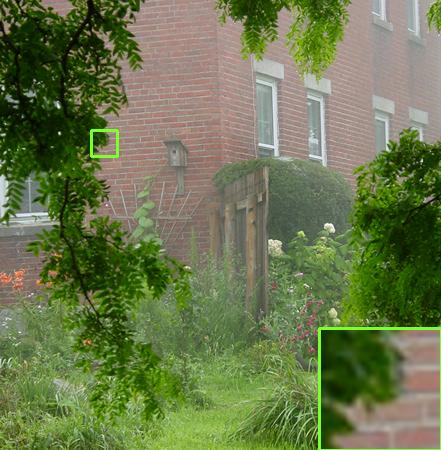}
	\end{minipage}
	\begin{minipage}[h]{0.118\linewidth}
		\centering
		\includegraphics[width=\linewidth]{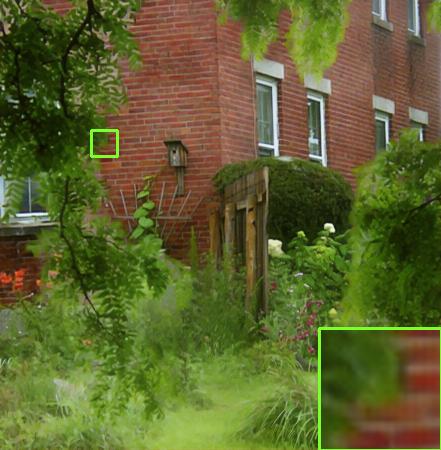}
	\end{minipage}
	\vspace{1mm}
	% 37real 2
	\begin{minipage}[h]{0.118\linewidth}
		\centering
		\includegraphics[width=\linewidth]{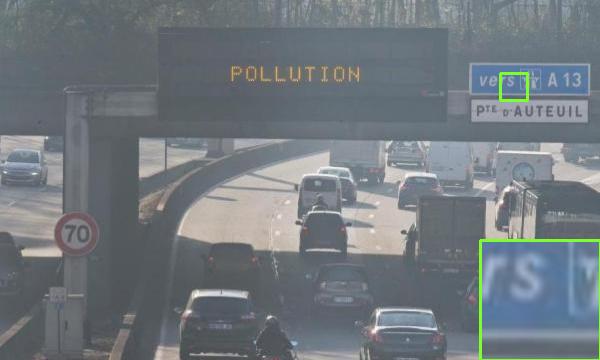}
	\end{minipage}
%	\begin{minipage}[h]{0.118\linewidth}
%		\centering
%		\includegraphics[width=\linewidth]{images/real/URHI/6/AOD_BL_Google_121.jpg}
%	\end{minipage}
	\begin{minipage}[h]{0.118\linewidth}
		\centering
		\includegraphics[width=\linewidth]{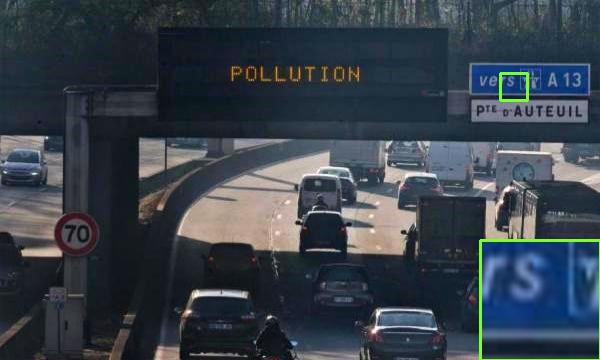}
	\end{minipage}
	\begin{minipage}[h]{0.118\linewidth}
		\centering
		\includegraphics[width=\linewidth]{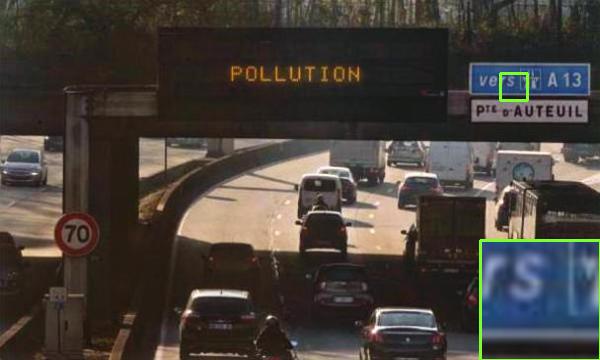}
	\end{minipage}
	\begin{minipage}[h]{0.118\linewidth}
		\centering
		\includegraphics[width=\linewidth]{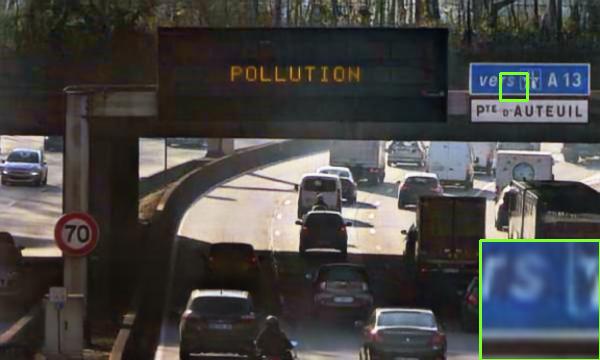}
	\end{minipage}
	\begin{minipage}[h]{0.118\linewidth}
		\centering
		\includegraphics[width=\linewidth]{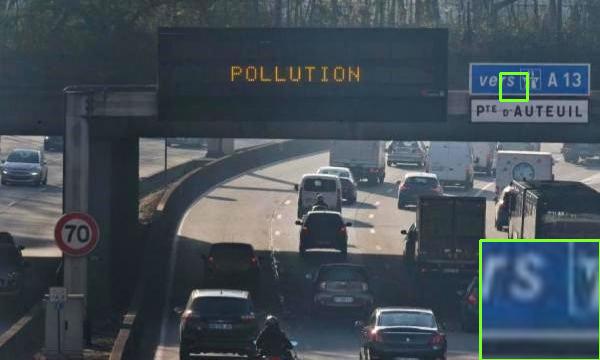}
	\end{minipage}
	\begin{minipage}[h]{0.118\linewidth}
		\centering
		\includegraphics[width=\linewidth]{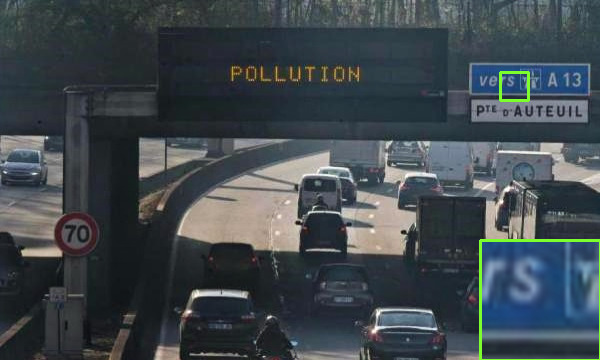}
	\end{minipage}
	\begin{minipage}[h]{0.118\linewidth}
		\centering
		\includegraphics[width=\linewidth]{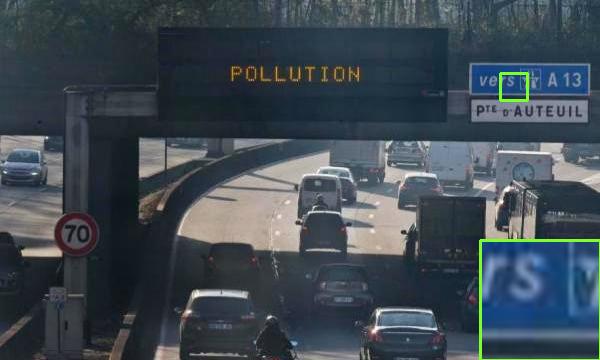}
	\end{minipage}
	\begin{minipage}[h]{0.118\linewidth}
		\centering
		\includegraphics[width=\linewidth]{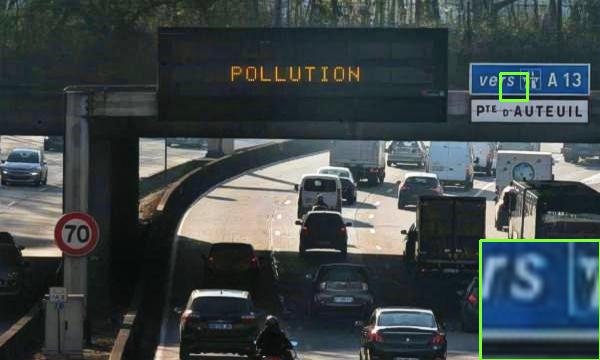}
	\end{minipage}
	\vspace{1mm}
	% URHI 1
	\begin{minipage}[h]{0.118\linewidth}
		\centering
		\includegraphics[width=\linewidth]{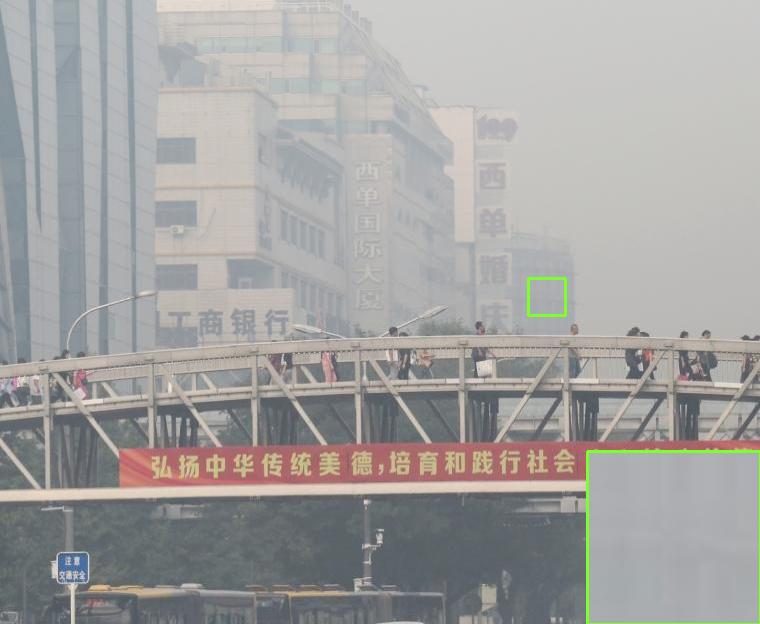}
	\end{minipage}
%	\begin{minipage}[h]{0.118\linewidth}
%		\centering
%		\includegraphics[width=\linewidth]{images/real/URHI/1/AOD_BJ_Baidu_606.jpg}
%	\end{minipage}
	\begin{minipage}[h]{0.118\linewidth}
		\centering
		\includegraphics[width=\linewidth]{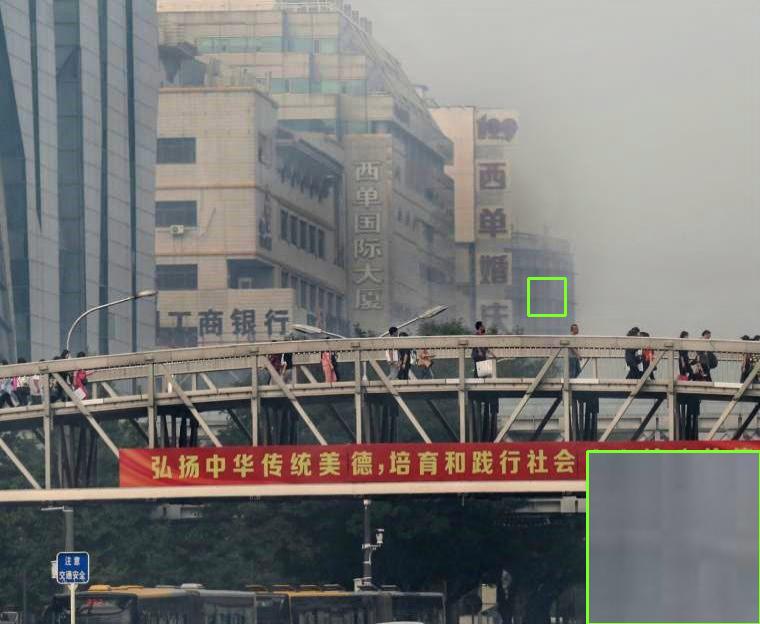}
	\end{minipage}
	\begin{minipage}[h]{0.118\linewidth}
		\centering
		\includegraphics[width=\linewidth]{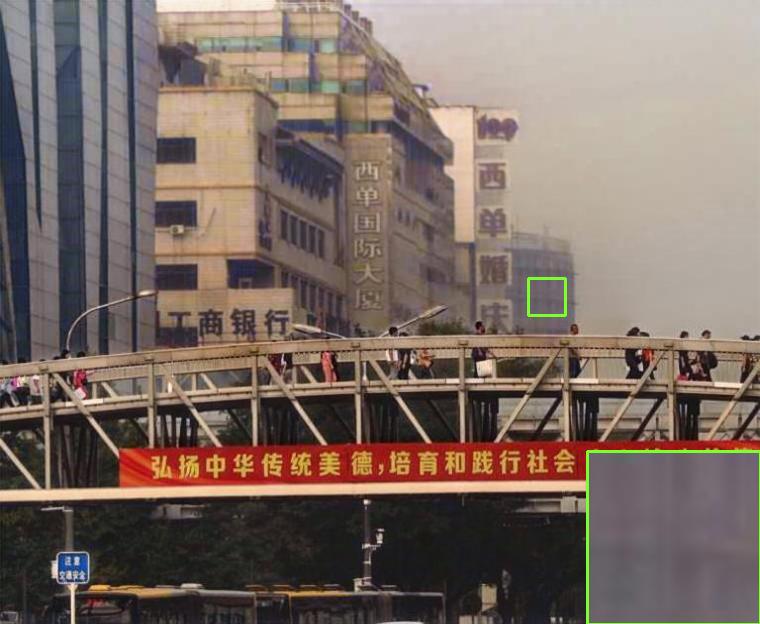}
	\end{minipage}
	\begin{minipage}[h]{0.118\linewidth}
		\centering
		\includegraphics[width=\linewidth]{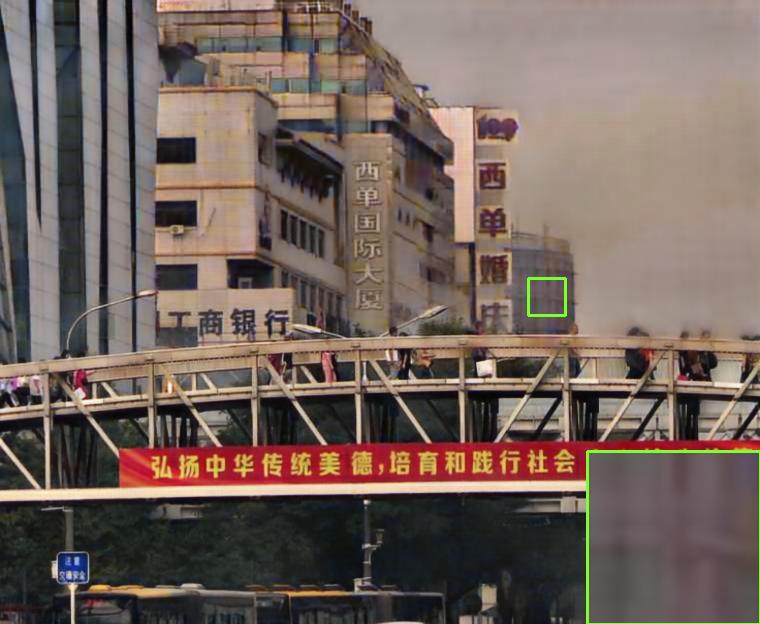}
	\end{minipage}
	\begin{minipage}[h]{0.118\linewidth}
		\centering
		\includegraphics[width=\linewidth]{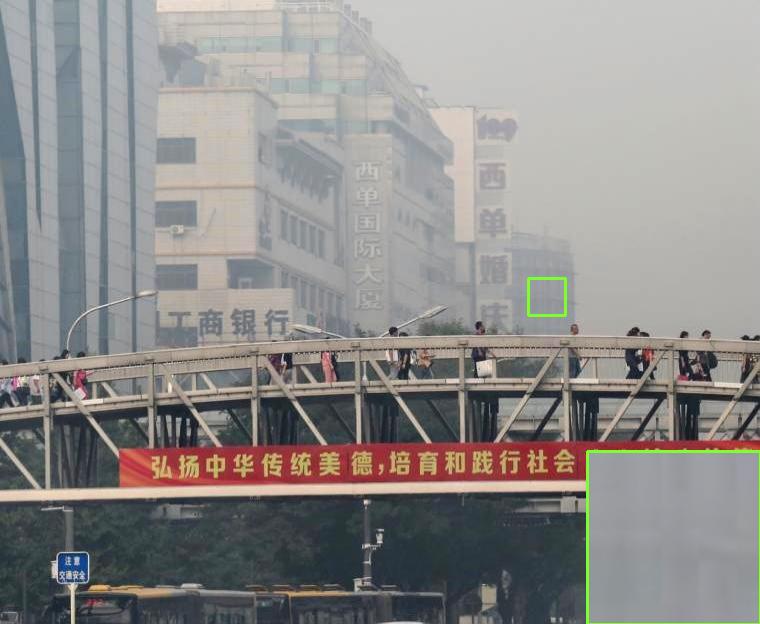}
	\end{minipage}
	\begin{minipage}[h]{0.118\linewidth}
		\centering
		\includegraphics[width=\linewidth]{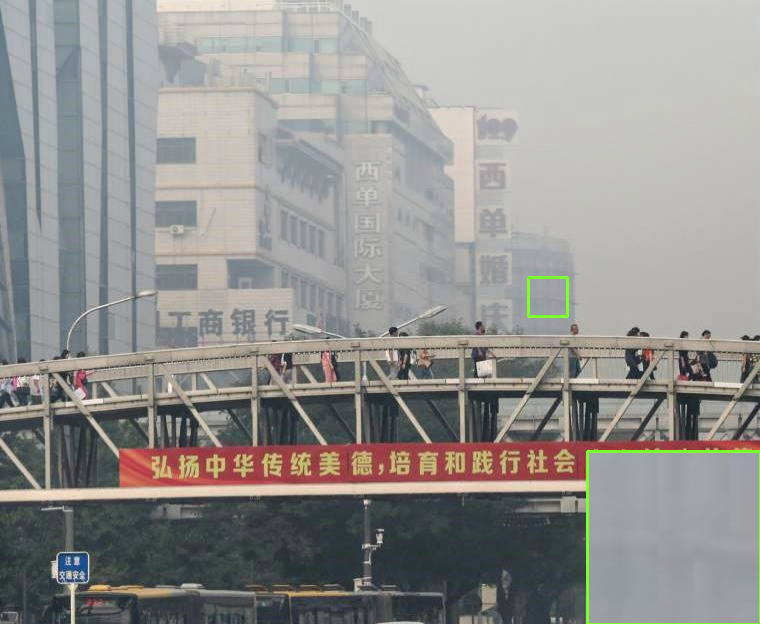}
	\end{minipage}
	\begin{minipage}[h]{0.118\linewidth}
		\centering
		\includegraphics[width=\linewidth]{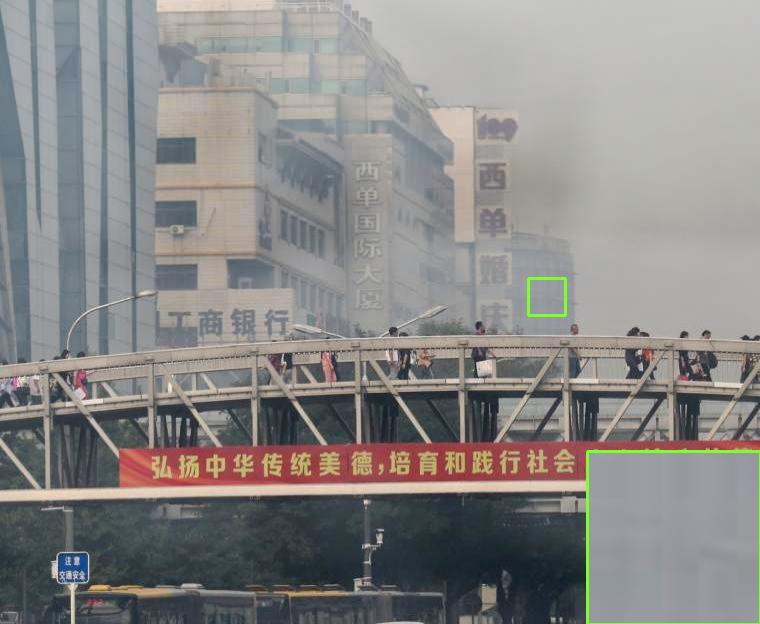}
	\end{minipage}
	\begin{minipage}[h]{0.118\linewidth}
		\centering
		\includegraphics[width=\linewidth]{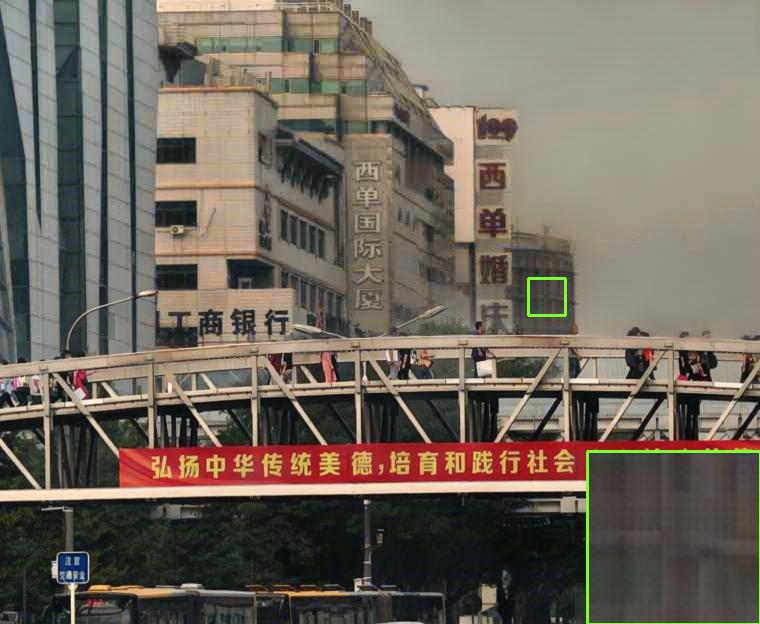}
	\end{minipage}
	\vspace{1mm}
	% URHI 2
	\begin{minipage}[h]{0.118\linewidth}
		\centering
		\includegraphics[width=\linewidth]{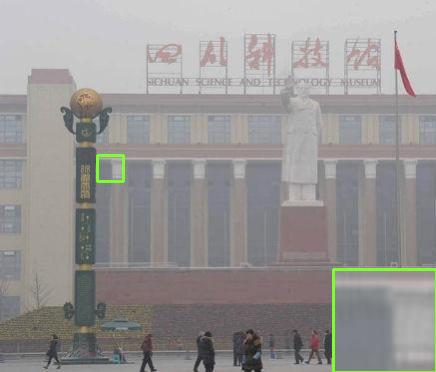}
	\end{minipage}
%	\begin{minipage}[h]{0.118\linewidth}
%		\centering
%		\includegraphics[width=\linewidth]{images/real/URHI/2/AOD_CD_Google_093.jpg}
%	\end{minipage}
	\begin{minipage}[h]{0.118\linewidth}
		\centering
		\includegraphics[width=\linewidth]{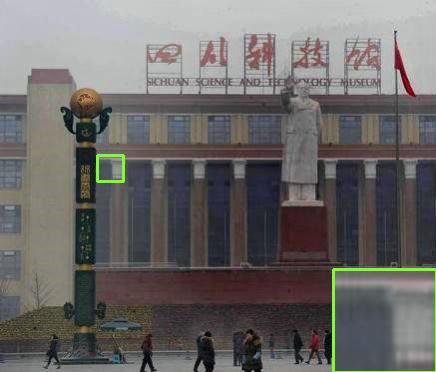}
	\end{minipage}
	\begin{minipage}[h]{0.118\linewidth}
		\centering
		\includegraphics[width=\linewidth]{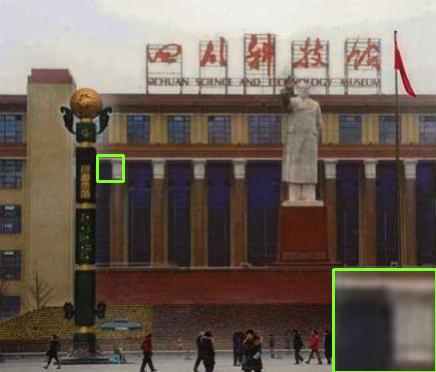}
	\end{minipage}
	\begin{minipage}[h]{0.118\linewidth}
		\centering
		\includegraphics[width=\linewidth]{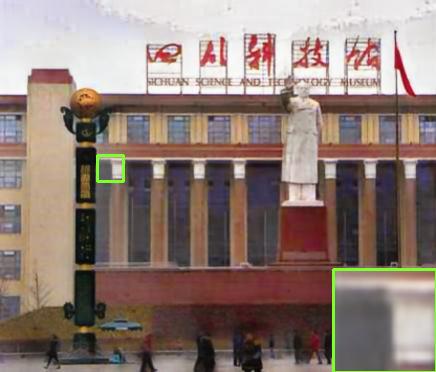}
	\end{minipage}
	\begin{minipage}[h]{0.118\linewidth}
		\centering
		\includegraphics[width=\linewidth]{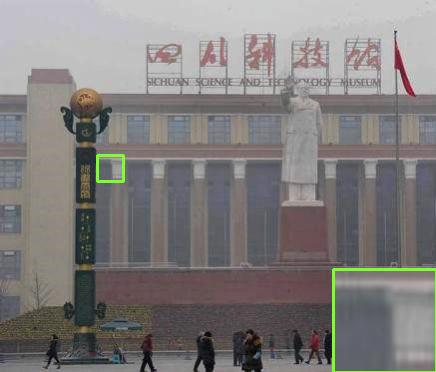}
	\end{minipage}
	\begin{minipage}[h]{0.118\linewidth}
		\centering
		\includegraphics[width=\linewidth]{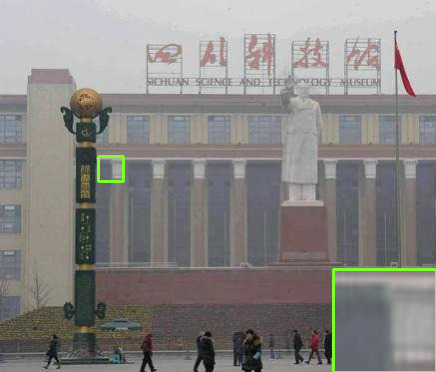}
	\end{minipage}
	\begin{minipage}[h]{0.118\linewidth}
		\centering
		\includegraphics[width=\linewidth]{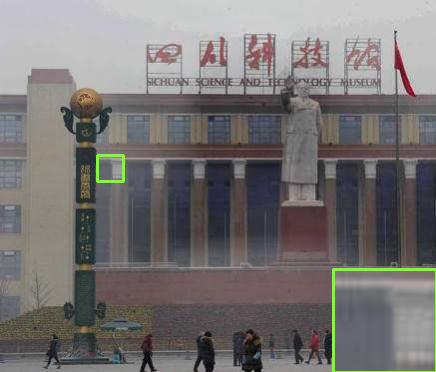}
	\end{minipage}
	\begin{minipage}[h]{0.118\linewidth}
		\centering
		\includegraphics[width=\linewidth]{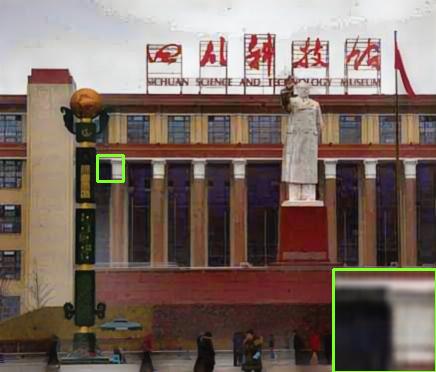}
	\end{minipage}
	\vspace{1mm}
	% URHI 3
	\begin{minipage}[h]{0.118\linewidth}
		\centering
		\includegraphics[width=\linewidth]{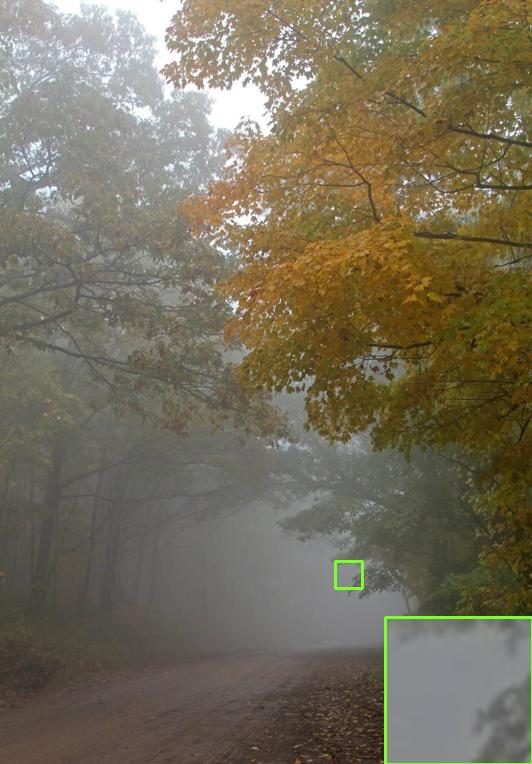}
	\end{minipage}
	\begin{minipage}[h]{0.118\linewidth}
		\centering
		\includegraphics[width=\linewidth]{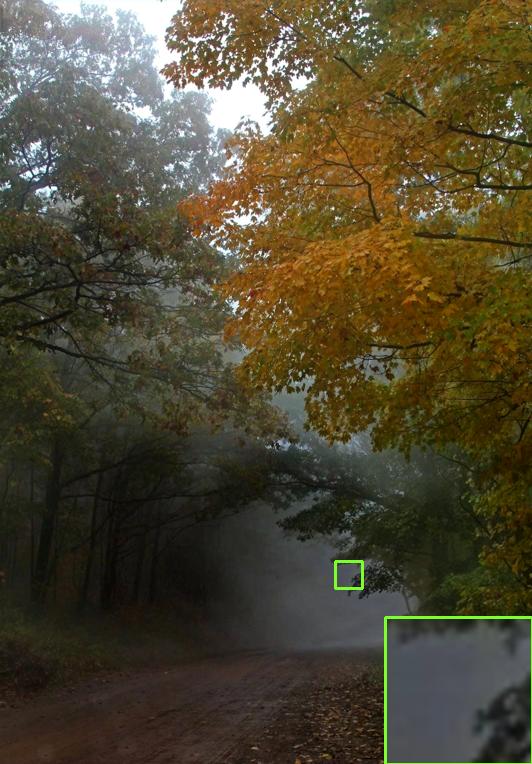}
	\end{minipage}
	\begin{minipage}[h]{0.118\linewidth}
		\centering
		\includegraphics[width=\linewidth]{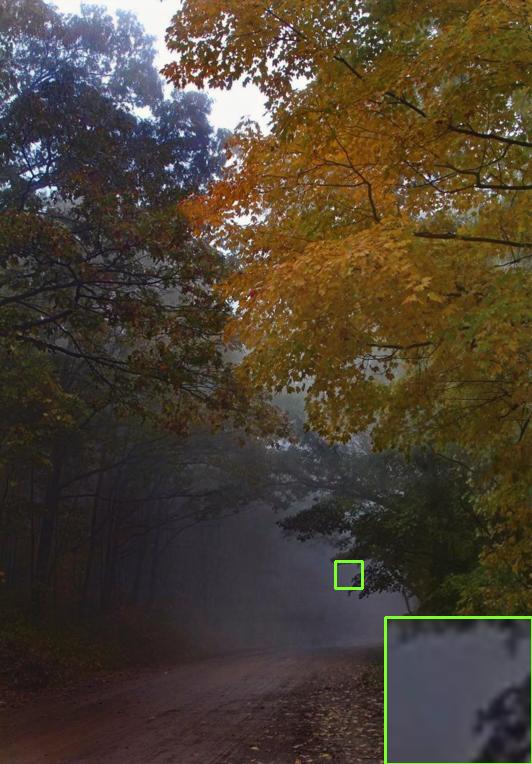}
	\end{minipage}
	\begin{minipage}[h]{0.118\linewidth}
		\centering
		\includegraphics[width=\linewidth]{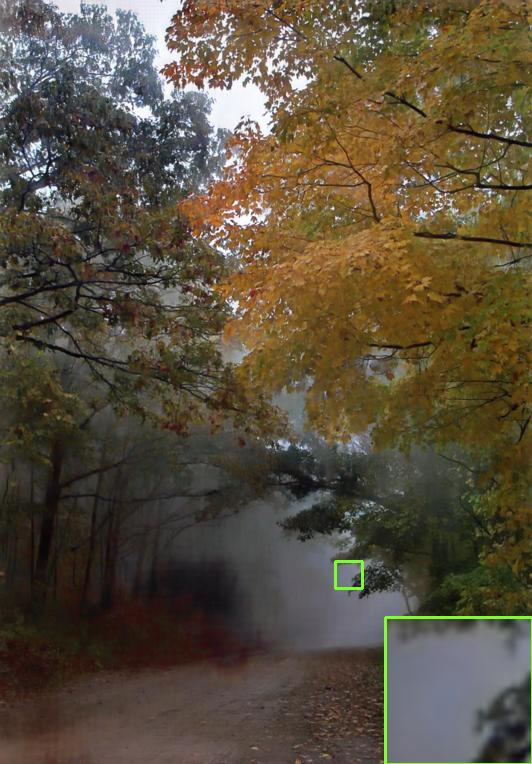}
	\end{minipage}
	\begin{minipage}[h]{0.118\linewidth}
		\centering
		\includegraphics[width=\linewidth]{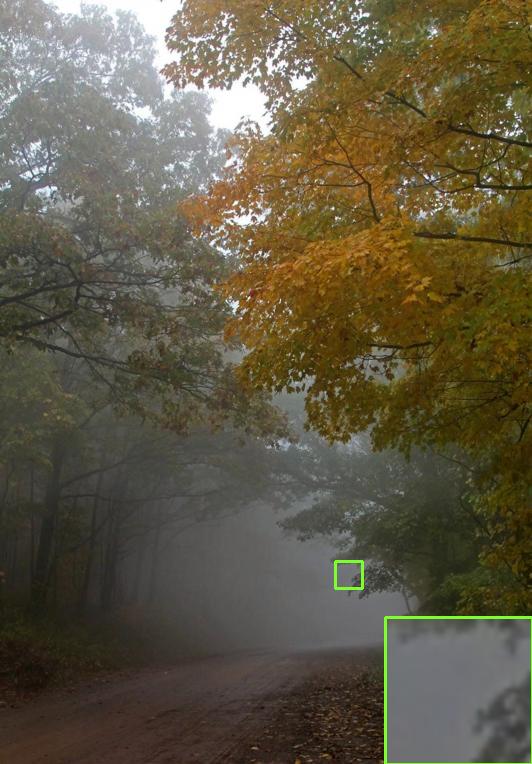}
	\end{minipage}
	\begin{minipage}[h]{0.118\linewidth}
		\centering
		\includegraphics[width=\linewidth]{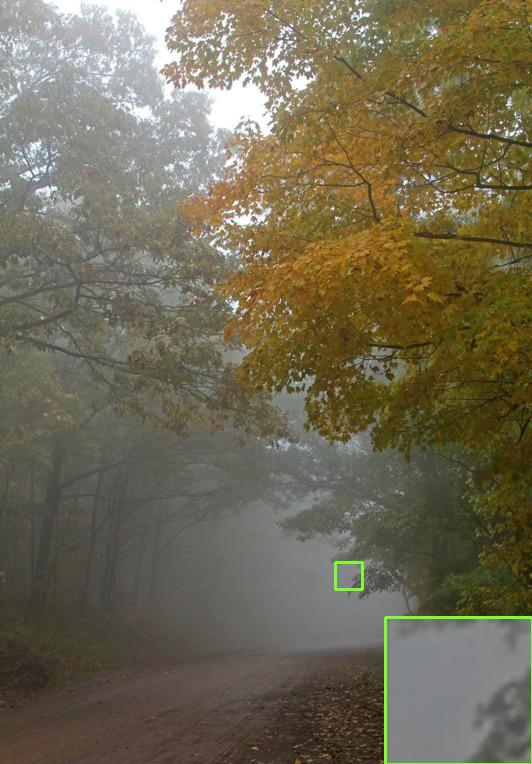}
	\end{minipage}
	\begin{minipage}[h]{0.118\linewidth}
		\centering
		\includegraphics[width=\linewidth]{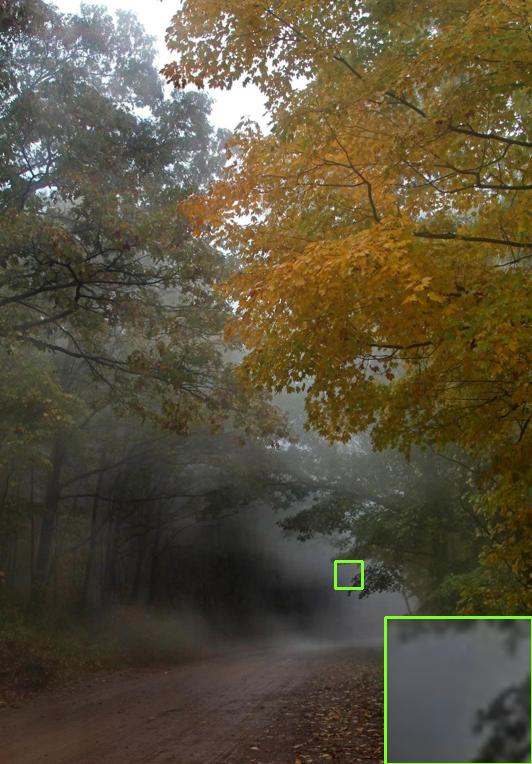}
	\end{minipage}
	\begin{minipage}[h]{0.118\linewidth}
		\centering
		\includegraphics[width=\linewidth]{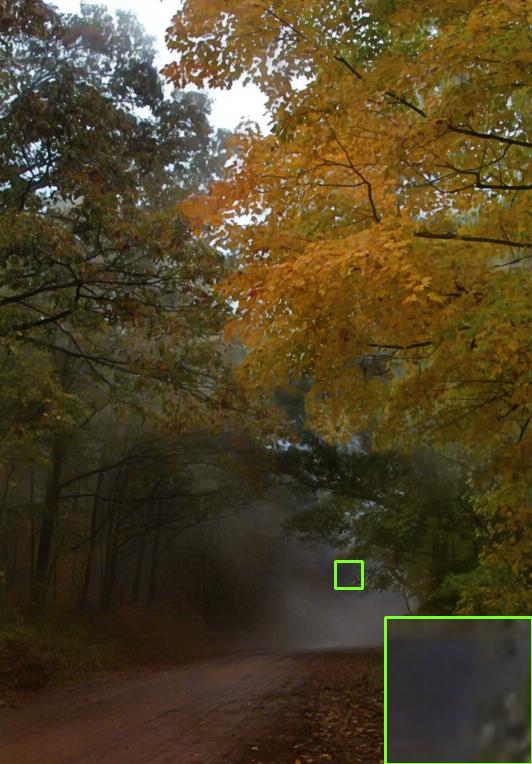}
	\end{minipage}
	\vspace{1mm}
	% URHI 4
	\begin{minipage}[h]{0.118\linewidth}
		\centering
		\includegraphics[width=\linewidth]{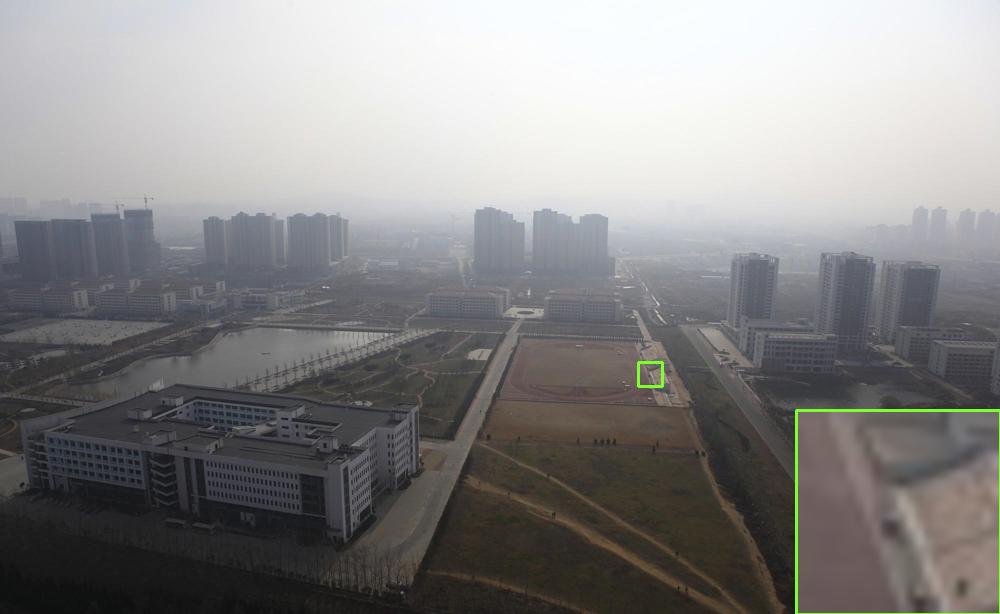}
		\scriptsize{(a) Hazy inputs}
	\end{minipage}
	\begin{minipage}[h]{0.118\linewidth}
		\centering
		\includegraphics[width=\linewidth]{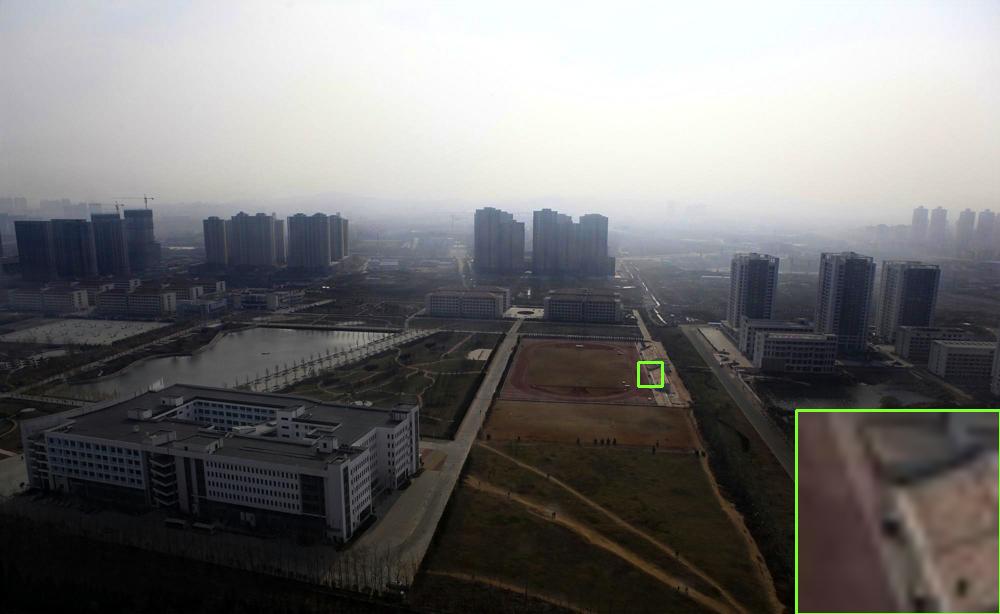}
		\scriptsize{(b) GFN}
	\end{minipage}
	\begin{minipage}[h]{0.118\linewidth}
		\centering
		\includegraphics[width=\linewidth]{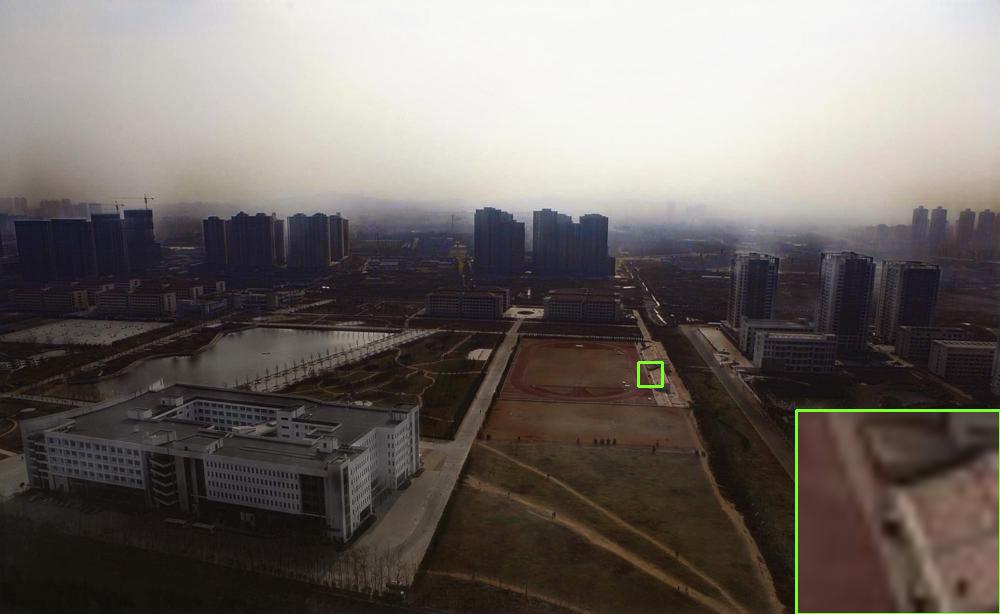}
		\scriptsize{(c) EPDN}
	\end{minipage}
	\begin{minipage}[h]{0.118\linewidth}
		\centering
		\includegraphics[width=\linewidth]{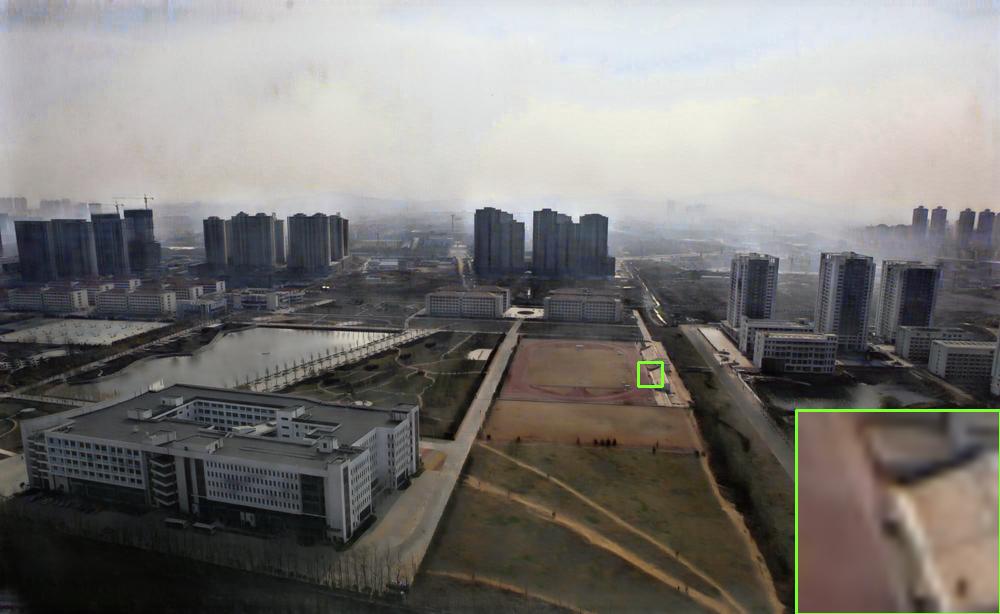}
		\scriptsize{(d) DADN}
	\end{minipage}
	\begin{minipage}[h]{0.118\linewidth}
		\centering
		\includegraphics[width=\linewidth]{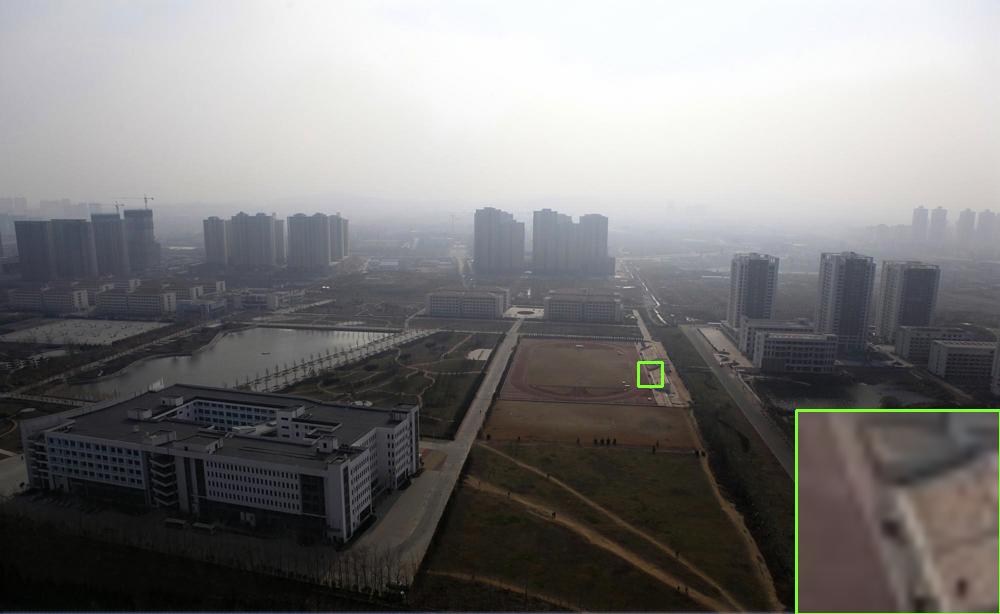}
		\scriptsize{(e) KDDN}
	\end{minipage}
	\begin{minipage}[h]{0.118\linewidth}
		\centering
		\includegraphics[width=\linewidth]{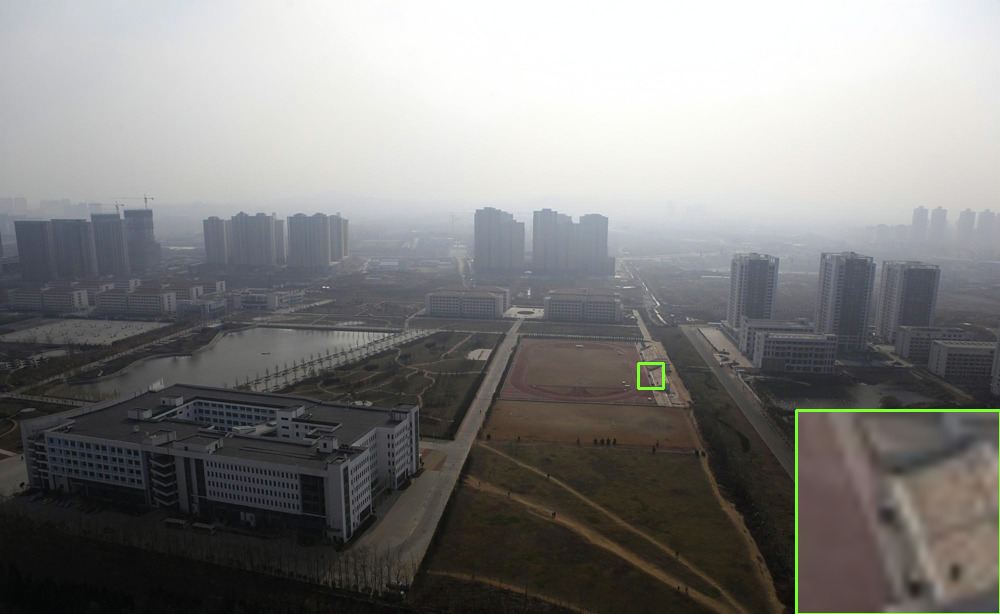}
		\scriptsize{(f) ACER-Net}
	\end{minipage}
	\begin{minipage}[h]{0.118\linewidth}
		\centering
		\includegraphics[width=\linewidth]{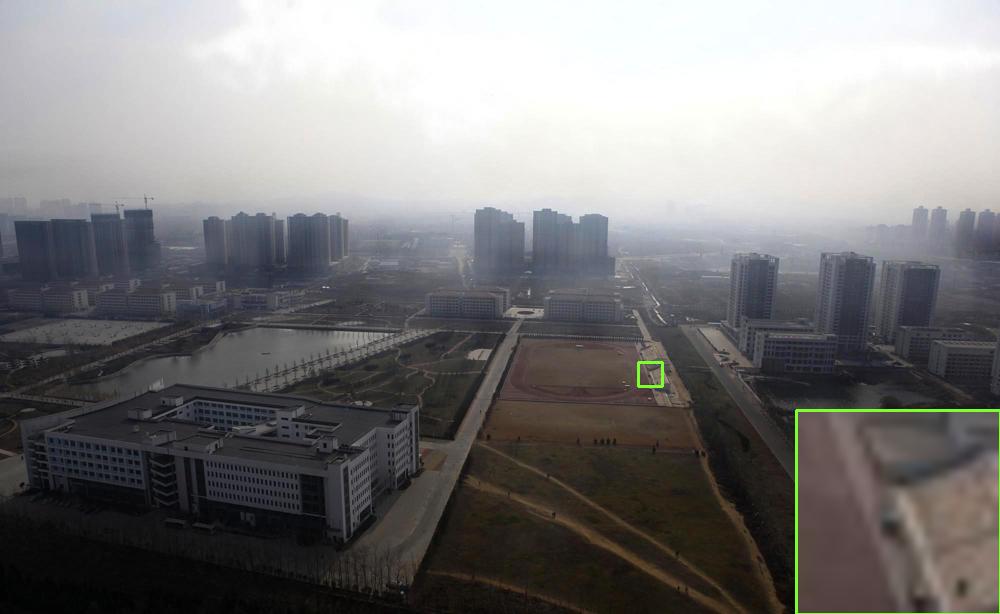}
		\scriptsize{(g) GDN}
	\end{minipage}
	\begin{minipage}[h]{0.118\linewidth}
		\centering
		\includegraphics[width=\linewidth]{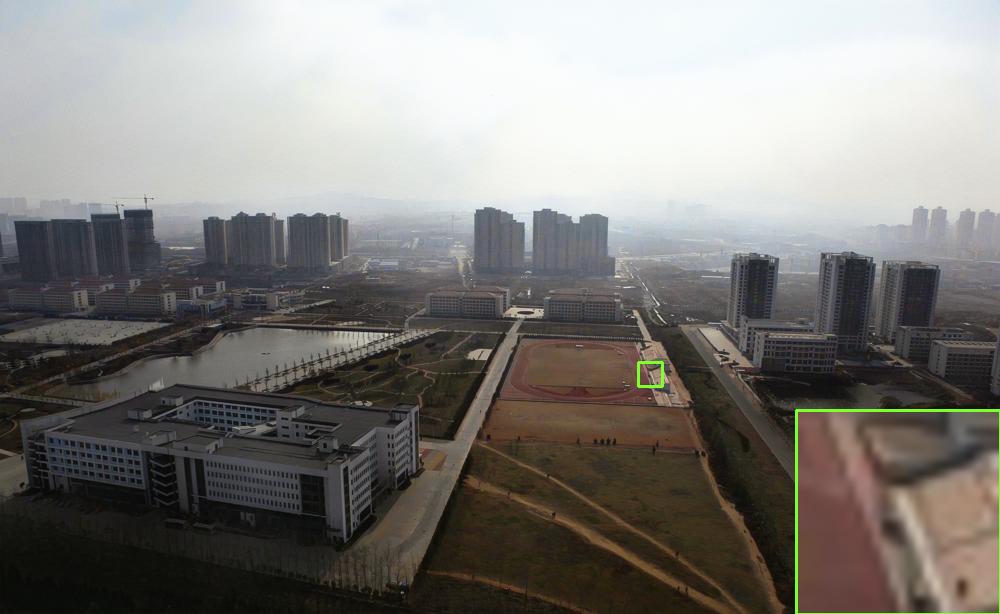}
		\scriptsize{(h) GDN+}
	\end{minipage}
%	\vspace{1mm}
%	% URHI 5
%	\begin{minipage}[h]{0.118\linewidth}
%		\centering
%		\includegraphics[width=\linewidth]{images/real/URHI/5/hazy_CQ_Baidu_009.jpg}
%		\scriptsize{(a) Hazy inputs}
%	\end{minipage}
%	\begin{minipage}[h]{0.118\linewidth}
%		\centering
%		\includegraphics[width=\linewidth]{images/real/URHI/5/AOD_CQ_Baidu_009.jpg}
%		\scriptsize{(b) AOD-Net}
%	\end{minipage}
%	\begin{minipage}[h]{0.118\linewidth}
%		\centering
%		\includegraphics[width=\linewidth]{images/real/URHI/5/GFN_CQ_Baidu_009.jpg}
%		\scriptsize{(c) GFN}
%	\end{minipage}
%	\begin{minipage}[h]{0.118\linewidth}
%		\centering
%		\includegraphics[width=\linewidth]{images/real/URHI/5/EPDN_CQ_Baidu_009.jpg}
%		\scriptsize{(d) EPDN}
%	\end{minipage}
%	\begin{minipage}[h]{0.118\linewidth}
%		\centering
%		\includegraphics[width=\linewidth]{images/real/URHI/5/DA_CQ_Baidu_009.jpg}
%		\scriptsize{(e) DADN}
%	\end{minipage}
%	\begin{minipage}[h]{0.118\linewidth}
%		\centering
%		\includegraphics[width=\linewidth]{images/real/URHI/5/KDDN_CQ_Baidu_009.jpg}
%		\scriptsize{(f) KDDN}
%	\end{minipage}
%	\begin{minipage}[h]{0.118\linewidth}
%		\centering
%		\includegraphics[width=\linewidth]{images/real/URHI/5/GDN_CQ_Baidu_009.jpg}
%		\scriptsize{(g) GDN}
%	\end{minipage}
%	\begin{minipage}[h]{0.118\linewidth}
%		\centering
%		\includegraphics[width=\linewidth]{images/real/URHI/5/GDN+_CQ_Baidu_009.jpg}
%		\scriptsize{(h) GDN+}
%	\end{minipage}
	\caption{Visual comparisons of different methods on real-world hazy images: the ones in the first 2 rows are from $37$Real, and the rest are from URHI. Zoom in for details.}
	\label{fig:real}
\end{figure*}

\begin{figure*}[t]
	\centering
	\begin{minipage}[h]{0.16\linewidth}
		\centering
		\includegraphics[width=\linewidth]{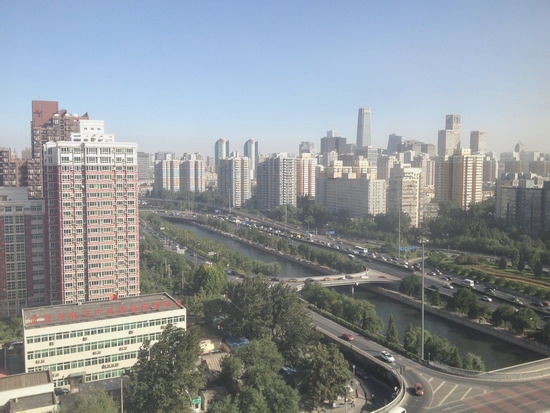}
		\scriptsize{(a) Hazy image}
	\end{minipage}
	\begin{minipage}[h]{0.16\linewidth}
		\centering
		\includegraphics[width=\linewidth]{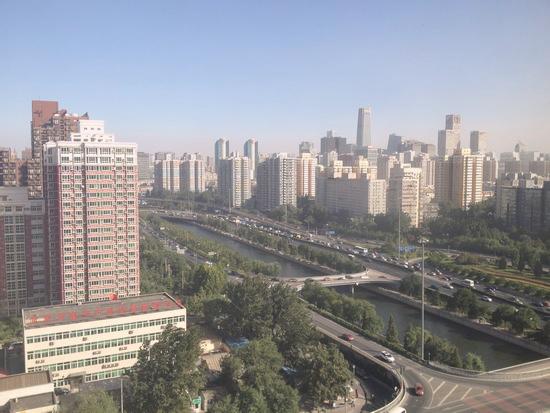}
		\scriptsize{(b) WB}
	\end{minipage}
	\begin{minipage}[h]{0.16\linewidth}
		\centering
		\includegraphics[width=\linewidth]{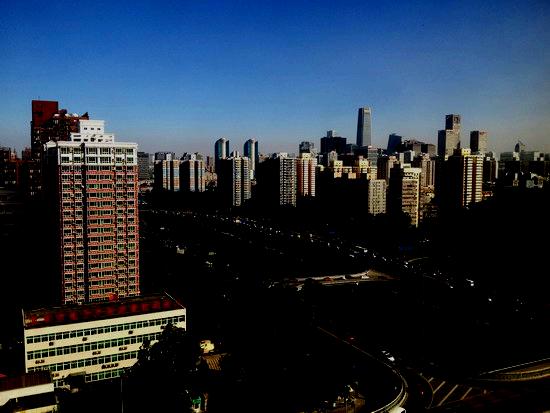}
		\scriptsize{(c) CE}
	\end{minipage}
	\begin{minipage}[h]{0.16\linewidth}
		\centering
		\includegraphics[width=\linewidth]{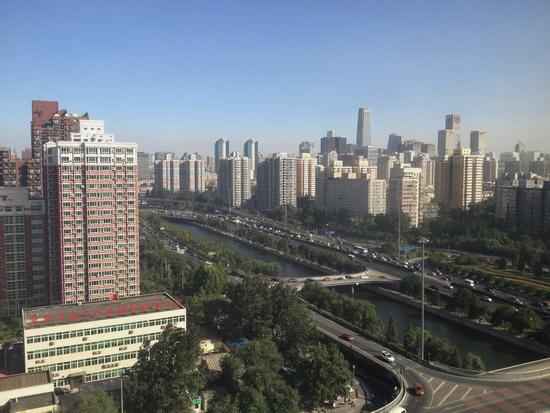}
		\scriptsize{(d) GC ($\gamma=1.5$)}
	\end{minipage}
	\begin{minipage}[h]{0.16\linewidth}
		\centering
		\includegraphics[width=\linewidth]{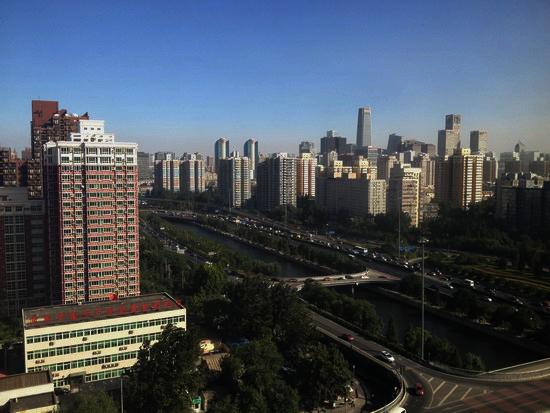}
		\scriptsize{(e) GC ($\gamma=2.5$)}
	\end{minipage}
	\begin{minipage}[h]{0.16\linewidth}
		\centering
		\includegraphics[width=\linewidth]{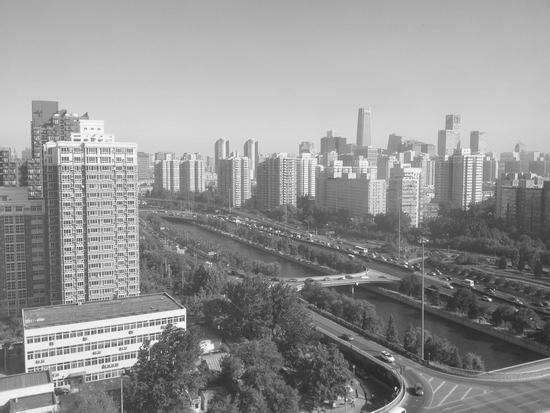}
		\scriptsize{(f) GS}
	\end{minipage}
	\begin{minipage}[h]{0.16\linewidth}
		\centering
		\includegraphics[width=\linewidth]{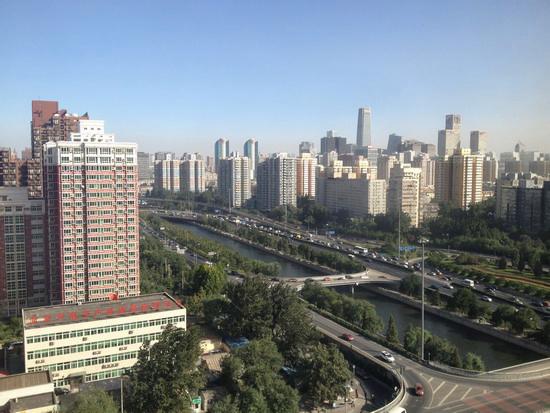}
		\scriptsize{(g) Dehazed image}
	\end{minipage}
	\begin{minipage}[h]{0.16\linewidth}
		\centering
		\includegraphics[width=\linewidth]{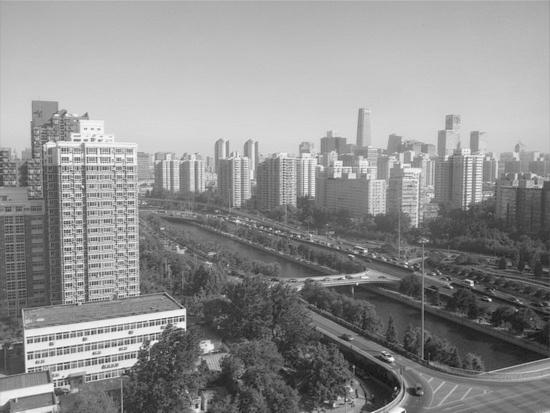}
		\scriptsize{(h) Learned input $\#1$}
	\end{minipage}
	\begin{minipage}[h]{0.16\linewidth}
		\centering
		\includegraphics[width=\linewidth]{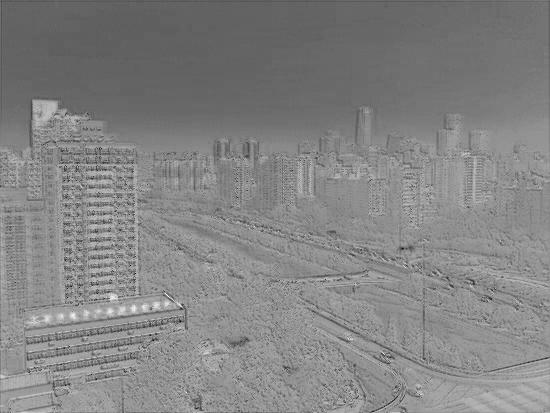}
		\scriptsize{(i) Learned input $\#3$}
	\end{minipage}
	\begin{minipage}[h]{0.16\linewidth}
		\centering
		\includegraphics[width=\linewidth]{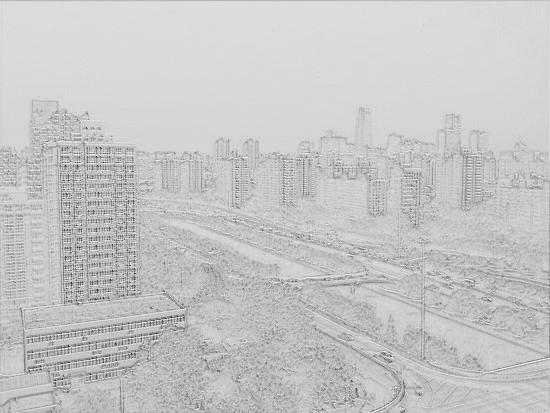}
		\scriptsize{(j) Learned input $\#4$}
	\end{minipage}
	\begin{minipage}[h]{0.16\linewidth}
		\centering
		\includegraphics[width=\linewidth]{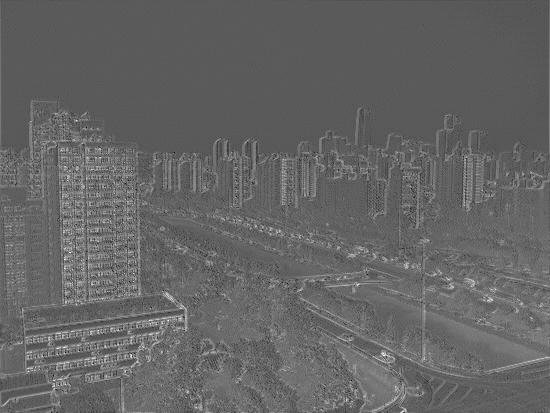}
		\scriptsize{(k) Learned input $\#5$}
	\end{minipage}
	\begin{minipage}[h]{0.16\linewidth}
		\centering
		\includegraphics[width=\linewidth]{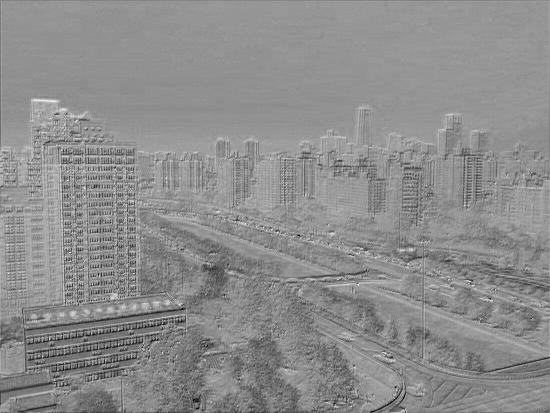}
		\scriptsize{(l) Learned input $\#15$}
	\end{minipage}
	%	\vspace*{0.3em}
	\caption{Visualization of the derived and learned inputs for a hazy image from SOTS.}
	\label{fig:CNN_priors}
\end{figure*}

\subsection{Evaluations on Synthetic Data}
We conduct evaluations on the $4$ synthetic datasets, \textit{i.e.}, SOTS, Middlebury, HazeRD, and O-HAZE. Comparisons in terms of average PSNR/SSIM values can be found in Tab~\ref{tab:syn1}. It is evident that the proposed GDN+ outperforms all other methods chosen for comparison, and has a significant improvement over its preliminary version GDN (\textit{e.g.}, $4.4$ dB on SOTS). Besides, for each method, we also demonstrate the number of trainable parameters in million (M). The proposed GDN+ has much fewer parameters than ACER-Net and DADN, and the comparison between GDN+ and GDN reveals that the adoption of self-attentions greatly improves the dehazing performance with a trivial impact on the model size (\textit{i.e.}, $+ 0.003$M).

%We conduct evaluations on the $4$ synthetic datasets, \textit{i.e.}, SOTS, Middlebury, HazeRD, and O-HAZE. Comparisons in terms of average PSNR/SSIM values can be found in  Tab~\ref{tab:syn1}. It is evident that the proposed GDN+ outperforms all other methods chosen for comparison, and has a significant improvement over its preliminary version GDN (\textit{e.g.}, $4.4$ dB on SOTS). Besides, for each method, we also demonstrate the number of trainable parameters in million (M), and the runtime in second (s) computed on a $1,080$P fake dataset, where all pixel values are set to $1$. Except for DCP that only works on CPU, the runtime of all other methods are tested on GPU. The proposed GDN+ has much fewer parameters than ACER-Net and DADN, and our un-optimized code takes about $0.7s$ to process one $1,080$P image that is faster than DADN but slower than ACER-Net. The comparison between the GDN+ and GDN reveals that although the adoption of self-attentions greatly improves the dehazing performance with a minimal impact on the model size (\textit{i.e.}, $+ 0.03$M), it comes at the cost of increasing the runtime (\textit{i.e.}, $+0.3$s).

Visual comparisons on the testing datasets can be found in Fig.~\ref{fig:synthetic}. 
%Note that the AOD-Net suffers from the halo artifacts in some cases (\textit{e.g.}, the $5$th row of Fig.~\ref{fig:synthetic} (b)). 
Note that GFN and EPDN exhibit limited performance on the images with dense haze (\textit{e.g.}, the $7$th row of Figs.~\ref{fig:synthetic} (b-c)). DADN tends to cause color distortions (\textit{e.g.}, the $4$th and $5$th rows in Fig.~\ref{fig:synthetic} (d)). KDDN and ACER-Net still retain a non-negligible amount of haze in some cases (\textit{e.g.}, the $4$st row of Fig.~\ref{fig:synthetic} (e-f)). In comparison, the dehazing results of GDN+ are visually most similar to the ground-truth as they are free of color distortion and contain very little residual haze. 

\begin{table}[t]
	\caption{Quantitative Evaluations on two real datasets using the FADE metric. The lower value indicates the better dehazing performance. We highlight the best one in \textbf{bold}.}
	\small
	\centering
	\begin{tabular}{c|ccccc}
		%\toprule
		\hline
		Method & Hazy & DADN & ACER-Net & GDN & GDN+ \\ \hline
		37Real & $1.0037$ & $0.5788$ & $0.6961$ & $0.6290$ & $\mathbf{0.5184}$ \\
		URHI & $2.3230$ & $1.0561$ & $1.5927$ & $1.4852$ & $\mathbf{0.9094}$ \\
		\hline
		%\bottomrule
	\end{tabular}
	\label{tab:real}
\end{table}

\subsection{Evaluation on Real Data}
We further compare the GDN+ with other methods on $2$ real datasets, \textit{i.e.}, 37Real and URHI. The results are shown in Table~\ref{tab:real}, where the lower value indicates the better dehazing performance. It is evident that GDN+ surpasses the SOTAs on FADE, and the dehazing performance on real data is significantly improved as compared to GDN. This improvement can be attributed to the proposed ITKT mechanism that successfully alleviates domain shift between synthetic data and real data.

%Here we shall only make qualitative evaluations since the haze-free counterparts of the real hazy images in these two datasets are not available. 

%The AOD-Net again suffers from the halo artifacts (\textit{e.g.}, the $6$th rows of Fig.~\ref{fig:real} (b)). 
Besides quantitative evaluations, visual comparisons can be found in Fig.~\ref{fig:real} as well. The dehazing results on real data are quite consistent with those on synthetic data. GFN and EPDN have difficulty in dealing with dense haze (\textit{e.g.}, the $4$th rows of Figs.~\ref{fig:real} (b-c)). DADN may cause severe color distortion (\textit{e.g.}, the $2$th row of Fig.~\ref{fig:real} (d)). Due to domain shift, KDDN and ACER-Net fail to produce desirable dehazing results on real-world hazy images.

Compared to these methods, the proposed GDN+ removes haze more thoroughly and is free of color distortion. The GDN+ also delivers better dehazing results on real data than GDN.

%by to assist the learning process

%where the translated data is utilized in training and its learning process is assisted by the distilled knowledge from the corresponding synthetic data.

%As compared to GDN, the dehazing performance on real data is largely improved owing to the proposed ITKT, where the translated data is utilized in training and its learning process is assisted by the distilled knowledge from the corresponding synthetic data.

%Moreover, we intentionally select one nighttime hazy image in testing. As shown in the last row of Fig.~\ref{fig:real}, even without the consideration of nighttime hazy image in training, our dehazing result still performs favorably against the SOTAs. This validates that the proposed GDN+ has good generalization ability to the nighttime haze.

\subsection{Necessity of Atmosphere Scattering Model}\label{sec:asm}

\begin{table}[t]
	\caption{Quantitative Comparisons of Different Estimation Strategies. The best one is highlighted in \textbf{bold}.}
	\small
	\centering
	\begin{tabular}{l|lllllllll}
		%\toprule
		\hline
		Method & Indirect & Direct (GDN+) \\ \hline
		SOTS & $31.82$/$0.9732$ & $\mathbf{32.15}$/$\mathbf{0.9777}$ \\
		HazeRD & $17.81$/$0.8455$ & $\mathbf{19.54}$/$\mathbf{0.8697}$ \\
		\hline
		%\bottomrule
	\end{tabular}
	\label{tab:estimation}
\end{table}

To gain a better understanding of the difference between the direct estimation strategy where the ASM is completely bypassed (denoted as \textit{Direct}), and the indirect estimation strategy where the transmission map and the atmospheric light intensity are first estimated in order to calculate the dehazing result (denoted as \textit{Indirect}), we adjust the GDN+ to make it follow the indirect estimation strategy instead. Specifically, we modify the convolution at the output end ({\textit{i.e.}, the rightmost Conv@K$3$S$1$ in Fig.~\ref{fig:GDN_main}) so that it outputs two feature maps rather than three. The first feature map is used as the estimated transmission map while the mean of the second one serves as the estimated atmospheric light intensity. These two estimates are then substituted into Eq.~(\ref{eq:asm}) to calculate the dehazing result. This variant of GDN+ is trained in the same way as detailed in Sec.~\ref{sec:implementation}. It is then quantitatively evaluated on SOTS and HazeRD, and compared with its original version. Note that both SOTS and HazeRD are synthetic datasets based on ASM. Therefore, as far as this kind of testing datasets are concerned, the indirect estimation strategy essentially takes advantage of the ASM as a perfect prior. However, as shown in Tab.~\ref{tab:estimation}, although adopting the ASM leads to a significant reduction in the number of parameters to be estimated, it in fact incurs  performance degradation. This indicates that incorporating the ASM into GDN+ does have a detrimental effect on the loss surface.

% Moreover, even with dimensional reduction, accurately estimating the transmission map for a hazy image is still challenging, and there is no evidence that this task is easier than the dehazing task. We visualize the estimated transmission map for a synthetic hazy image in Fig~\ref{fig:indirect} as an example. The quality of this map is awful that influences the final dehazing results. 

\begin{table}[t]
	\caption{Quantitative Comparisons of Different Input Types. The best one is highlighted in \textbf{bold}.}
	\small
	\centering
	\begin{tabular}{l|lllllllll}
		%\toprule
		\hline
		Method & Original & Derived & Learned (GDN+) \\ \hline
		SOTS & $30.04$/$0.9697$ & $29.08$/$0.9635$ & $\mathbf{32.15}$/$\mathbf{0.9777}$ \\
		HazeRD & $17.69$/$0.8486$ & $18.04$/$0.8580$ & $\mathbf{19.54}$/$\mathbf{0.8697}$ \\
		\hline
		%\bottomrule
	\end{tabular}
	\label{tab:input}
\end{table}

\begin{table*}[t]
	\caption{Quantitative Comparisons of Different Variants of GDN+. The best one is highlighted in \textbf{bold}.}
	\small
	\centering
	\begin{tabular}{l|ll|lll|l|l}
		%\toprule
		\hline
		Method & EDNet & MSNet & w/o SCAB & w/o CAB & w/o SAB  & w/o post-processing & Our GDN+ \\ \hline
		SOTS & $25.98$/$0.9446$ & $27.65$/$0.9467$ & $27.85$/$0.9657$ & $29.31$/$0.9734$ & $31.80$/$0.9759$ & $31.79$/$0.9755$ & $\mathbf{32.15}$/$\mathbf{0.9777}$ \\
		HazeRD & $15.86$/$0.8182$ & $16.51$/$0.8208$ & $15.91$/$0.8264$ & $16.12$/$0.8295$ & $19.05$/$0.8648$ & $19.28$/$0.8669$ &$\mathbf{19.54}$/$\mathbf{0.8697}$\\
		\hline
		%\bottomrule
	\end{tabular}
	\label{tab:syn3}
\end{table*}

\subsection{Utility of Learned Inputs}

The pre-processing module of GDN+ produces $16$ learned inputs in total. Here we build two variants of GDN+ to demonstrate the diversity gain offered by these learned inputs. For the first variant (denoted as \textit{Original}), we remove the pre-processing module and replace the first $3$ learned inputs by the RGB channels of the given RGB hazy image and the rest by all-zero feature maps. For the second variant (denoted as \textit{Derived}), the learned inputs are substituted with the same number of derived inputs generated by hand-selected pre-processing methods. More specifically, we generate $16$ derived inputs, $3$ from the given hazy image, $3$ from the White Balanced (WB) image, $3$ from the Contrast Enhanced (CE) image, $6$ from two Gamma Corrected (GC) images with $\gamma$ set to 1.5 and 2.5 respectively, and $1$ from the Gray-Scaled (GS) image. Fig.~\ref{fig:CNN_priors} shows the derived and learned inputs of a hazy image.

Although the hand-selected pre-processing methods can create diversified inputs,  our pre-processing module is considerably more flexible and adaptive in finetuning the given image to better suit the follow-up process (\textit{e.g.,} the learned inputs $\#3$ and $\#5$ enhance different aspects of the given hazy image and are complement to each other). More interestingly, the learned input $\#1$ resembles a GS image, even though this is not prescribed. This shows that our pre-processing module is capable of mimicking hand-selected pre-processing methods when it is beneficial to do so.

To further validate the effectiveness of learned inputs, we follow the same experimental setup to train both variants, and quantitatively evaluate their dehazing performance on the SOTS and HazeRD. Tab.~\ref{tab:input} shows that the GDN+ with learned inputs (denoted as \textit{Learned}) outperforms the \textit{Original} and \textit{Derived} versions in terms of PSNR and SSIM metrics.

\subsection{Validation of Overall Design}

The proposed GDN+ is a multi-scale network that is enhanced in two aspects: 1)  a grid structure with dense connections across different scales to facilitate the information exchange, and 2) a novel SCAB that is capable of fusing features based on their relative importance. To demonstrate the effectiveness of the adopted grid structure, we consider the following two variants: 1) the Encoder-Decoder Network (\textit{EDNet}) obtained by pruning the GDN+ (see the red path in Fig.~\ref{fig:GDN_main}), and 2) the conventional multi-scale network (\textit{MSNet}) that removes all exchange branches except for the first and the last ones in order to maintain the minimum connection. To validate the proposed SCAB, we consider the following three variants: 1) the GDN+ without SCABs (\textit{w/o SCAB}), 2) the GDN+ with CAB-absent SCABs (\textit{w/o CAB}), and 3) the GDN+ with SAB-absent SCABs (\textit{w/o SAB}). In addition, we build a variant of GDN+ that has no post-processing module (\textit{w/o post-processing}). All these variants are trained in the same way as before and are tested on the SOTS and HazeRD.  

The quantitative comparisons are shown in Table~\ref{tab:syn3}. Compared to \textit{EDNet} and \textit{MSNet}, the proposed GDN+ achieves favorable dehazing results owing to the superiority of the grid structure. Besides, it can be seen that the variants~\textit{w/o SAB} and \textit{w/o CAB} both outperform the baseline \textit{w/o SCAB} though the performance gain from CAB appears to be more significant than that from SAB. Benefiting from the contributions of both CAB and SAB, the GDN+ with SCABs delivers further elevated performance.  As compared to GDN+, the dehazing performance of \textit{w/o post-processing} is inferior owing to the potential residual artifacts from the backbone module. The above results provide a fairly comprehensive justification for the overall design of GDN+.

%From our experiments, the overall design in GDN+  is validated.
%In Fig.~\ref{label}, we show the dehazing results from different variants for a hazy image on SOTS. Our GDN+ successively removes the haze while the others remain part of the haze in their dehazing results.

\subsection{Effectiveness of Intra-Task Knowledge Transfer} \label{sec:ITKT}

To convincingly demonstrate the effectiveness of the proposed ITKT mechanism, we consider a variant (\textit{w/o ITKT}) that trains the GDN+ directly on translated data. We also convert the original SOTS to a translated version, named SOTS-T, for quantitative comparisons in terms of PSNR and SSIM metrics. Besides \textit{w/o ITKT} and \textit{w/ ITKT} (\textit{i.e., GDN+}), the GDN+ pre-trained on synthetic data (\textit{pre-trained}) is also tested on SOTS-T.

According to Tab.~\ref{tab:ITKT},  the synthetic domain knowledge does benefit  the learning process on translated data. Indeed, \textit{w/ ITKT} achieves higher PSNR and SSIM values on SOTS-T as compared to \textit{w/o ITKT}. Also from Figs.~\ref{fig:ITKT} (c-d), \textit{w/ ITKT} removes haze more thoroughly than \textit{w/o ITKT}, and produces more appealing dehazing results. As for \textit{pre-trained}, although it works well on synthetic data, the dehazing performance on real data is rather limited as shown in Fig.~\ref{fig:ITKT} (b). This dramatic performance drop is owing to the domain shift between training and testing data. Therefore, it is necessary to conduct training on real data or those with (approximately) the same distribution. This is exactly the rationale of creating and utilizing the translated data to finetune the GDN+.

It is worth emphasizing that the proposed ITKT is generic in nature and can be easily employed in other learning-based dehazing methods to improve their performance on real-world hazy images.

 %train the dehazing network with the translated data that conforms the same haze distribution as in the real scenario. This fact also motivates us to propose the teacher-student structure with ITKT. 

 \begin{table}[t]
	\caption{Quantitative Comparisons of ITKT-Related Variants on SOTS-T. The best one is highlighted in \textbf{bold}.}
	\small
	\centering
	\begin{tabu}{l|lllllllll}
		%\toprule
		\hline
		Method & pre-trained & w/o ITKT  & w/ ITKT (GDN+) \\ \hline
		SOTS-T & $21.08$/$0.8004$ & $23.70$/$0.8606$ & $\mathbf{24.66}$/$\mathbf{0.8751}$ \\
		\hline
		%\bottomrule
	\end{tabu}
	\label{tab:ITKT}
\end{table}

\begin{figure}[t]
	\centering
	\begin{minipage}[b]{0.24\linewidth}
		\centering
		\includegraphics[width=\linewidth]{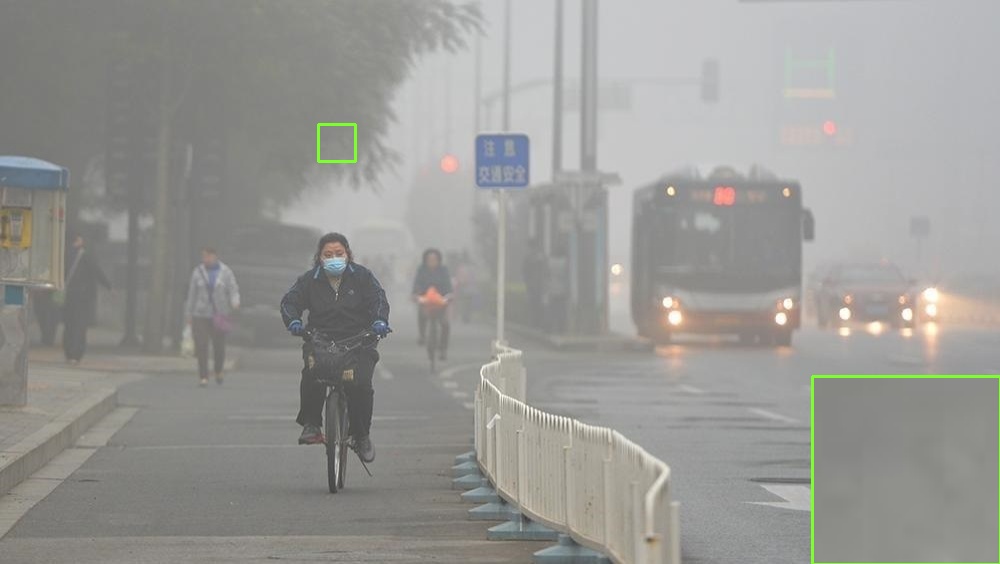}
	\end{minipage}
	\begin{minipage}[b]{0.24\linewidth}
		\centering
		\includegraphics[width=\linewidth]{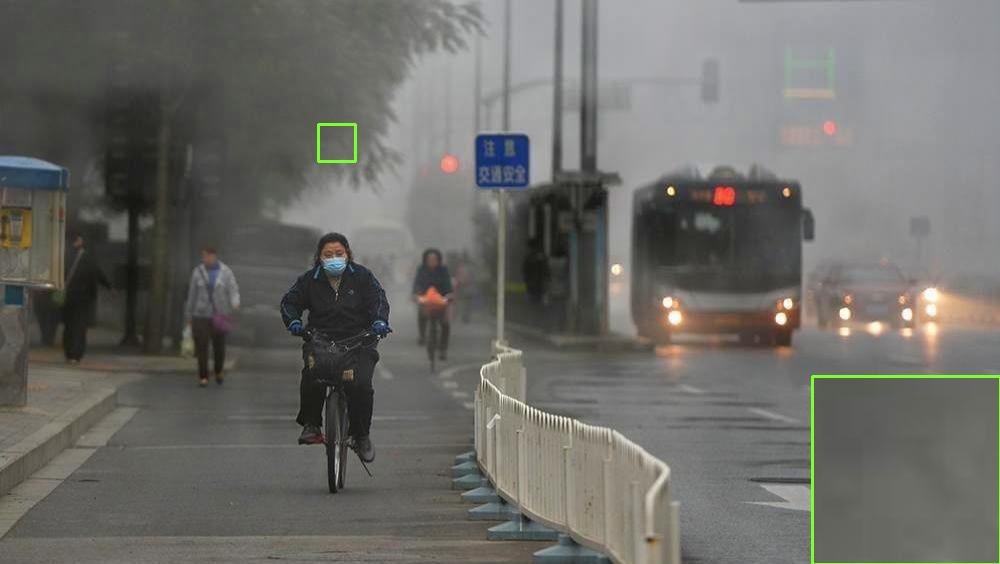}
	\end{minipage}
	\begin{minipage}[b]{0.24\linewidth}
		\centering
		\includegraphics[width=\linewidth]{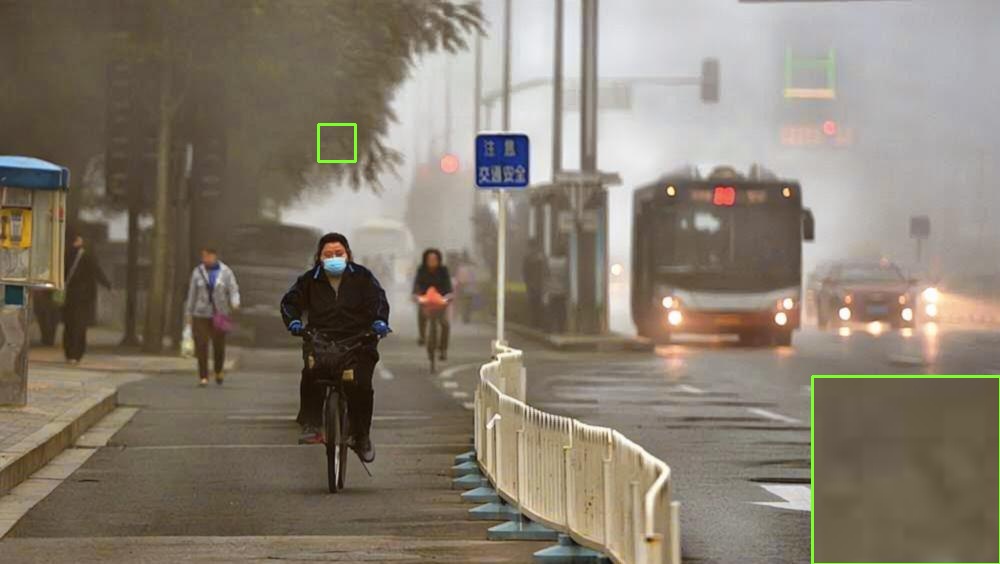}
	\end{minipage}
	\vspace{1mm}
	\begin{minipage}[b]{0.24\linewidth}
		\centering
		\includegraphics[width=\linewidth]{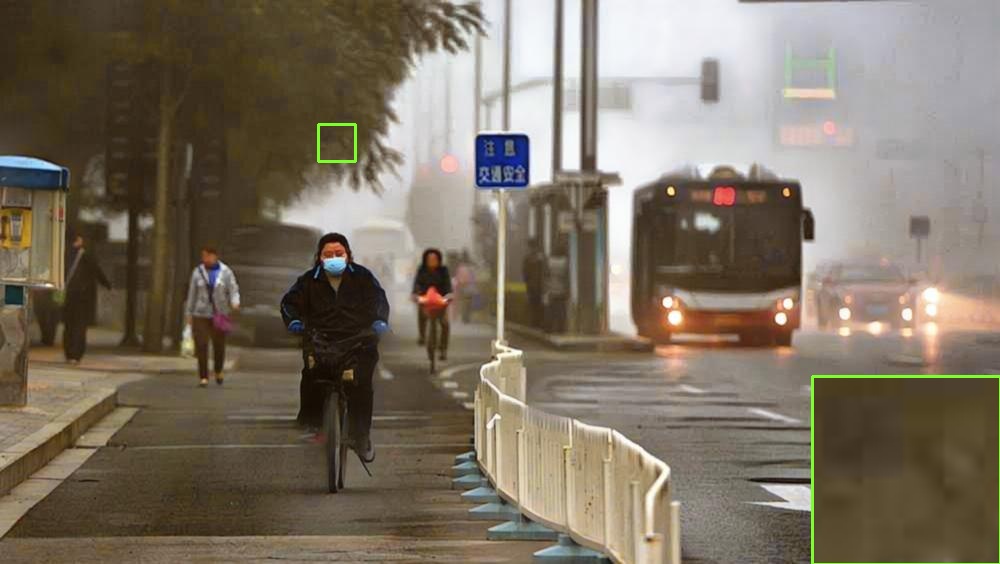}
	\end{minipage}
	\begin{minipage}[b]{0.24\linewidth}
		\centering
		\includegraphics[width=\linewidth]{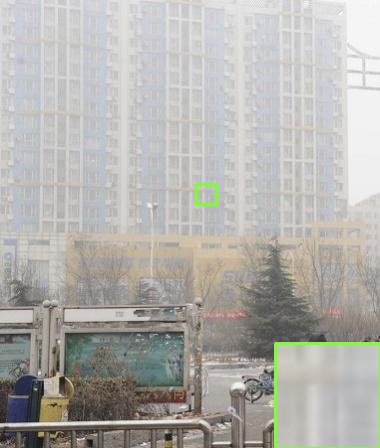}
		\scriptsize{(a) Hazy image}
	\end{minipage}
	\begin{minipage}[b]{0.24\linewidth}
		\centering
		\includegraphics[width=\linewidth]{images/ITKT/1/BJ_Bing_483_hazy.jpg}
		\scriptsize{(b) Pre-trained}
	\end{minipage}
	\begin{minipage}[b]{0.24\linewidth}
		\centering
		\includegraphics[width=\linewidth]{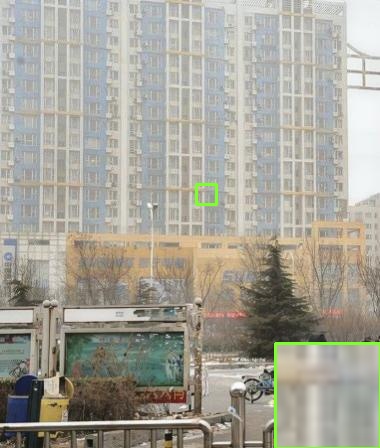}
		\scriptsize{(c) w/o ITKT}
	\end{minipage}
	\begin{minipage}[b]{0.24\linewidth}
		\centering
		\includegraphics[width=\linewidth]{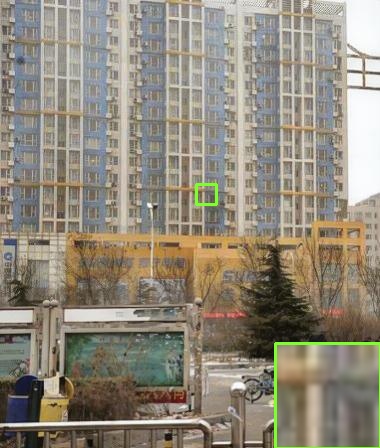}
		\scriptsize{(d) w/ ITKT}
	\end{minipage}

	%	\vspace*{0.3em}
	\caption{Qualitative comparisons of ITKT-related variants on real hazy images from URHI.}
	\label{fig:ITKT}
\end{figure}

\section{Conclusion}
%-------------------------------------------------------------------------
We have proposed an enhanced multi-scale network and demonstrated its competitive performance for single image dehazing. The design of this network involves several ideas.
We adopt a densely connected  grid structure to facilitate the information exchange  across different scales. A Novel SCAB, constructed based on the idea of self-attentions, is placed at the junctions of the grid structure to enable adaptive feature fusion. The issue of domain shift is addressed by converting synthetic data to translated data with the distribution matched to that of real-world hazy images. We further propose a novel ITKT mechanism that leverages the synthetic domain knowledge to assist the learning process on translated data.

%To address the domain shift, we convert the synthetic data to the translated data that conforms the real haze distribution, and propose a new Intra-Task Knowledge Transferring (ITKT) that leverages the distilled knowledge from synthetic data to assist the learning process on translated data. 

%Extensive experiments validate the overall design of our network. 

Due to the generic nature of its building components, the proposed network is expected to be applicable to a wide range of image restoration problems. Investigating such applications is an endeavor well worth undertaking. 

Our work also sheds some light on the puzzling phenomenon regarding the use of the ASM in image dehazing, and suggests the need to rethink the role of physical models in the design of image restoration algorithms.

%As for the future work, we plan to add a color correction processing that can deal with chromatic casts in real hazy images to further improve our dehazing results.

% if have a single appendix:
%\appendix[Proof of the Zonklar Equations]
% or
%\appendix  % for no appendix heading
% do not use \section anymore after \appendix, only \section*
% is possibly needed

% use appendices with more than one appendix
% then use \section to start each appendix
% you must declare a \section before using any
% \subsection or using \label (\appendices by itself
% starts a section numbered zero.)
%

%\appendices
%\section{Proof of the First Zonklar Equation}
%Appendix one text goes here.
%
%% you can choose not to have a title for an appendix
%% if you want by leaving the argument blank
%\section{}
%Appendix two text goes here.
%
%
%% use section* for acknowledgment
%\section*{Acknowledgment}
%
%
%The authors would like to thank...

% Can use something like this to put references on a page
% by themselves when using endfloat and the captionsoff option.
\ifCLASSOPTIONcaptionsoff
\newpage
\fi

% trigger a \newpage just before the given reference
% number - used to balance the columns on the last page
% adjust value as needed - may need to be readjusted if
% the document is modified later
%\IEEEtriggeratref{8}
% The "triggered" command can be changed if desired:
%\IEEEtriggercmd{\enlargethispage{-5in}}

% references section

% can use a bibliography generated by BibTeX as a .bbl file
% BibTeX documentation can be easily obtained at:
% http://mirror.ctan.org/biblio/bibtex/contrib/doc/
% The IEEEtran BibTeX style support page is at:
% http://www.michaelshell.org/tex/ieeetran/bibtex/
%\bibliographystyle{IEEEtran}
% argument is your BibTeX string definitions and bibliography database(s)
%\bibliography{IEEEabrv,../bib/paper}
%
% <OR> manually copy in the resultant .bbl file
% set second argument of \begin to the number of references
% (used to reserve space for the reference number labels box)
%\begin{thebibliography}{1}
%
%\bibitem{IEEEhowto:kopka}
%H.~Kopka and P.~W. Daly, \emph{A Guide to \LaTeX}, 3rd~ed.\hskip 1em plus
%  0.5em minus 0.4em\relax Harlow, England: Addison-Wesley, 1999.
%
%\end{thebibliography}
\bibliographystyle{IEEEtran}
\bibliography{IEEEabrv,egbib}

% biography section
% 
% If you have an EPS/PDF photo (graphicx package needed) extra braces are
% needed around the contents of the optional argument to biography to prevent
% the LaTeX parser from getting confused when it sees the complicated
% \includegraphics command within an optional argument. (You could create
% your own custom macro containing the \includegraphics command to make things
% simpler here.)
%\begin{IEEEbiography}[{\includegraphics[width=1in,height=1.25in,clip,keepaspectratio]{mshell}}]{Michael Shell}
% or if you just want to reserve a space for a photo:

% You can push biographies down or up by placing
% a \vfill before or after them. The appropriate
% use of \vfill depends on what kind of text is
% on the last page and whether or not the columns
% are being equalized.

%\vfill

% Can be used to pull up biographies so that the bottom of the last one
% is flush with the other column.
%\enlargethispage{-5in}

% that's all folks
\end{document}